\documentclass[lettersize,journal]{IEEEtran}

\usepackage{amsmath,amsfonts}
\usepackage{tabularx}
\usepackage{algorithmic}
\usepackage{algorithm}
\usepackage{array}
\usepackage[caption=false,font=normalsize,labelfont=sf,textfont=sf]{subfig}
\usepackage{textcomp}
\usepackage{stfloats}
\usepackage{verbatim}
\usepackage{graphicx}
\usepackage{pifont}

\newcolumntype{C}[1]{>{\centering\arraybackslash}m{#1}}

\usepackage{cite}
\usepackage{multirow}
\usepackage{bbding}
\usepackage{pifont}
\usepackage{graphicx}
\usepackage{tikz}
\usepackage{multirow}
\usepackage{array}
\usepackage{caption}
\usepackage{colortbl}
\usepackage{makecell}
\usepackage{xcolor}
\usepackage{longtable}
\usepackage{ltxtable}
\usepackage{rotating}
\definecolor{datasetcolor}{RGB}{0,102,204}
\definecolor{networkcolor}{RGB}{255,140,0}
\definecolor{losscolor}{RGB}{0,128,0}
\usepackage{pgfplots}
\pgfplotsset{compat=newest}
\pgfplotsset{compat=1.16}
\usepackage{tikz}
\usepackage{pgfplots}
\pgfplotsset{compat=newest}
\usepackage{booktabs}   
\usepackage{pifont}     
\usepackage{tabularx}   
\usepackage{array} 
\usepackage{tabularx} 
\usepackage{ragged2e} 
\usepackage{booktabs} 
\usepackage{xcolor}   
\usepackage{caption}
\usepackage{array}
\usepackage{ragged2e}
\usepackage[table]{xcolor}
\usepackage{tabularx}
\usepackage{MnSymbol}
\usepackage{array}
\usepackage{placeins}

\usepackage[colorlinks=true, linkcolor=black, urlcolor=blue, citecolor=black]{hyperref}
\usepackage{cleveref}

\begin{document}

\title{Privacy-Preserving Action Recognition: Taxonomy, Methods, and Privacy–Utility Trade-offs}

\author{Sareer Ul Amin, Muhammad Ayaz, Muhammad Munsif, Sanghyun Seo

\thanks{
This research was Funded by the Ministry of Science and ICT (NRF, National Research Foundation of Korea), Mar. 2025~ Feb. 2030 (This work was supported by the National Research Foundation of Korea (NRF) grant funded by the Korea government (MSIT) (No. RS-2023-00218176)). (Corresponding author: Sanghyun Seo)}

\thanks{ Sareer Ul Amin and Muhammad Ayaz are with the Department of Applied Art and Technology, Chung-Ang University, Anseong 17546, South Korea. (Email: sareer2021@cau.ac.kr; ayaz2025@cau.ac.kr). }

\thanks{ Muhammad Munsif is with the Ulsan National Institute of Science and Technology (UNIST), Ulsan 44919, Republic of Korea. (Email: munsif@unist.ac.kr). }

\thanks{ Sanghyun Seo is with the College of Art and Technology, Chung-Ang University, Anseong, 17546, South Korea. (Email: sanghyun@cau.ac.kr). }

}



\maketitle

\begin{abstract}
Video surveillance in public safety, healthcare, and smart environments has made continuous human monitoring routine, raising real risks to personal identity and appearance. Privacy-preserving action recognition (PPAR) tackles the tension between the utility of video understanding and this exposure, and has drawn fast-growing interest. However, existing surveys remain narrow. Most catalog a single mechanism family, predate recent adversarial and hybrid work, or barely address evaluation. The result is a fragmented literature with incompatible threat models, inconsistent metrics, and no shared evaluation standard. We address this with a PRISMA-guided review of 32 peer-reviewed papers (2018--2026) drawn from 885 screened records. Methods sort into five families, namely adversarial learning (52\%), skeleton-based (20\%), cryptographic (12\%), differential privacy (8\%), and hybrid (8\%), each with distinct privacy, utility, and efficiency trade-offs. Evaluation is the weak point. Only 10\% of papers adopt a formal privacy definition, 65\% rely on ad-hoc metrics, and 40\% report an inconsistently defined cMAP. The trade-offs are steep. Skeleton methods reach about 85\% accuracy but drop appearance, adversarial methods hold near 80\% utility at moderate privacy (cMAP 0.9 to 0.3--0.5), and differential privacy often falls below 70\%. Harder conditions stay under-tested, with fewer than 15\% of papers checking cross-dataset generalization, under 10\% testing adaptive attackers, and real-time edge deployment nearly untouched. We contribute a two-dimensional privacy-space taxonomy, a formal threat model, a comparative trade-off analysis, the PPAR Unified Evaluation Protocol, and a roadmap centered on benchmark standardization. With this grounding, we argue PPAR can move from prototypes toward deployment, with lessons extending to face recognition and medical imaging.

\end{abstract}


\section{Introduction}
\begin{figure*}[t]
\centering
\includegraphics[width=\textwidth]{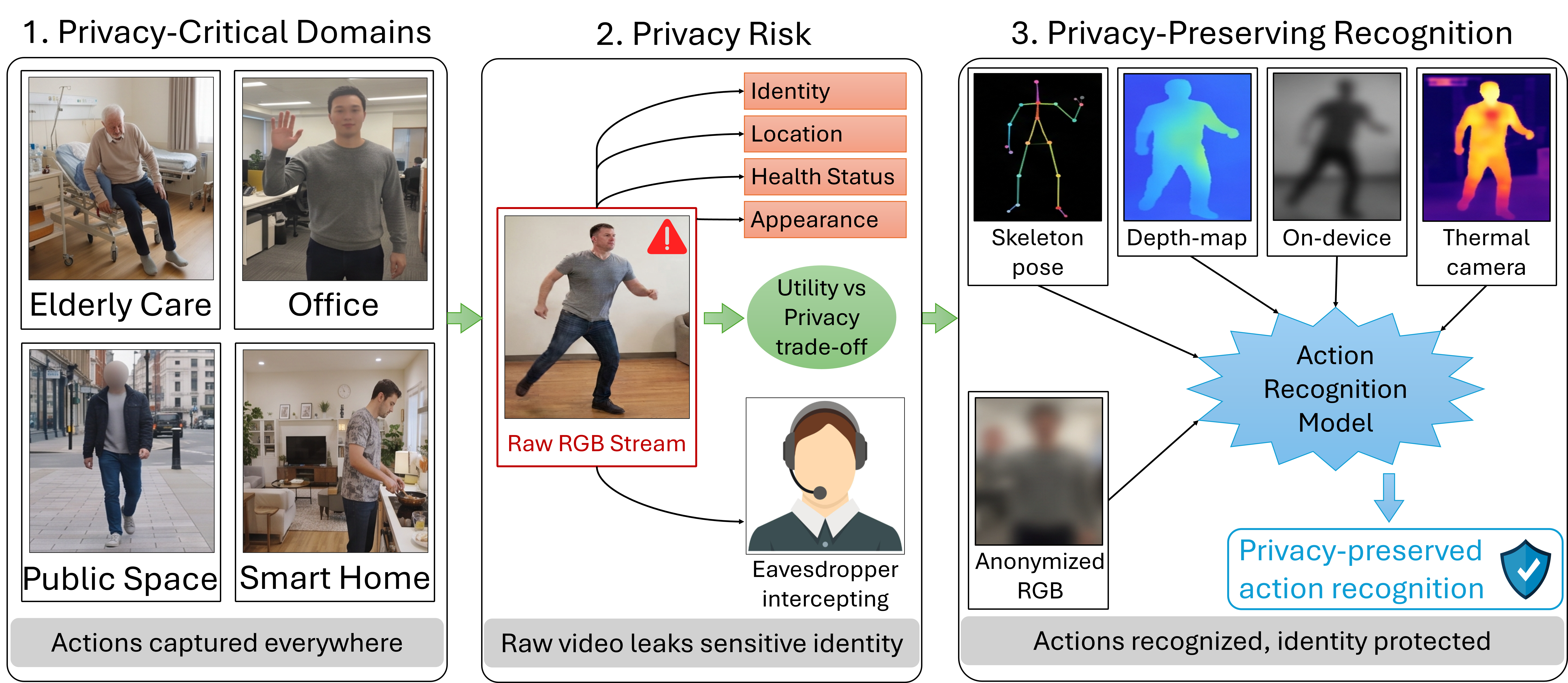}
\caption{Motivation for privacy-preserving action recognition (PPAR). \textbf{(Left)~Actions captured everywhere.} Everyday actions such as falling, waving, walking, and cooking are recorded across diverse settings such as elderly care, offices, public spaces, and smart homes. \textbf{(Center)~Raw video leaks sensitive identity.} A raw RGB stream reveals sensitive attributes such as identity, location, health status, and appearance. An eavesdropper can intercept these attributes, which creates a utility-versus-privacy trade-off. \textbf{(Right)~Actions recognized, identity protected.} The input is converted into privacy-preserving modalities such as skeleton pose, depth map, on-device processing, thermal camera, and anonymized RGB. These modalities are fed to the action recognition model, which enables privacy-preserving action recognition while hiding identity.}
\label{fig:motivation}
\end{figure*}

\IEEEPARstart{R}{ecognizing} human actions in video is central to modern visual analytics. It supports surveillance, anomaly detection, healthcare monitoring, smart homes, sports analysis, and human-computer interaction ~\cite{zhang_chaotic_2025,park_human_2025,gutev_depth-based_2025,gao_privacy-preserving_2025}. This capability comes at a serious privacy cost. The same visual features that enable accurate recognition, such as facial appearance, body geometry, gait, clothing, skin tone, and location context, also reveal identity, demographics, behavioral preferences, and other sensitive information~\cite{li2024privacy}, as illustrated in Fig. \ref{fig:motivation}. Video analytics is now moving toward large-scale deployment in public, commercial, and institutional settings. As a result, the tension between surveillance utility and individual privacy has become urgent and unavoidable. PPAR is the technical response to this tension. It recognizes actions accurately while actively suppressing the inference of sensitive attributes. Post-hoc anonymization methods such as face-blurring and pixelation degrade privacy and utility indiscriminately. PPAR instead pursues a principled co-design. It integrates privacy objectives into the learning pipeline, so that privacy-critical features are removed while action-discriminative information is preserved. The field covers several directions. These include adversarial learning that yields representations resistant to attribute inference, modality selection that replaces appearance-rich RGB with skeleton, depth, or motion representations, differential privacy adapted to video, and cryptographic and secure-computation approaches.
 
\definecolor{LightGray}{rgb}{0.9,0.9,0.9}
\newcolumntype{C}[1]{>{\centering\arraybackslash}m{#1}}
\newcolumntype{L}[1]{>{\raggedright\arraybackslash}m{#1}}
\newcolumntype{J}[1]{>{\justifying\noindent\arraybackslash}m{#1}}
\renewcommand\tabularxcolumn[1]{m{#1}}
\begin{table}[t]
\centering
\caption*{Notation used throughout the survey.}
\renewcommand{\arraystretch}{1.4}
\begin{tabularx}{\columnwidth}{@{}
        L{1.5cm}   
        >{\arraybackslash}X    
@{}}
\toprule
\textbf{Symbol} & \textbf{Meaning} \\
\midrule
$x$                        & Raw input (video, skeleton, or other modality) \\ \hline
$y$                        & Utility label (action class) \\ \hline
$p$                        & Privacy-sensitive attribute(s) \\ \hline
$f_\theta$                 & Privacy transformer / anonymizer / encoder  \\ \hline
$z = f_\theta(x)$          & Transformed representation (privacy-protected output of $f_\theta$) \\ \hline
$g_\phi$                   & Utility predictor / action classifier (parameterized by $\phi$) \\ \hline
$\hat{y} = g_\phi(z)$      & Predicted action from transformed representation \\ \hline
$a_\psi$                   & Privacy adversary / attribute classifier (parameterized by $\psi$) \\ \hline
$\hat{p} = a_\psi(z)$      & Predicted private attribute from transformed representation \\ \hline
$\mathcal{L}_{\text{util}}$ & Utility loss (e.g., cross-entropy for action recognition) \\ \hline
$\mathcal{L}_{\text{priv}}$ & Privacy loss (expected to be minimized) \\ \hline
$\lambda$                  & Trade-off coefficient balancing utility and privacy losses \\ \hline
$\tau$                     & Privacy threshold in the constrained formulation \\ \hline
$\operatorname{Privacy}(\cdot)$ & Privacy functional measuring protection level  \\ \hline
$\text{s.t.}$              & ``Subject to'' (constraint qualifier) \\ \hline
$\min,\ \max$              & Minimization / maximization over model parameters \\ \hline
$\varepsilon, \delta$      & Differential privacy parameters \\ \hline
$\mathcal{A}$              & Randomized mechanism / algorithm \\ \hline
$D,\ D'$                   & Adjacent datasets (differing in one record) \\ \hline
$S$                        & Measurable subset of the output space (event) \\
\bottomrule
\end{tabularx}
\end{table}
The field matured rapidly over 2021--2026, with more publications, more sophisticated methods, and coalescing benchmarks~\cite{zhao_visual_2025}. Even so, it remains fragmented. Papers adopt different privacy definitions such as cMAP, $\varepsilon$-differential privacy, attribute-classification accuracy, and informal claims. They assume different attacker models such as naive classifiers, adaptive attackers, transfer attacks, and membership inference. They evaluate on different benchmarks such as HMDB51, UCF101, and privacy-aware variants, and they report results at varying levels of rigor. This makes it hard to tell which methods work, under what conditions, and how to deploy them responsibly. This survey addresses that fragmentation. We conduct a systematic review of 32 peer-reviewed papers from 2018 to 2026, identified from 885 screened records. We then synthesize them into a unified taxonomy, identify mechanism families and their trade-offs, propose standardized evaluation protocols, and chart a research agenda for field maturation.
\subsection{The Paradigm Shift}
PPAR marks a paradigm shift in video understanding. Rather than treating privacy as an external constraint or a post-hoc annotation, it embeds privacy directly into learning and inference. It optimizes for task utility and leakage suppression at the same time. The field moved from binary obfuscation to learned anonymization, and from informal privacy claims to formal definitions. Early work by Wu et al.\ (2018) and Wang et al.\ (2019) established that privacy-aware sensing and processing need not destroy action discriminability~\cite{wu_towards_2018,wang_privacy-preserving_2019}. CIAGAN (Maximov et al., 2020) marked a turning point by transforming facial identity while preserving action structure~\cite{maximov_ciagan_2020}. Tomei et al.\ (2021) then quantified how face obfuscation affects downstream recognition and established a utility-privacy measurement framework~\cite{tomei_estimating_2021}. Methodological maturation followed. Liu et al.\ (2021) embedded privacy defenses in compressed-sensing pipelines~\cite{liu_video_2021}. SPAct (Dave et al., 2022) introduced self-supervised privacy learning with multi-attribute evaluation~\cite{dave_spact_2022}. STPrivacy (Li et al., 2023) proposed spatio-temporal anonymization with standardized dataset variants such as VP-HMDB51 and VP-UCF101~\cite{li_stprivacy_2023}. MPPAR (Peng et al., 2023) addressed adaptive-attacker generalization through meta-learning~\cite{peng_joint_2023}. Formal guarantees emerged in parallel. Luo et al.\ (2024) applied differential privacy to video action recognition~\cite{luo_differentially_2024}. Modality-aware approaches showed that privacy can also be preserved by design. Gao et al.\ (2025), Moon et al.\ (2023), Jain et al.\ (2024), and Hirose et al.\ (2022) demonstrated that skeleton, depth, and silhouette representations inherently reduce appearance leakage~\cite{gao_privacy-preserving_2025,moon_anonymization_2023,jain_privacy-preserving_2024,hirose_anonymization_2022}. Recent work extends the field beyond RGB recognition to generative restoration and multimodal settings (Feng et al., 2025). It also includes noise-collaborative, encryption-based, and segmentation-driven frameworks~\cite{feng_priva_2025,ismail_stealthguard_2025,li_supplementary_2025,kim_secure_2022,yan_image_2020}. PPAR now spans four concerns. Sensing asks which modalities expose less identity. Encoding compares learned and hand-crafted representations. Learning covers adversarial, self-supervised, and meta-learning methods. Evaluation covers cMAP, differential privacy, and membership inference. Deployment in surveillance, smart homes, healthcare, assisted living, and edge analytics now demands that systems address privacy explicitly~\cite{climent-perez_privacy-preserving_2022,al-obaidi_privacy_2019,al-obaidi_modeling_2020,li_multi-objective_2025,singh_human_2021,perelli_analysis_2024,zhang_chaotic_2025}. PPAR is therefore no longer a subtopic of action recognition. It is a distinct research area that requires its own taxonomy, theory, and standards.
\definecolor{LightGray}{rgb}{0.9,0.9,0.9}
\newcolumntype{C}[1]{>{\centering\arraybackslash}m{#1}}
\newcolumntype{L}[1]{>{\raggedright\arraybackslash}m{#1}}
\newcolumntype{J}[1]{>{\justifying\noindent\arraybackslash}m{#1}}
\DeclareRobustCommand{\covfull}{\tikz[baseline=-0.45ex]{\fill (0,0) circle (0.6ex);\draw[line width=0.3pt] (0,0) circle (0.6ex);}}                                   
\DeclareRobustCommand{\covpart}{\tikz[baseline=-0.45ex]{\fill (0,0) -- (90:0.6ex) arc (90:270:0.6ex) -- cycle;\draw[line width=0.3pt] (0,0) circle (0.6ex);}}      
\DeclareRobustCommand{\covnone}{\tikz[baseline=-0.45ex]{\draw[line width=0.3pt] (0,0) circle (0.6ex);}}                                                             
\sloppy
\begin{table*}[t]
\centering
\caption{Positioning of this survey against the most related privacy and visual-security surveys, along the scope axes that discriminate them. \textbf{PPAR} denotes privacy-preserving action recognition as the primary target. \textbf{Skel.} denotes skeleton/pose-modality privacy. \textbf{Formal} denotes model-space or formal (differential-privacy) guarantees. \textbf{Multi-mech.} denotes breadth across the privacy-injection mechanism families such as input/acquisition, representation, model, compression/encryption, and output/deployment. Citation counts are from Google Scholar, July 2026. \covfull~denotes fully covered, \covpart~partially covered, and \covnone~not covered.}
\label{tab:survey_comparison}
\renewcommand{\arraystretch}{1.4}
\renewcommand\tabularxcolumn[1]{m{#1}}
\footnotesize
\begin{tabularx}{\textwidth}{@{}
        C{0.4cm}   
        >{\raggedright\arraybackslash}p{1.8cm}   
        C{0.5cm}   
        C{0.6cm}   
        >{\raggedright\arraybackslash}p{1.7cm}   
        C{0.55cm}  
        C{0.55cm}  
        C{0.55cm}  
        C{1.75cm}   
        >{\arraybackslash}X 
@{}}
\toprule
\multirow{2}{*}{\textbf{\#}} &
\multirow{2}{*}{\raggedright\arraybackslash\textbf{Survey}} &
\multirow{2}{*}{\textbf{Year}} &
\multirow{2}{*}{\textbf{Cit.}} &
\multirow{2}{*}{\textbf{Venue}} &
\multicolumn{4}{c}{\textbf{Focus}} &
\multirow{2}{*}{\raggedright\arraybackslash\textbf{Description}} \\[-0.2em]
\cmidrule(lr){6-9}
& & & & & \textbf{PPAR} & \textbf{Skel.} & \textbf{Formal} & \textbf{Multi-mech.} & \\
\midrule
1 & GAN privacy survey~\cite{cai_generative_2021}          & 2021 & 533 & ACM Comput. Surv.     & \covnone & \covnone & \covpart  & \covpart & Surveys GAN-based privacy and security applications across vision, language, and other domains. It takes a generative-model view and is not action-centric. \\ \hline
2 & Surveillance paradox~\cite{cucchiara_video_2024}       & 2024 & 6   & IEEE Computer         & \covpart & \covnone & \covnone & \covnone & Reports AI experiments such as action recognition and NL description that reconcile surveillance and privacy. It offers no method taxonomy. \\ \hline
3 & Visual content privacy~\cite{zhao_visual_2025}         & 2025 & 77  & ACM Comput. Surv.     & \covnone & \covnone & \covnone & \covpart & Surveys visual-content privacy through a CV-/HV-adversary framework that categorizes protection methods. It is not action-recognition focused. \\ \hline
4 & Elderly monitoring~\cite{houshidari_three_2025} & 2025 & 1   & IEEE Smart Grid Conf. & \covnone & \covnone & \covnone & \covnone & Bibliometric review of 4,864 papers from 1990 to 2024 on noninvasive older-adult monitoring. It maps privacy trends rather than methods. \\ \hline
5 & PPAR survey~\cite{li2024privacy}          & 2024 & 1   & PRCV                  & \covfull & \covpart & \covnone & \covpart & Concise survey of PPAR methods and categories. It is narrow in scope, with no systematic protocol or empirical trade-off. \\ \hline
6 & Skeleton VR privacy~\cite{carr_review_2024}            & 2024 & 1   & IEEE MetaCom          & \covpart & \covfull & \covpart & \covnone & Reviews the privacy-utility of skeleton and pose data in VR/metaverse settings. It covers the skeleton modality only, with no broader taxonomy. \\ \hline
\textbf{7} & \textbf{Our survey} & \textbf{2026} & \textbf{--} & \textbf{--} & \covfull & \covfull & \covfull & \covfull & Systematic PRISMA review of PPAR across all injection stages and modalities, with a unified taxonomy, standardized protocol, and empirical trade-offs. \\
\bottomrule
\end{tabularx}
\end{table*}
\subsection{Problem Definition and Scope}
We define PPAR as follows. The input is a spatiotemporal observation, such as RGB video, a skeleton sequence, depth frames, or a compressed-domain representation. The output is an action label or a task-specific prediction, such as an activity class or an anomaly score. The privacy constraint is to limit the leakage of sensitive attributes, including identity, demographics, appearance, behavior, and location. The goal is to keep utility high while keeping an attacker's attribute or identity inference low. We measure utility by recognition accuracy or detection rate. This definition separates PPAR from related problems. Generic action recognition ignores privacy. Face anonymization protects only the face, whereas PPAR covers full-body action, gait, and behavior. Differential privacy for tabular data must here be adapted to high-dimensional, correlated video. Encryption protects data in transit, whereas PPAR protects semantics during learning, which makes a violation harder regardless of policy. PPAR is therefore a multi-objective learning problem. Its utility and privacy goals are task- and attacker-dependent, and they often conflict. Four trade-offs govern this space. The first is utility against privacy. The second is generalization against attacker-specific defenses. The third is formal guarantees against practical empirical metrics such as cMAP. The fourth is reconstruction resistance against fine-grained discriminability. Representative methods span this space. Adversarial and self-supervised approaches trade roughly 10--20\% accuracy for large privacy gains~\cite{dave_spact_2022,li_stprivacy_2023,wu_privacy-preserving_2022}. Differential privacy attains formal guarantees only at a steep utility cost~\cite{luo_differentially_2024}. Modality-aware methods decouple privacy and utility by design~\cite{moon_anonymization_2023,gao_privacy-preserving_2025}. We scope the survey to video, skeleton, depth, and compressed-domain recognition with explicit, empirically evaluated privacy mechanisms. We exclude non-video modalities such as WiFi, RF, and mmWave. We also exclude generic privacy tools not tailored to visual action tasks, unevaluated privacy claims, and face anonymization detached from action-recognition utility. We unify these fragmented variants under one framework, formalized in Section~\ref{sec:2}. A system accepts spatiotemporal input, applies a privacy-aware transformation such as representation learning, encryption, or modality selection, and outputs action predictions subject to privacy constraints defined against explicit attacker models.
\subsection{Comparison with Existing Surveys}
Several broad surveys touch on privacy in visual computing. Cai et al.\ review GANs and their privacy applications~\cite{cai_generative_2021}. Cucchiara et al.\ cover surveillance and privacy from both policy and technical angles~\cite{cucchiara_video_2024}. Zhao et al.\ survey privacy-preserving visual analytics more generally~\cite{zhao_visual_2025}. Others treat GAN taxonomies for vision~\cite{wang_generative_2021}, adversarial robustness of 2D/3D models~\cite{li_survey_2024}, privacy attacks and defenses in federated learning~\cite{zhao_federation_2025}, generative facial de-identification~\cite{park_privacy-driven_2025}, and three decades of privacy-preserving older-adult monitoring~\cite{houshidari_three_2025}. Two concurrent reviews come closest to our topic. Li et al.~\cite{li2024privacy} focus on privacy-preserving action recognition, and Carr et al.~\cite{carr_review_2024} focus on skeleton-data privacy in VR metaverses.
None of these gives the PPAR-specific, evaluation-centered synthesis the field now needs. They fall short in six recurring ways. First, their scope is broad. General privacy surveys fold face anonymization, activity monitoring, identity obfuscation, and crowd analysis into one discussion, without isolating action recognition under privacy. Second, they are not evaluation-centric. They enumerate methods without comparing utility metrics, privacy metrics, and attacker assumptions. They also do not confront how inconsistently papers report cMAP, $\varepsilon$ values, or informal privacy claims. Third, they offer no unified taxonomy. PPAR spans input-space methods such as coded aperture and blurring, representation-space methods such as adversarial and self-supervised learning, model-space methods such as architecture design and federated learning, and output-space interventions. Prior surveys treat these as separate topics rather than one taxonomy organized by injection level and mechanism family. Fourth, they conflate empirical attribute hiding with formal privacy. Empirical attribute hiding means a classifier cannot recover the attribute from protected features. Formal privacy means differential-privacy bounds or information-theoretic guarantees. Fifth, they leave benchmark fragmentation unaddressed. Papers draw on PA-HMDB51, HMDB51, UCF101, Kinetics, and custom variants with mismatched splits, attacker protocols, and metrics. Sixth, they stay descriptive. They summarize past work without prescribing how results should be evaluated and reported. Table~\ref{tab:survey_comparison} positions our survey against the six most closely related reviews, published between 2021 and 2025, along the four scope axes that discriminate them. These axes are PPAR as the primary target, skeleton/pose-modality privacy, formal guarantees at the model-space or differential-privacy level, and breadth across the privacy-injection mechanism families. The general privacy and GAN surveys~\cite{cai_generative_2021,zhao_visual_2025} reach several mechanism families but cover neither PPAR nor skeleton privacy. They catalog protection methods across many visual tasks, GANs in one case and a CV-/HV-adversary framework in the other, without isolating action recognition or testing privacy claims under comparable protocols. The surveillance- and domain-specific reviews~\cite{cucchiara_video_2024,houshidari_three_2025} cover the least of the space. The former only partially targets action recognition and reports AI experiments that reconcile surveillance with privacy, but it offers no method taxonomy. The latter is a bibliometric review of noninvasive older-adult monitoring, so it maps privacy trends rather than mechanisms and covers none of the four axes. Even the closest concurrent reviews~\cite{li2024privacy,carr_review_2024} span only part of the space. Li et al.\ target PPAR directly, yet they treat skeleton privacy and mechanism breadth only partially and formal guarantees not at all. Carr et al.\ cover skeleton privacy fully, but they cover PPAR and formal guarantees only partially and offer no broader mechanism taxonomy. Both also stay method-descriptive. Neither assembles a systematic corpus, prescribes a standardized evaluation protocol, or measures privacy--utility trade-offs empirically. Ours is the only entry with full coverage on all four axes. No prior survey treats PPAR as a standalone problem with its own threat models, taxonomy, and evaluation standards, and that is the gap this survey fills.
\begin{figure*}[t]
\centering
\includegraphics[width=\textwidth]{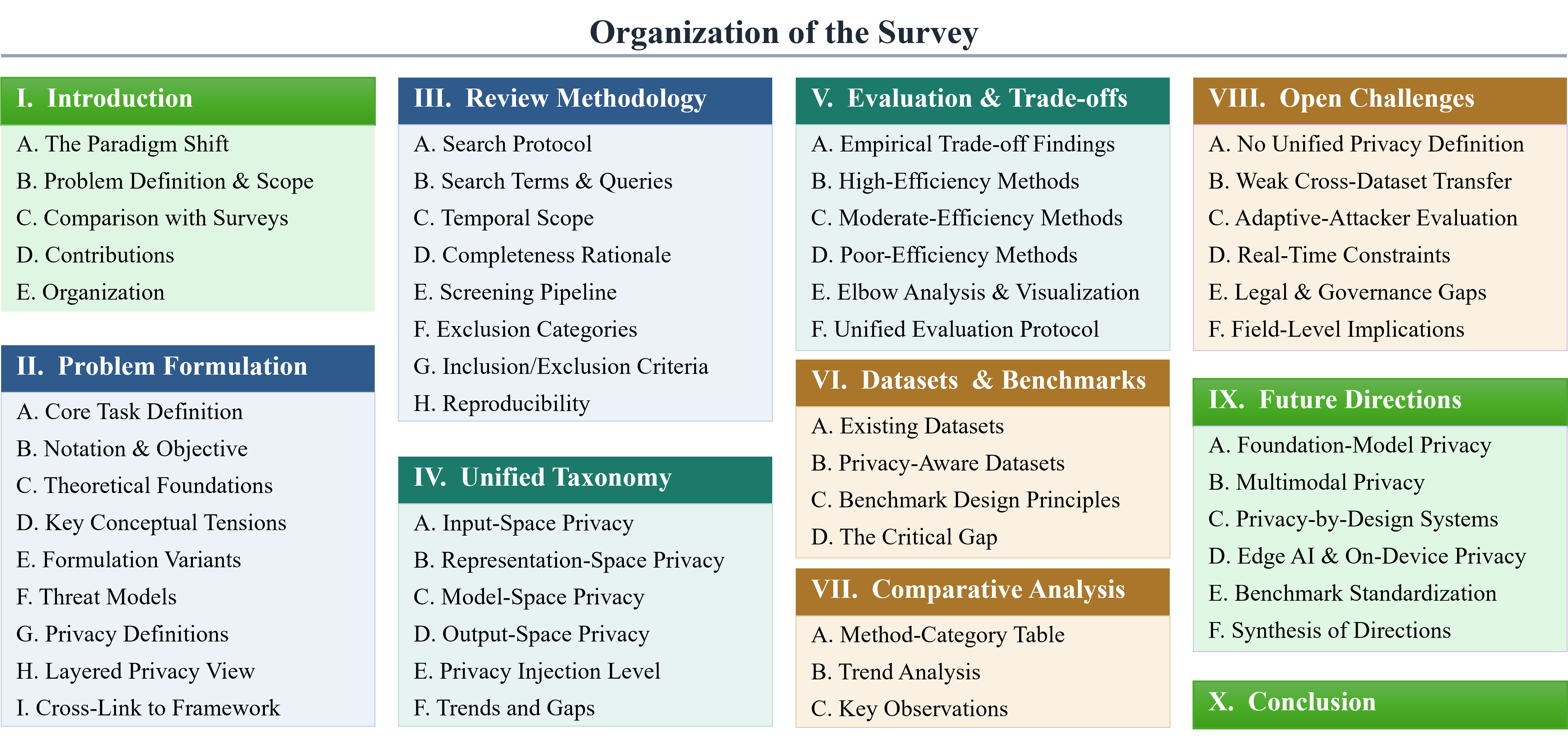}
\caption{Section-level organization of this survey, from problem formulation and taxonomy through evaluation, benchmarks, and open challenges to future directions.}
\label{fig:survey_organization}
\end{figure*}
\subsection{Contributions}
This survey is distinguished on every axis. It is the first systematic (PRISMA) review dedicated to PPAR, and it screens 885 records down to a 32-paper corpus. It organizes methods under a unified two-dimensional privacy-space taxonomy of injection stage and mechanism family, rather than the single method-type axis of prior PPAR surveys. It is the only work to propose a standardized evaluation protocol for reproducible privacy claims. It is the only one to characterize privacy-utility trade-offs empirically through Pareto frontiers and practical operating points. It is also the only one to chart a phased research roadmap toward deployment. Together these establish the present work as the evaluation-centered reference the field currently lacks. The main contributions of this work to the PPAR research community are as follows.
\begin{enumerate}
  \item A unified taxonomy of PPAR methods organized by privacy injection level (input-, representation-, model-, output-space) and mechanism family (adversarial learning, skeleton-based, cryptographic, differential privacy, and emerging hybrid methods), enabling readers to locate and position new methods relative to existing work.
  \item Formalization of the PPAR problem, with notation for input/output, privacy attributes, attacker models (white-box/black-box, adaptive/transfer), and privacy metrics (cMAP, $\varepsilon$-DP, membership inference), enabling rigorous comparison across papers.
  \item A critical comparative analysis of the corpus across utility metrics (Top-1 accuracy, per-class breakdown, degradation \%), privacy metrics (cMAP, $\varepsilon$-DP, attribute accuracy), attacker assumptions (threat model, knowledge, training data), and empirical trade-offs with Pareto-frontier analysis, identifying high-efficiency methods.
  \item A proposed PPAR Unified Evaluation Protocol that specifies minimum requirements for utility reporting (Top-1 and degradation \%), privacy evaluation (a primary metric of cMAP or $\varepsilon$-DP, and a secondary metric of membership inference or transfer), attack specification (an explicit threat model), reproducibility (code, splits, hyperparameters), and baseline comparisons.
  \item Identification of open challenges and future directions, including the absence of a universal PPAR benchmark suite, formal privacy-utility trade-off characterization, generalization and robustness under adaptive attackers, privacy-preserving edge deployment under real-time constraints, extension to multimodal and generative settings, and privacy-aware data collection and annotation.
\end{enumerate}
\subsection{Organization}
The rest of this survey is organized as follows, and Fig.~\ref{fig:survey_organization} gives a visual overview of the section structure. Section~2 formalizes the PPAR problem, threat models, and privacy definitions. Section~3 details the systematic review methodology, including the search and screening protocol. Section~4 presents our unified taxonomy of PPAR methods across input-, representation-, model-, and output-space privacy. Section~5 analyzes evaluation protocols and privacy-utility trade-offs, and it introduces the PPAR Unified Evaluation Protocol with its five standardized components. Section~6 reviews datasets and benchmarks. Section~7 provides a quantitative comparative analysis across mechanism families. Section~8 identifies open challenges. Section~9 outlines future directions. Section~10 concludes.

\section{Problem Formulation}\label{sec:2}
PPAR is not a single problem with a universal mathematical formulation. It is a family of related problems that vary along four axes, namely sensing modality, privacy target, threat assumptions, and privacy definition. The literature often mixes these variants without making the differences explicit. This creates confusion about which methods solve which problems and whether their claims are comparable. This section therefore establishes a unified notation and theoretical framework. We define the core PPAR problem, characterize the threat models, and formalize the privacy notions used across the corpus. This gives the common language on which the taxonomy, evaluation, and comparative analysis of later sections depend.
\subsection{Core Task Definition}
\textbf{Input.} A spatiotemporal observation such as RGB video, skeleton sequences, depth streams, compressed-domain representations, or other structured spatiotemporal modalities. \textbf{Utility Task.} Given $x$, predict an action label $y \in \mathcal{Y}$, where $\mathcal{Y} = \{1, 2, \ldots, K\}$ is a discrete set of $K$ action classes, such as \textit{walking} or \textit{running} for recognition. \textbf{Privacy Target.} Sensitive attributes such as identity, demographics, appearance, and location or context. \textbf{Goal.} Learn a typically multi-stage function that maximizes utility while minimizing privacy leakage.
 
\subsection{Notation and the Joint Optimization Objective}
Following the notation adopted throughout this survey, the general PPAR objective is a constrained multi-objective problem.
\begin{equation}
    \min_{\theta, \phi}\ \mathcal{L}_{\text{util}}\big(g_\phi(f_\theta(x)),\, y\big) \quad \text{s.t.}\quad \operatorname{Priv}\!\big(a_\psi, f_\theta(x)\big) \ge \tau .
    \label{eq:constrained}
\end{equation}
It also has an unconstrained form with trade-off parameter $\lambda > 0$.
\begin{equation}
    \min_{\theta,\phi}\ \mathcal{L}_{\text{util}}\big(g_\phi(f_\theta(x)),\, y\big)
    + \lambda\, \mathcal{L}_{\text{priv}}\big(a_\psi(f_\theta(x)),\, p\big).
    \label{eq:weighted}
\end{equation}
Here $\mathcal{L}_{\text{priv}}$ rewards making attribute prediction harder. Trained jointly or sequentially, $a_\psi$ maximizes attribute prediction while $g_\phi$ preserves action accuracy. This creates a minimax game, and its adversarial form is made explicit in Eq.~\eqref{eq:minimax}.
\subsection{Theoretical Foundations Used in the Corpus}
The PPAR literature employs six complementary theoretical frameworks.
\subsubsection{Adversarial Representation Learning}
The utility network and the privacy adversary are trained in opposition. The network learns representations that support action recognition while fooling the adversary. Wu et al. pioneered this approach for visual privacy~\cite{wu_towards_2018}. They later validated it empirically and reduced privacy leakage (cMAP) by 60--70\% while maintaining reasonable action accuracy~\cite{wu_privacy-preserving_2022}. The optimization typically takes the following form.
\begin{equation}
    \min_{\theta,\phi}\ \mathcal{L}_{\text{util}}
    + \lambda \max_{\psi}\ \mathcal{L}_{\text{priv}} .
    \label{eq:minimax}
\end{equation}
\subsubsection{Self-Supervised and Weakly Supervised Privacy Learning}
Self-supervised and weakly supervised methods learn privacy-aware representations from unlabeled data or weak signals, which avoids dense attribute labels. SPAct~\cite{dave_spact_2022} uses contrastive learning to group same-action clips while separating clips of the same person performing different actions. This encodes action while suppressing identity. STPrivacy~\cite{li_stprivacy_2023} applies spatio-temporal anonymization with self-supervised objectives at the frame and video levels.
\subsubsection{Meta-Learning and Generalization-Aware Privacy}
MPPAR~\cite{peng_joint_2023} applies meta-learning so that privacy-preserving representations generalize across attacker models and unseen attributes. It trains the privacy mechanism against multiple attackers at once, which improves robustness to the worst-case adversary rather than to a single fixed one.
\subsubsection{Structured Modality-Based Privacy}
Skeleton, depth, and motion-based representations reduce appearance leakage by design. Privacy therefore follows from the choice of representation rather than from adversarial learning or encryption. Gao et al.~\cite{gao_privacy-preserving_2025} show that skeleton-based recognition with graph convolutional networks preserves $>$95\% action accuracy while substantially reducing identity inference. Moon et al.~\cite{moon_anonymization_2023} and Jain et al.~\cite{jain_privacy-preserving_2024} report similar trade-offs for skeleton and depth.
\subsubsection{Formal Privacy and Differential Privacy}
Luo et al.~\cite{luo_differentially_2024} apply differential privacy to action recognition by adding calibrated Gaussian noise during training (DP-SGD). The resulting model satisfies $(\varepsilon, \delta)$-differential privacy, so an adversary observing it cannot reliably infer whether any individual's data was in the training set.
\subsubsection{Secure and Reversible Privacy}
Encrypted and reversible pipelines protect privacy by design. Feng et al.~\cite{feng_priva_2025} introduce Priva, a reversible system that performs analysis in an encrypted representation space and restores the original video post-analysis using diffusion models.
Across these frameworks, privacy is obtained either empirically or formally. Empirical methods include adversarial, self-supervised, and modality-based approaches, whereas formal methods include differential privacy and encryption. The formal guarantees are stronger, but they typically incur greater utility or computational cost. The survey examines this trade-off throughout.
\subsection{Key Conceptual Tensions}
PPAR is shaped by four unavoidable tensions. The first is the \emph{privacy-utility trade-off}. Protecting identity and demographic cues often requires removing fine-grained, action-relevant detail, so this is an information-theoretic constraint rather than a merely computational one. The second is \emph{generalization versus specificity}. A defense tuned to a specific attacker or dataset may fail against different attackers or cross-domain inputs. This forces a balance between strong privacy on known threats and robustness to unforeseen ones. The third is \emph{empirical versus formal privacy}. Empirical privacy is flexible but attacker-dependent and offers weak guarantees, whereas formal privacy (DP) provides provable bounds at often prohibitive utility cost. No method has yet achieved both at once. The fourth is \emph{computation and latency}. Strong mechanisms such as secure multiparty computation and carefully calibrated DP-SGD impose overhead that hinders real-time deployment on edge devices.
\subsection{Formulation Variants Across the Literature}
The literature varies systematically along three dimensions. The first is the \emph{privacy objective}. Some papers learn privacy adversarially (Wu et al., SPAct, STPrivacy, MPPAR)~\cite{wu_towards_2018,wu_privacy-preserving_2022,dave_spact_2022,li_stprivacy_2023,peng_joint_2023}, others learn it by modality choice such as skeleton or depth, and others rely on differential privacy or encryption. Each implies different optimization landscapes and convergence properties. The second is \emph{privacy measurement}. Empirical privacy is measured through attribute-classifier accuracy, where cMAP dominates, formal privacy through $\varepsilon$-DP bounds, and anonymization through reconstruction or re-identification attacks. These metrics are not directly comparable. The third is the \emph{pipeline stage}. Privacy can be injected at the input (obfuscation, modality substitution), the representation (adversarial learning, DP noise), the model (training-time DP), or the output (label sanitization, visualization obfuscation). Early injection prevents leakage before processing, whereas late injection preserves training-time utility but risks intermediate leakage. Later sections classify methods along these dimensions to enable systematic comparison and trend analysis.
\begin{table*}[t]
\centering
\caption{Attacker capability dimensions and their implications.}
\label{tab:attack_dimensions}
\renewcommand{\arraystretch}{1.4}
\begin{tabularx}{\textwidth}{@{}
        C{0.4cm}   
        L{2.5cm}     
        L{5cm}     
        >{\raggedright\arraybackslash}X   
@{}}
\toprule
\textbf{\#} & \textbf{Dimension} & \textbf{Options} & \textbf{Implication} \\
\midrule
1 & Access          & White-box vs.\ black-box              & White-box enables gradient attacks. Black-box requires query-based attacks \\ \hline
2 & Training data   & Supervised vs.\ unsupervised          & Supervised is stronger. Unsupervised requires fewer assumptions \\ \hline
3 & Scope           & Single- vs.\ multi-attribute          & Multi-attribute is more realistic \\ \hline
4 & Generalization  & Within- vs.\ cross-domain             & Cross-domain tests robustness \\ \hline
5 & Temporal        & Frame-level vs.\ temporal aggregation & Temporal attacks leverage consistency \\
\bottomrule
\end{tabularx}
\end{table*}
\subsection{Threat Models}
A privacy claim means little without a stated adversary. A transform that defeats a naive classifier can still fail against an adaptive attacker that knows the defense. Attacker strength also sets how a metric is read. A strong attacker gives an upper bound on privacy and a weak one gives a lower bound. We formalize the attacker models used or implied across the corpus and group them by objective and by capability. Table~\ref{tab:attack_dimensions} summarizes the capability dimensions along which these attackers vary.
\emph{Attribute inference} is the dominant model in PPAR. The adversary $a_\psi$ receives a transformed input $z = f_\theta(x)$ and minimizes $\mathcal{L}_{\text{attr}}(a_\psi(z), p)$ for a ground-truth attribute $p$. Success is measured by classification accuracy or cMAP~\cite{wu_privacy-preserving_2022,dave_spact_2022,li_stprivacy_2023,peng_joint_2023,fioresi_ted-spad_2023,ilic_selective_2024}. \emph{Identity re-identification} matches an anonymized video $z$ against a gallery by similarity score. It reports the rank-1 rate or CMC curves~\cite{maximov_ciagan_2020,moon_anonymization_2023,hirose_anonymization_2022,hellmann_ganonymization_2024}. \emph{Reconstruction} applies to compressed-sensing, encryption, and reversible schemes. It trains an inversion model $\mathrm{Inv}(z) \approx x$ to recover the input, scored by reconstruction error (PSNR, SSIM) or semantic recovery, such as whether a restored face can be identified~\cite{wang_privacy-preserving_2019,ishikawa_learnable_2024,zheng_fast_2024,feng_priva_2025}. \emph{Membership inference} asks whether a query sample was in the training set, scored by the true-positive rate at a fixed false-positive rate or by ROC-AUC. This model is understudied in PPAR, though related re-identification risks are noted by Dou et al.~\cite{dou_person_2026}.
Two models capture stronger capabilities. The \emph{adaptive} attacker is the realistic upper bound. It holds white-box knowledge of $f_\theta$ and $g_\phi$ and computes gradients to mount optimization attacks. STPrivacy and MPPAR test this setting through retraining~\cite{li_stprivacy_2023,peng_joint_2023}. The \emph{transfer} attacker trains on a source domain and is evaluated on a target domain, which measures how attack success survives domain shift. STPrivacy shows this on VP-HMDB51 and VP-UCF101~\cite{li_stprivacy_2023}. Most PPAR papers assume a limited attacker with black-box access, single-attribute prediction, and within-domain evaluation. This is the easiest setting to defend. Stronger models such as adaptive, transfer, and multi-attribute attackers are tested by only a few papers, chiefly STPrivacy and MPPAR. A privacy claim is therefore only as strong as its threat model, so comparisons should state it explicitly and note which method faces the stronger adversary.
\subsection{Privacy Definitions}
\subsubsection{Empirical Privacy}
Empirical privacy defines a representation as private if an attacker trained to predict sensitive attributes achieves lower-than-baseline accuracy. It is measured with cMAP (confusion-matrix average precision) or attribute-classification accuracy. An unprotected baseline scores cMAP $\approx$ 0.9, and protected representations target $<$ 0.4 for moderate privacy. Its central limitation is that it is task- and attacker-dependent and offers no formal guarantee against unmeasured attack vectors, so a different classifier or training procedure may yield a different cMAP. It is nonetheless the dominant metric in the PPAR literature (Wu et al.~\cite{wu_privacy-preserving_2022}, SPAct~\cite{dave_spact_2022}, STPrivacy~\cite{li_stprivacy_2023}, MPPAR~\cite{peng_joint_2023}).
\subsubsection{Formal Privacy (Differential Privacy)}
Formal privacy provides a mathematical guarantee. A randomized algorithm or model satisfies $(\varepsilon, \delta)$-differential privacy if, for any two adjacent datasets $D$ and $D'$ that differ in one record,
\begin{equation}
    \Pr[\mathcal{A}(D) = S] \le e^{\varepsilon}\, \Pr[\mathcal{A}(D') = S] + \delta .
    \label{eq:dp}
\end{equation}
where $\mathcal{A}$ is the algorithm and $S$ is any outcome set. Here $\varepsilon$ quantifies privacy loss, where a lower $\varepsilon$ means stronger privacy, and $\delta$ is a failure probability that is typically $10^{-5}$. Its advantages are provable, model-independent guarantees and composition theorems for combining multiple mechanisms. Its drawback is that strong privacy demands a small $\varepsilon$ and therefore sacrifices utility, so it remains impractical at scale for video action recognition. Luo et al.~\cite{luo_differentially_2024} illustrate the cost, since $\varepsilon = 5$ achieves 76\% accuracy while $\varepsilon = 1$ drops to $\sim$45\%.
\subsubsection{Anonymization / De-identification Privacy}
Anonymization, or de-identification, achieves privacy by removing or transforming identifying information so that identity and related demographic attributes cannot be recovered, even visually. Typical mechanisms include face blurring or pixelation~\cite{tomei_estimating_2021}, face swapping and anonymization (CIAGAN, Hellmann et al.), skeleton-only representation~\cite{gao_privacy-preserving_2025,moon_anonymization_2023}, and silhouette deformation~\cite{hirose_anonymization_2022}. These are evaluated through visual inspection, reconstruction, and re-identification attacks, with metrics such as reconstruction PSNR/SSIM and rank-1 re-identification accuracy. The approach is interpretable and easily deployable with simple image processing. It is also ad hoc, potentially invertible under strong attacks, and without formal guarantees. Representative works include Maximov et al.~\cite{maximov_ciagan_2020}, Moon et al.~\cite{moon_anonymization_2023}, Hirose et al.~\cite{hirose_anonymization_2022}, and Hellmann et al.~\cite{hellmann_ganonymization_2024}.
\subsubsection{Cryptographic / Secure Privacy}
Cryptographic, or secure, privacy keeps data encrypted during transmission and processing, so that only authorized parties can decrypt. Mechanisms include coded aperture~\cite{wang_privacy-preserving_2019}, learnable encryption~\cite{ishikawa_learnable_2024}, encrypted inference~\cite{feng_priva_2025}, and homomorphic encryption. These are evaluated through reconstruction attacks on encrypted data, decryption attempts, and information leakage under model-extraction attacks. The approach offers strong formal confidentiality guarantees and is hardware-deployable. It is also computationally expensive, adds latency, and introduces key-management complexity. Representative works include Wang et al.~\cite{wang_privacy-preserving_2019}, Ishikawa et al.~\cite{ishikawa_learnable_2024}, Zheng et al.~\cite{zheng_fast_2024}, and Feng et al.~\cite{feng_priva_2025}.
\subsubsection{Task-Conditioned Privacy}
Task-conditioned privacy is defined relative to the target task. Data are private if they prevent inference of sensitive attributes while still enabling action recognition. This prevailing PPAR interpretation treats privacy as task-specific rather than absolute, so identity may be unrecoverable while action remains recognizable. It is evaluated by jointly measuring utility (action accuracy) and privacy (attribute accuracy) and reporting the resulting trade-off. The approach is practical and suits deployments where perfect privacy is unnecessary as long as task utility is maintained. Its guarantee depends on the task definition and is not universally transferable. Most PPAR papers implicitly adopt this definition (Wu et al., SPAct, STPrivacy, MPPAR, Gao et al., Jain et al.)~\cite{wu_privacy-preserving_2022,dave_spact_2022,li_stprivacy_2023,peng_joint_2023,gao_privacy-preserving_2025,jain_privacy-preserving_2024}.
\subsubsection{Critical Distinctions}
Four distinctions are essential to interpreting privacy claims correctly. First, a privacy \emph{guarantee}, \emph{metric}, and \emph{threat} are not the same thing. A guarantee is a formal mathematical bound (DP), a metric is an empirical measurement such as cMAP or reconstruction error, and a threat is an attack model. A method may therefore satisfy a guarantee yet still fail empirically against unmeasured attacks. Second, cMAP is a metric rather than a guarantee, so a low cMAP against one classifier does not guarantee privacy against different classifiers or modalities. Third, DP is a guarantee rather than merely a metric, since $\varepsilon$-DP formally bounds information leakage for any attacker, although practical $\varepsilon$ values often make utility infeasible. Fourth, anonymization claims require explicit evaluation, because removing faces or adopting a skeleton representation does not by itself guarantee privacy, which must be validated under explicit re-identification or attribute attacks.
\subsection{Layered Privacy Interpretation}
For the remainder of the survey we group the preceding definitions into three interpretive layers. Empirical privacy is task-specific attribute hiding, measured through cMAP-style metrics. Formal privacy is differential-privacy bounds on information leakage. Task-conditioned privacy is utility-aware privacy relative to the target task. Distinguishing these layers clarifies what privacy each method actually claims and whether competing claims are comparable at all.
\subsection{Cross-Linking to the Survey Framework}
This formalization directly shapes the rest of the survey. It determines how methods are grouped in the taxonomy, by privacy injection point and mechanism family, and which methods can be fairly compared, namely those that share threat models and privacy definitions. It also frames the open challenges as direct consequences of these definitions, including the scarcity of membership-inference evaluation, weak transfer-attack testing, and formal privacy at practical utility.

\begin{table}[t]
\centering
\caption{Search domains used in query construction.}
\label{tab:searchterms}
\renewcommand{\arraystretch}{1.25}
\renewcommand\tabularxcolumn[1]{m{#1}}
\begin{tabularx}{\columnwidth}{@{}
        C{0.45cm}
        L{3.0cm}
        >{\arraybackslash}X @{}}
\toprule
\textbf{\#} & \textbf{Domain} & \textbf{Queries} \\
\midrule
1 & Action Recognition Core &
Action recognition, Activity recognition, Human action, Activity analysis, Action understanding, Temporal action \\
\midrule
2 & Privacy and Defense &
Privacy-preserving, Privacy-aware, Anonymization, De-identification, Privacy protection, Privacy mechanism \\
\midrule
3 & Visual Privacy Specific &
Visual privacy, Video privacy, Appearance protection, Identity privacy, Face privacy, Attribute privacy \\
\midrule
4 & Video/Skeleton Specific &
Skeleton-based, Depth-based, Pose-based, Keypoint, Compressed video, Coded aperture \\
\midrule
5 & Learning-Based Defense &
Adversarial learning, Representation learning, Differential privacy, Secure learning, Encrypted video \\
\midrule
6 & Applications &
Surveillance privacy, Smart home, Healthcare monitoring, Assisted living, Edge analytics \\
\bottomrule
\end{tabularx}
\end{table}

\section{Systematic Review Methodology}
The PPAR literature is fragmented across the action recognition, computer vision, privacy and security, anonymization, and surveillance communities. These communities adopt divergent terminology, experimental settings, and privacy definitions. A systematic review protocol is therefore essential to minimize selection bias and maximize reproducibility. We searched peer-reviewed research on privacy-preserving mechanisms for video- and skeleton-based action recognition, and we combined database queries with manual supplementation across vision, security, and multimedia venues. The resulting 885 records were reduced through two screening stages, title/abstract screening with 99 retained and full-text screening with 32 finally included. Only papers that meet rigorous, auditable criteria inform the taxonomy, comparative analysis, and evaluation protocol that follow.
\subsection{Search Protocol}
We searched four databases (Web of Science, Scopus, IEEE Xplore, and ACM Digital Library) together with targeted hand-searches of recent proceedings (CVPR, ICCV, ECCV, NeurIPS, ICML, CCS, USENIX Security) and arXiv preprints. This multi-source strategy captured recent, high-impact work across disciplinary boundaries.
\subsection{Search Terms and Query Design}
We combined domain terms with Boolean operators (AND, OR) plus venue and time filters, and we refined the query on pilot searches to balance recall and precision. The query spanned six semantic domains, summarized in Table \ref{tab:searchterms}. For example, terms from the Action Recognition Core and Privacy and Defense clusters were joined as \texttt{("action recognition" OR "activity recognition") AND ("privacy-preserving" OR "anonymization")}, and then intersected with the remaining domains to widen coverage.
\subsection{Temporal Scope and Justification}
The search was restricted to 2018 onward. This is the period in which the techniques underpinning modern PPAR (deep networks, adversarial training, self-supervised representation learning, and differential privacy) matured and saw widespread adoption. Wu et al. pioneered explicit privacy-aware visual recognition~\cite{wu_towards_2018}, and Wang et al. introduced coded-aperture acquisition for privacy~\cite{wang_privacy-preserving_2019}. These works mark the boundary of this era.
\subsection{Search Completeness Rationale}
The query deliberately spanned terminological boundaries. It captured papers that address core PPAR problems without using the phrase "privacy-preserving action recognition", for example work framed as "video" or "skeleton anonymization." These cross-terms surfaced papers a narrow query would miss, including CIAGAN~\cite{maximov_ciagan_2020} ("face anonymization" plus "action"), Tomei et al.~\cite{tomei_estimating_2021} ("face obfuscation" plus "recognition"), skeleton anonymisation~\cite{moon_anonymization_2023} ("skeleton" plus "privacy"), compressed-domain privacy~\cite{zheng_fast_2024} ("encrypted video" plus "action recognition"), and differential privacy~\cite{luo_differentially_2024} ("differential privacy" plus "action recognition").
\subsection{Screening Pipeline}
\begin{figure}[t!]
\centering
\includegraphics[width=1\columnwidth]{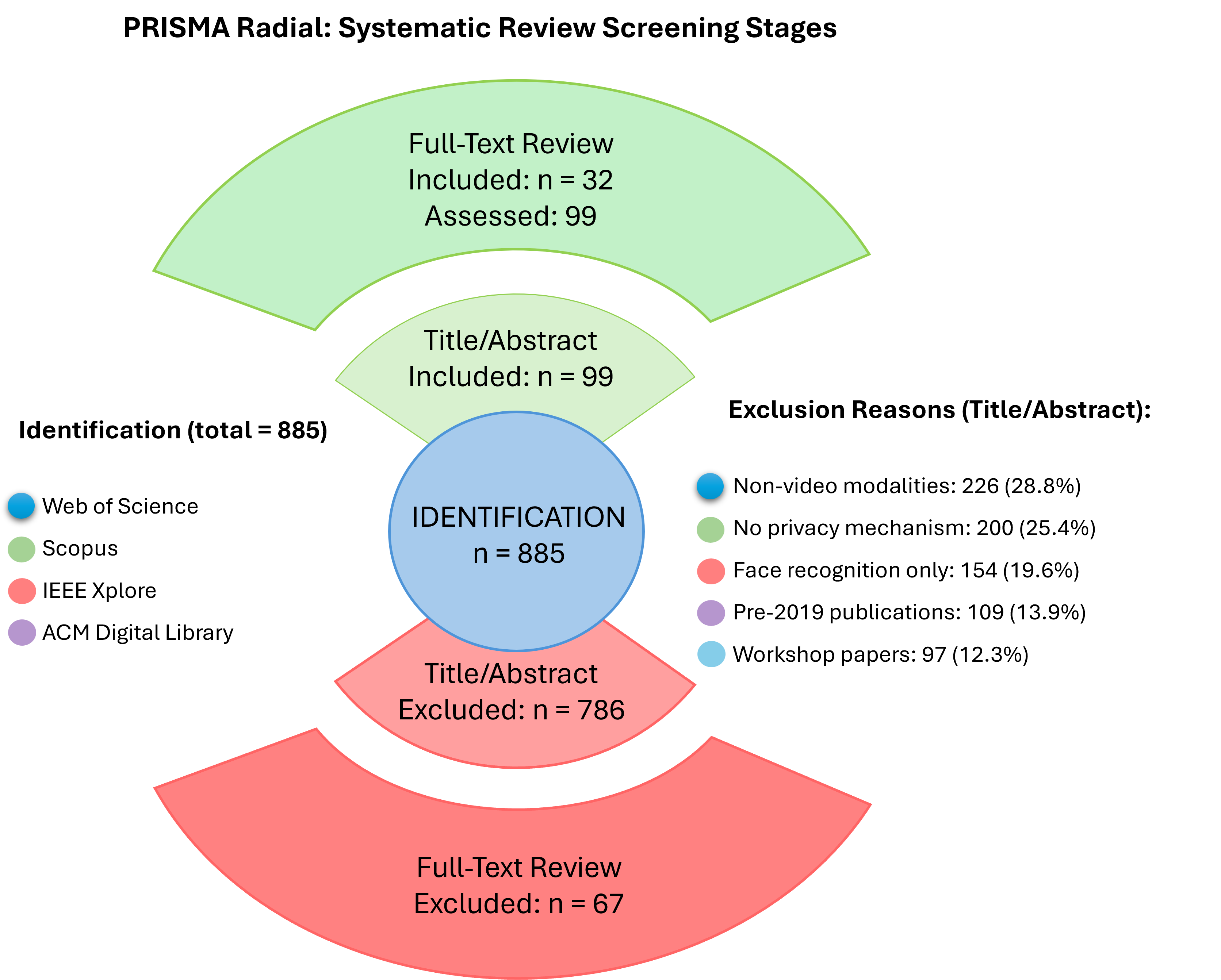}
\caption{PRISMA flow of the study-selection process, from 885 identified records to the 32 papers included in this review.}
\label{fig:prisma_flow}
\end{figure}

We reduced the 885 records in two stages. The process follows the PRISMA flow of Fig.~\ref{fig:prisma_flow} and applies the inclusion and exclusion criteria defined in Table \ref{tab:searchterms} and Table \ref{tab:inclusion_criteria}.
\subsubsection{Title and Abstract Screening (Stage 1)}
At the title and abstract level we kept records that met four conditions. A record had to concern video-based action or activity recognition. It had to mention privacy or anonymization. It had to appear in 2018 or later. Finally, it had to come from a peer-reviewed or otherwise substantive venue. We removed records on non-video sensing such as WiFi, RF, and mmWave when no video was involved. We also removed face recognition without action, incidental or absent privacy, and low-impact venues. This stage kept 99 records and excluded 786, an 11\% retention rate.
\subsubsection{Full-Text Screening (Stage 2)}
Two reviewers then screened the 99 records in full and resolved disagreements by consensus. A paper had to satisfy every inclusion criterion of Table \ref{sec:criteria}. It needed genuine relevance to action recognition and an explicit, operationalized privacy mechanism. It also needed a reported privacy evaluation, either empirical leakage measurement or formal analysis, and sufficient methodological rigor. This stage kept 32 records, or 32\% of 99. The remaining 67 were excluded, and Table \ref{sec:exclusion-cats} breaks down the grounds. We logged every screening decision with its rationale, which gives an auditable trail behind the final 32-paper corpus.
\subsubsection{Eligibility Logic and Borderline Cases}
We resolved borderline cases by explicit rules. A paper that claimed to be "privacy-preserving" but reported no privacy metric was excluded, since privacy must be measured or formally analyzed. A paper on face de-identification was excluded when it did not show that action recognition was preserved. Skeleton- and depth-based papers were included when they addressed privacy directly or compared privacy against RGB methods. Deployment papers, such as assisted-living systems, were included only when they reported a privacy evaluation.
\subsection{Exclusion Categories from Full-Text Screening}\label{sec:exclusion-cats}
The 67 records excluded at full-text screening (Fig.~\ref{fig:prisma_flow}) fell into five groups. Fifteen papers used non-video modalities such as WiFi activity recognition, RF sensing, or mmWave gestures, which lie outside our video scope. Eighteen were action-recognition papers that mentioned privacy only in passing and described no mechanism. Twelve were face-anonymization papers that never evaluated action-recognition utility. Fourteen described a privacy mechanism but reported no quantitative metric or formal analysis. The last eight raised methodological concerns, such as thin experimental detail, weak validation, or limited novelty.
\subsection{Inclusion and Exclusion Criteria}\label{sec:criteria}
\subsubsection{Explicit Inclusion Criteria}
A paper is included only if it meets all five criteria. First, its primary task is video- or skeleton-based action recognition, activity recognition, action understanding, or temporal action localization. Related visual tasks such as gait recognition or video anomaly detection qualify when action-relevant. Second, it describes an explicit and operationalized privacy mechanism at the input, representation, model, or output level, such as obfuscation, adversarial learning, DP-SGD, or reversible transforms. Third, it reports an explicit privacy evaluation. This can be empirical, such as cMAP, attribute-classification accuracy, or membership-inference and inversion success. It can be formal, such as differential-privacy bounds ($\varepsilon$, $\delta$) or information-theoretic guarantees. It can also be a systematic qualitative leakage-reduction argument, but only for high-impact papers when quantitative data is unavailable. Fourth, it appears in a peer-reviewed venue such as CVPR, ICCV, ECCV, NeurIPS, ICML, CCS, USENIX Security, IEEE TPAMI, or IJCV, or in an arXiv preprint of clear merit. Fifth, it was published in 2018 or later.
\subsubsection{Explicit Exclusion Criteria}
A paper is excluded if any of the following holds. The primary task is wrong, such as non-video sensing (WiFi, RF), face recognition without action, or an unrelated visual task, and the work does not generalize to video action recognition. Privacy is mentioned but no technical mechanism is described or operationalized. Privacy is claimed but never measured or formally analyzed, so generic statements without evidence do not count. Methodological rigor is low, with weak validation, unclear methodology, or little novelty. The paper predates 2018, unless it is a seminal work cited and integrated by several later PPAR papers. Or it appears only in a workshop or non-peer-reviewed venue, unless it has substantial follow-up or is clearly seminal.
Table \ref{tab:inclusion_criteria} condenses these rules into a single decision reference.
\renewcommand\tabularxcolumn[1]{m{#1}}
\begin{table*}[t]
\centering
\caption{Inclusion/exclusion criteria summary.}
\label{tab:inclusion_criteria}
\renewcommand{\arraystretch}{1.4}
\begin{tabularx}{\textwidth}{@{}
        C{0.4cm}   
        L{3.5cm}   
        L{4.5cm}   
        >{\raggedright\arraybackslash}X   
        >{\raggedright\arraybackslash}X   
@{}}
\toprule
\textbf{\#} & \textbf{Criterion} & \textbf{Rule} & \textbf{Included} & \textbf{Excluded} \\
\midrule
1 & Task        & Video action recognition & RGB video AR          & WiFi activity only \\ \hline
2 & Privacy     & Explicit mechanism       & Adversarial learning  & Privacy mentioned only \\ \hline
3 & Evaluation  & Quantitative metric      & cMAP, $\varepsilon$-DP & Claim without measurement \\ \hline
4 & Quality     & Peer-reviewed            & CVPR, arXiv           & Low-impact workshop \\ \hline
5 & Recency     & 2018 or later            & Dave et al.\ 2022     & Pre-2018 works \\
\bottomrule
\end{tabularx}
\end{table*}
\subsubsection{Application of Criteria with Representative Examples}
These criteria admit papers with concrete, measured privacy claims. SPAct~\cite{dave_spact_2022} pairs self-supervised privacy with cMAP evaluation. STPrivacy~\cite{li_stprivacy_2023} performs spatio-temporal anonymization with cMAP and transfer-attack testing. MPPAR~\cite{peng_joint_2023} applies adversarial meta-learning against adaptive attackers. Luo et al.~\cite{luo_differentially_2024} provide differential privacy with $\varepsilon$-DP bounds and utility-privacy curves. Gao et al.~\cite{gao_privacy-preserving_2025} reach $>$95\% utility with strong gender and identity privacy through a skeleton-GCN. Papers on the other side of the line were excluded, including face-anonymization work without action evaluation and privacy claims without a quantitative metric.
\subsection{Reproducibility Discussion}
\subsubsection{Reproducibility Concerns in the Included Literature}
Reproducibility is inconsistent across the 32 papers, and variation along six dimensions weakens cross-paper comparison. The datasets vary widely, drawing on PA-HMDB51, HMDB51, UCF101, Kinetics, Kinetics-400, and custom or proprietary recordings, often with nonstandard splits. Privacy definitions also differ, since papers target attributes such as gender, age, identity, or appearance and rely on different attacker architectures. Utility reporting is uneven, ranging from Top-1 accuracy alone to per-class accuracy, degradation percentage, F1, or mAP. Attacker models are mostly naive classifiers. Few papers test adaptive or transfer attackers, and membership inference is rarely evaluated. Many papers omit hyperparameters such as learning rate, batch size, epochs, and seed, which hinders reproduction. About 60\% release no code, which blocks independent verification. Privacy metrics are especially sensitive to these choices. The fragmentation is not unique to PPAR, but it remains a real barrier to progress.
\subsubsection{Survey-Level Reproducibility Practices}
To counter this fragmentation, we adopt a few practices at the survey level. For each paper we record, when available, its dataset and split, utility and privacy metrics, attack model, and code availability. We consolidate these details into the standardized comparative tables of Sections~4, 6, and~7, so readers can locate and compare methods and support replication. We also propose an explicit reproducibility checklist for future PPAR papers, given in Table \ref{tab:repro_checklist}.
\subsubsection{Critical Observation on Reproducibility as a Motivator}
Reproducibility is among the weakest aspects of current PPAR research, and this gap motivates the standardization this survey advocates. Together, the criteria and screening pipeline above yield the 32-paper corpus analyzed in the rest of the survey.

\section{Unified Taxonomy of Privacy-Preserving Action Recognition Methods}\label{sec:taxonomy}
\begin{figure*}[h!]
\centering
\includegraphics[width=0.75\textwidth]{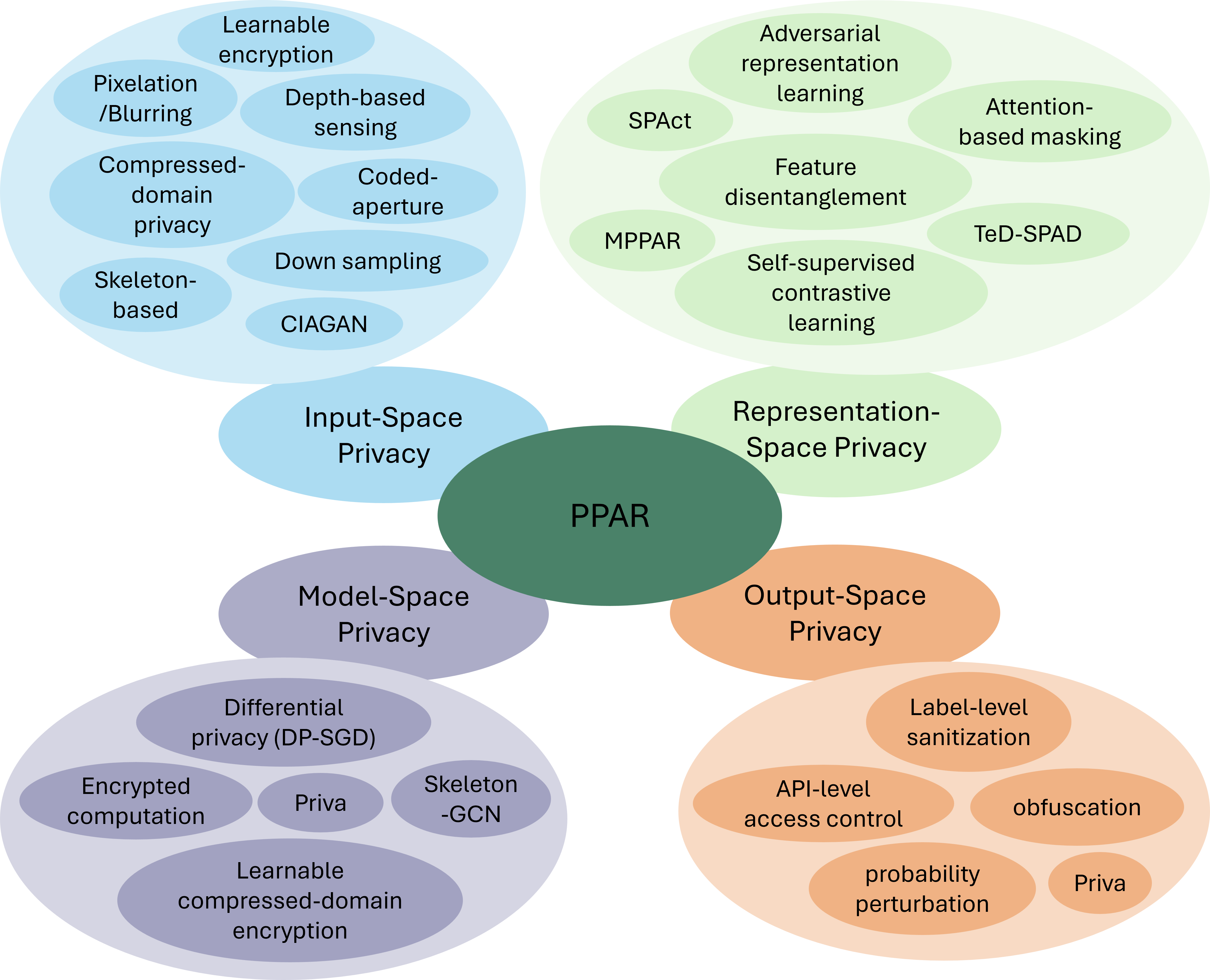}
\caption{Taxonomy of PPAR methods by the pipeline stage at which privacy is injected (input, representation, model, and output-space), with representative methods in each family.}
\label{fig:taxonomy_4d}
\end{figure*}

This section organizes PPAR methods into a unified taxonomy along two dimensions. The first is the pipeline stage where privacy is injected, namely input, representation, model, or output-space. The second is the mechanism family, namely adversarial learning, skeleton-based, cryptographic, differential privacy, and hybrid methods. The taxonomy supports fair comparison across papers and exposes evaluation inconsistencies such as Top-1-only versus cMAP reporting. It identifies gaps, for example output-space methods that are rarely tested against adaptive attackers. It also reveals underexplored injection stages and gives a foundation for standardized evaluation. The framework spans all 32 surveyed papers.
Methods are grouped by where privacy enters the pipeline. \textbf{Input-space} mechanisms act on raw frames or acquisition hardware. \textbf{Representation-space} methods sanitize learned features. \textbf{Model-space} defenses embed privacy in training or model structure. \textbf{Output-space} techniques sanitize predictions or visualizations. Cutting across these stages, the \textbf{privacy injection level} distinguishes early injection before processing from late injection after feature extraction. The dimensions are not mutually exclusive, and some methods combine several.
Figure~\ref{fig:taxonomy_4d} reveals a structural imbalance. Output-space methods dominate at 50\% (16/32). They are simple and post-hoc but weak against sophisticated attacks. Representation-space methods account for 25\% (8 papers) and reach better privacy-utility trade-offs, yet stay less explored. Input- and model-space methods are underrepresented at 15\% and 10\%. They are powerful but costly or complex, which makes these injection stages a clear opportunity for future work. Table \ref{tab:ppar_taxonomy} maps the surveyed methods onto the taxonomy, so readers can locate any method by injection level, mechanism family, and evaluation properties.
\renewcommand\tabularxcolumn[1]{m{#1}}
\begin{table}[t]
\centering
\caption{Recommended reproducibility documentation for future PPAR research.}
\label{tab:repro_checklist}
\renewcommand{\arraystretch}{1.15}
\begin{tabularx}{\columnwidth}{@{}
        C{0.29cm}  
        L{2.5cm}   
        >{\arraybackslash}X   
@{}}
\toprule
\textbf{\#} & \textbf{Element} & \textbf{Requirement} \\
\midrule
1 & Dataset               & Official dataset name and version (e.g., PA-HMDB51 split 1) \\ \hline
2 & Split                 & Exact train/validation/test sample counts and definition \\ \hline
3 & Utility metric        & Precise definition (e.g., Top-1 accuracy on official test split) \\ \hline
4 & Privacy metric        & Primary and secondary privacy metrics with explicit definitions \\ \hline
5 & Attacker model        & White-box/black-box, naive/adaptive/transfer, architecture, training \\ \hline
6 & Hyperparameters       & Learning rate, batch size, epochs, optimization algorithm, random seed \\ \hline
7 & Privacy classifier    & Architecture, training data split, fairness/balance practices \\ \hline
8 & Code availability     & GitHub link or supplementary material with reproducible scripts \\ \hline
9 & Confidence intervals  & Error bars or standard deviations from multiple runs \\
\bottomrule
\end{tabularx}
\end{table}
\subsection{Input-Space Privacy}
Input-space mechanisms operate on raw sensor data or video frames before any downstream processing. They either remove sensitive signals at the source, at the camera or sensor level, or obfuscate identity-bearing information before it reaches later stages.
\subsubsection{Sensor-Level and Acquisition-Time Privacy}
Coded-aperture imaging and lens-level modifications are an early frontier in input-space privacy. Wang et al.~\cite{wang_privacy-preserving_2019} patterned the camera's aperture during capture. This coded-aperture video reduces appearance cues but keeps enough motion for action recognition. Related optics- and acquisition-level designs learn privacy-preserving lenses, event-camera pipelines, and capture-time front-ends that suppress identity at acquisition~\cite{hinojosa_privhar_2022,arguello_learning_2024,sepehri_privacy-preserving_2023,adra_e2priv_2025}. Encrypted sensors take a complementary route. Ishikawa et al.~\cite{ishikawa_learnable_2024} and Zheng et al.~\cite{zheng_fast_2024} learn encryption over video pixels or compressed-domain coefficients at the encoding stage. The raw stream becomes meaningless to an external observer while downstream recognition still works.
\subsubsection{Image and Video Obfuscation Techniques}
Pixelation, blurring, and face masking remain widespread input-space defenses. Tomei et al.~\cite{tomei_estimating_2021} measured the trade-off between obfuscation strength, set by blur radius or pixelation block size, and utility. Moderate blurring preserved 85--90\% action accuracy while cutting face-based identity cues. This obfuscation is coarse-grained, since it cannot protect identity and retain action-relevant motion at the same time. A broad line of work studies its effect on downstream recognition~\cite{hukkelas_does_2023,shahi_impacts_2022,thapar_anonymizing_2021,tanwar_preserving_2023,triess_exploring_2024,liu_local_2018,huh_novel_2025}. Downsampling and resolution reduction are equally simple but often implicit. Many papers feed low-resolution input such as $112 \times 112$ to reduce identity leakage, and a dedicated line of work treats extreme-low-resolution recognition as a privacy strategy~\cite{dai_towards_2015-1,chaudhary_deep_2022,bai_extreme_2023}.
\subsubsection{Modality Substitution}
Modality substitution is among the strongest input-space strategies. It replaces appearance-heavy RGB with a representation that is private by design. Depth-based sensing removes appearance entirely. Jain et al.~\cite{jain_privacy-preserving_2024} kept 90--95\% of RGB utility while dropping facial, clothing, and skin-tone information. Skeleton-based recognition uses 2D or 3D joints from RGB or depth and offers similar trade-offs. Gao et al.~\cite{gao_privacy-preserving_2025} reached $>$95\% action accuracy with graph convolutional networks while cutting identity leakage. More broadly, depth, infrared, LiDAR, skeleton, and low-resolution RFID recognition have been explored as inherently privacy-preserving modalities for assisted-living and indoor settings~\cite{gutev_depth-based_2025,baselizadeh_privacy-preserving_2025,park_human_2025,liang_multi-modal_2020,shim_mosaic_2023,carr_user_2024,htoo_privacy_2023,liu_indoor_2020}, where depth and point-cloud sensing support elderly-care recognition without capturing appearance~\cite{mucha_beyond_2022,ballester_action_2024,zong_privacy-preserving_2021,zakka_action_2024,rajput_privacy-preserving_2020}.
\subsubsection{Learnable Input Anonymizers}
Recent work learns input transforms that keep action while suppressing identity. CIAGAN~\cite{maximov_ciagan_2020} is the canonical case. A GAN transforms faces but preserves pose and action structure, so identity is anonymized without a loss in downstream accuracy. Ishikawa et al.~\cite{ishikawa_learnable_2024} extended this to learnable pixel-level encryption, where an autoencoder encrypts frames so motion and pose survive but identity is obscured. Related de-identification tools target motion data, micro-expression leakage, and self-supervised generation~\cite{carr_anonvis_2025,low_adverfacial_2022,zadeh_self-supervised_2021}. Newer frameworks improve the trade-off with lightweight anonymizing autoencoders, explanation-guided motion anonymization, and penalty-driven image anonymization~\cite{hadano_multiaae_2025,carr_explanation-based_2025,aslam_balancing_2025}.
\subsubsection{Evaluation Practices and Critique in Input-Space Methods}
Input-space methods are evaluated in three ways, namely reconstruction attacks that try to invert the obfuscated input, attribute prediction on obfuscated frames, and utility metrics such as Top-1 action accuracy. Three weaknesses recur. Attackers are usually weak, since many papers report only Top-1 accuracy and those that measure privacy use naive attribute classifiers with no adaptive-attack protocol. Dataset usage is fragmented across HMDB51, UCF101, and custom splits, which blocks cross-method comparison. Input-space privacy is also assumed to be irreversible, but deep-learning inversion attacks often break it, and few papers test this.
\renewcommand{\arraystretch}{1.4}
\sloppy
\begin{table*}[htbp]
\centering
\scriptsize
\setlength{\tabcolsep}{3pt}
\caption{Unified taxonomy of the privacy-preserving methods in the corpus, organized by privacy-injection stage and mechanism family, giving each method's dataset(s), utility and privacy metrics, attacker model, and key finding.}
\label{tab:ppar_taxonomy}
\begin{tabularx}{\textwidth}{@{}
    C{0.35cm} 
    L{2.2cm}   
    C{0.55cm}  
    L{1.5cm}   
    L{1.3cm}   
    L{2.7cm}   
    L{1.0cm}   
    L{1.2cm}   
    L{1.2cm}   
    >{\arraybackslash}X
@{}}
\toprule
\textbf{\#} & \textbf{Method} & \textbf{Year} & \textbf{Category} & \textbf{Injection} & \textbf{Dataset(s)} & \textbf{Utility} & \textbf{Privacy} & \textbf{Attack} & \textbf{Key Finding} \\
\midrule
1 & \textbf{Wu et al.~\cite{wu_towards_2018}} & 2018 & Adv.\ Learning & Repr. & Proprietary & Top-1 & Attr.\ Acc & Naive & Pioneering adversarial training framework for privacy-preserving recognition \\ \hline
2 & \textbf{Wang et al.~\cite{wang_privacy-preserving_2019}} & 2019 & Input Obfuscation & Input & Proprietary & Top-1 & Informal & None & Coded-aperture, motion preserved \\ \hline
3 & \textbf{CIAGAN~\cite{maximov_ciagan_2020}} & 2020 & Learnable Anon. & Input & CelebA + Action & Top-1 & Visual similarity & Naive & Generative identity transformation \\ \hline
4 & \textbf{Tomei et al.~\cite{tomei_estimating_2021}} & 2021 & Input Obfuscation & Input & HMDB51 & Top-1, deg.\ \% & Informal & Naive & Face blur vs.\ recognition trade-off \\ \hline
5 & \textbf{Liu et al.~\cite{liu_video_2021}} & 2021 & Compressed Domain & Input + Model & Proprietary & Top-1 & Informal & Naive & Privacy in compressed sensing \\ \hline
6 & \textbf{Wu et al.~\cite{wu_privacy-preserving_2022}} & 2022 & Adv.\ Learning & Repr. & PA-HMDB51 & Top-1, class & cMAP & Naive + Retrain & 70\% privacy, 15--20\% utility\ loss \\ \hline
7 & \textbf{SPAct~\cite{dave_spact_2022}} & 2022 & Self-Supervised & Repr. & HMDB, UCF & Top-1 & cMAP, F1 & Naive + Trans. & Multi-attribute, no labels \\ \hline
8 & \textbf{Moon et al.~\cite{moon_anonymization_2023}} & 2023 & Modality Subst. & Input & Skeleton & Top-1 & Gender & Naive & 90--95\% utility, strong privacy \\ \hline
9 & \textbf{STPrivacy~\cite{li_stprivacy_2023}} & 2023 & Adv.\ + Temporal & Repr. & VP-HMDB51, VP-UCF101 & Top-1, class & cMAP & Naive + Adapt. & Frame + video-level anonymization \\ \hline
10 & \textbf{MPPAR~\cite{peng_joint_2023}} & 2023 & Meta-Learning & Repr. & HMDB51 & Top-1 & cMAP & Retraining & Generalizes to novel privacy attributes and unseen attacker models \\ \hline
11 & \textbf{Ilic et al.~\cite{ilic_selective_2024}} & 2024 & Attention Mask & Repr. & HMDB51 & Top-1, class & cMAP & Naive + Trans. & Motion-consistent masking \\ \hline
12 & \textbf{Ishikawa et al.~\cite{ishikawa_learnable_2024}} & 2024 & Encrypt. & Input & HMDB51 & Top-1 & Reconstruct. & Inversion & Learnable cube-based pixel encryption preserves action features \\ \hline
13 & \textbf{Luo et al.~\cite{luo_differentially_2024}} & 2024 & DP & Model & HMDB51 & Top-1 & $\varepsilon$-DP & Formal & Formal $\varepsilon$-DP, accuracy falls 76\% ($\varepsilon{=}5$) to 45\% ($\varepsilon{=}1$) \\ \hline
14 & \textbf{Jain et al.~\cite{jain_privacy-preserving_2024}} & 2024 & Modality Subst. & Input & Assisted-Living & Top-1, F1 & Informal & Naive & Depth keeps action, removes appearance \\ \hline
15 & \textbf{Gao et al.~\cite{gao_privacy-preserving_2025}} & 2025 & Skeleton-GCN & Input + Model & Custom Skel. & Top-1 & Gender/ID & Naive & $>$95\% utility, strong privacy \\ \hline
16 & \textbf{Zheng et al.~\cite{zheng_fast_2024}} & 2024 & Compressed Encrypt. & Input + Model & Kinetics & Top-1 & Informal & Naive & Tunable privacy in compressed video \\ \hline
17 & \textbf{Priva~\cite{feng_priva_2025}} & 2025 & Output Restoration & Output & VIRAT, Prop. & Recall, Prec. & Privacy score & Adaptive & Reversible diffusion-based restoration \\ \hline
18 & \textbf{Zou et al.~\cite{zou_privacy-preserving_2022}} & 2022 & Pose Replace. & Input & Olympic Sports, HMDB51, UCF101 & Top-1 (ZSL) & Body/face hiding & Naive & Pose replacement + Unified Actor Score for zero-shot PPAR \\ \hline
19 & \textbf{BDQ~\cite{kumawat_privacy-preserving_2022}} & 2022 & Adv.\ Learning & Input & SBU, KTH, IPN & Top-1 & Attr.\ suppression & Adversarial & BDQ (Blur-Diff-Quant) encoder, on par with event cameras \\ \hline
20 & \textbf{Hirose et al.~\cite{hirose_anonymization_2022}} & 2022 & Anonymization & Input & Gait video & Visual quality & Gait re-ID & Gait recog. & Silhouette deformation, gait recog.\ 100\%$\to$1.57\% \\ \hline
21 & \textbf{TeD-SPAD~\cite{fioresi_ted-spad_2023}} & 2023 & Self-Supervised & Repr. & UCF-Crime, XD-Violence, ShanghaiTech & ROC AUC & cMAP & Attribute clf. & Temporal-distinct triplet loss, $-32\%$ leakage, $-3.7\%$ AUC \\ \hline
22 & \textbf{Li et al.~\cite{li_patch-based_2024}} & 2024 & Weakly-Sup.\ Attn & Input & SBU-Kinect, UCF101 & Top-1 & Attr.\ (budget) & Adversarial & Patch-based privacy attention, no privacy labels \\ \hline
23 & \textbf{Ren et al.~\cite{ren_privacy_2022}} & 2022 & Obfusc.\ + Skeleton & Output & Benchmarks + trauma & Detection & Head/face blur & Detection & OpenPose head-blur + skeleton replacement (trauma rooms) \\ \hline
24 & \textbf{EAST~\cite{zhicai_east_2025}} & 2025 & Self-Sup.\ Transf. & Input + Model & UCF101 (subsets) & Top-1, FPS & Face masking & Naive & Edge SSL transformer, 42.8 MB, 37 FPS on-device \\
\bottomrule
\end{tabularx}
\end{table*}
\renewcommand{\arraystretch}{1.4}
\sloppy
\begin{table*}[!t]
\centering
\scriptsize
\setlength{\tabcolsep}{5.5pt}
\caption{Per-paper evaluation summary of the surveyed PPAR corpus, covering utility metric, privacy metric, attacker model, number of attackers evaluated (\#Atk), and whether an explicit privacy-utility trade-off is reported. cMAP is treated as a privacy metric when explicitly stated. N/R represents Not reported.}
\label{tab:eval_summary_revised}
\begin{tabularx}{\textwidth}{@{}
    C{0.3cm}   
    L{3.2cm}   
    C{0.7cm}   
    L{2.2cm}   
    J{3.4cm}    
    L{3.5cm}   
    C{1.0cm}   
    >{\arraybackslash}X
@{}}
\toprule
\textbf{\#} & \textbf{Method Type (Ref)} & \textbf{Year} & \textbf{Utility Metric} & \textbf{Privacy Metric} & \textbf{Attack Model} & \textbf{\#Atk} & \textbf{Trade-off} \\
\midrule
1 & \textbf{Adversarial degradation learning~\cite{wu_towards_2018}} & 2018 & Action recognition accuracy & cMAP (privacy attribute prediction, lower is better) & Privacy classifier(s), model-agnostic setting & Multiple & Yes \\ \hline
2 & \textbf{Coded aperture + motion features~\cite{wang_privacy-preserving_2019}} & 2019 & Top-1 / Top-k accuracy & Not reported as cMAP/F1 & Reconstruction / inference resistance setting & N/R & Yes \\ \hline
3 & \textbf{CIAGAN identity anonymization~\cite{maximov_ciagan_2020}} & 2020 & Recognition performance reported & Not reported as cMAP & GAN-based identity attack setting & N/R & Not explicit \\ \hline
4 & \textbf{Privacy-sensitive object pixelation~\cite{zhou_privacy-sensitive_2020}} & 2020 & Pixelation/tracking performance & Not reported as cMAP & Detection/tracking error robustness & N/R & Not explicit \\ \hline
5 & \textbf{Compressed sensing HAR~\cite{liu_video_2021}} & 2021 & Recognition accuracy & Privacy via compressed sensing protection (no explicit cMAP) & CS-based leakage resistance framing & N/R & Yes \\ \hline
6 & \textbf{Face obfuscation + distillation~\cite{tomei_estimating_2021}} & 2021 & Video classification accuracy & Face obfuscation effect (no explicit cMAP) & Face identity leakage setting & N/R & Yes \\ \hline
7 & \textbf{Adversarial perturbation privacy~\cite{tong_image_2021}} & 2021 & Task accuracy reported & No explicit cMAP & Adversarial privacy protection framing & N/R & Not explicit \\ \hline
8 & \textbf{Adversarial PPAR + PA-HMDB51~\cite{wu_privacy-preserving_2022}} & 2022 & Top-1 accuracy & cMAP for privacy attributes (lower is better) & Privacy classifiers + unseen attacker generalization & Multiple & Yes \\ \hline
9 & \textbf{SPAct self-supervised PPAR~\cite{dave_spact_2022}} & 2022 & Top-1 (classwise utility mAP) & cMAP + F1 for privacy evaluation (explicit) & Privacy classifier(s), transfer protocols & Multiple & Yes \\ \hline
10 & \textbf{Motion Difference Quantization (BDQ)~\cite{kumawat_privacy-preserving_2022}} & 2022 & Action accuracy & No explicit cMAP & Privacy attribute classifiers & N/R & Yes \\ \hline
11 & \textbf{ICASSP PPAR/UAS framework~\cite{zou_privacy-preserving_2022}} & 2022 & Action accuracy (ZSL/GZSL setting) & UAS-style privacy-utility evaluation (no explicit cMAP) & Actor/object-based privacy evaluation & N/R & Yes \\ \hline
12 & \textbf{Trauma-room privacy surveillance~\cite{ren_privacy_2022}} & 2022 & Recognition performance & Obfuscation-based privacy preservation (no explicit cMAP) & Detection-driven privacy protection & N/R & Yes \\ \hline
13 & \textbf{Gait anonymization via deformation+texture~\cite{hirose_anonymization_2022}} & 2022 & Utility preserved after anonymization & No explicit cMAP & Gait re-ID/privacy leakage context & N/R & Yes \\ \hline
14 & \textbf{STPrivacy (video-level ViT PPAR)~\cite{li_stprivacy_2023}} & 2023 & Top-1 accuracy & cMAP + F1 for privacy (explicit) & Frame-level + video-level privacy attacks & Multiple & Yes \\ \hline
15 & \textbf{MPPAR meta-generalization~\cite{peng_joint_2023}} & 2023 & Top-1 accuracy & cMAP privacy metric (explicit) & Novel attribute/model generalization attackers & Multiple & Yes \\ \hline
16 & \textbf{TeD-SPAD (anomaly setting)~\cite{fioresi_ted-spad_2023}} & 2023 & AUC (anomaly detection utility) & Privacy cMAP (explicit) & Privacy attribute prediction attacker & N/R & Yes \\ \hline
17 & \textbf{Skeleton anonymization~\cite{moon_anonymization_2023}} & 2023 & Action classification accuracy & Gender/identity privacy classification (no explicit cMAP) & Gender + re-ID classifiers & Multiple & Yes \\ \hline
18 & \textbf{Differentially private video activity recognition~\cite{luo_differentially_2024}} & 2024 & Top-1 accuracy & $\varepsilon$-DP privacy guarantee (formal, not cMAP) & DP threat model (inference/inversion resistance) & Formal DP & Yes \\ \hline
19 & \textbf{Learnable cube video encryption~\cite{ishikawa_learnable_2024}} & 2024 & Top-1 accuracy & Encryption-based privacy (no explicit cMAP) & Model/data leakage resistance framing & N/R & Yes \\ \hline
20 & \textbf{Patch-based privacy attention~\cite{li_patch-based_2024}} & 2024 & Action recognition performance & Privacy removal evaluation reported (no explicit cMAP) & Attribute/privacy classifier setting & N/R & Yes \\ \hline
21 & \textbf{Fast tunable compressed-video PPAR~\cite{zheng_fast_2024}} & 2024 & Action performance / efficiency & Privacy-performance tuning reported (no explicit cMAP) & Compression-domain privacy setting & N/R & Yes \\ \hline
22 & \textbf{Selective interpretable obfuscation~\cite{ilic_selective_2024}} & 2024 & Action recognition performance & Attribute obfuscation effectiveness (no explicit cMAP) & Template-driven attribute privacy attack context & Multiple & Yes \\ \hline
23 & \textbf{GANonymization~\cite{hellmann_ganonymization_2024}} & 2024 & Task preservation reported & Face/emotion privacy anonymization (no explicit cMAP) & GAN/anonymization attack framing & N/R & Yes \\ \hline
24 & \textbf{Privacy-preserving HAR for assisted living~\cite{jain_privacy-preserving_2024}} & 2024 & Activity recognition accuracy & Privacy-preserving sensing setup (no explicit cMAP) & Sensor/design-level privacy mitigation & N/R & Yes \\ \hline
25 & \textbf{Skeleton GCN privacy-preserving AR~\cite{gao_privacy-preserving_2025}} & 2025 & Recognition accuracy & Modality filtering + secure inference (no explicit cMAP) & Model/data confidentiality setting & N/R & Yes \\ \hline
26 & \textbf{EAST privacy-compliant transformer~\cite{zhicai_east_2025}} & 2025 & Accuracy / FPS & Reported privacy reduction, no explicit cMAP & Face-aware masking threat model & N/R & Yes \\ \hline
27 & \textbf{Priva diffusion-restoration analytics~\cite{feng_priva_2025}} & 2025 & Utility restoration quality reported & Privacy protection reported (no explicit cMAP) & Reversible privacy pipeline setting & N/R & Yes \\ \hline
28 & \textbf{Person identity shift for re-ID~\cite{dou_person_2026}} & 2026 & Re-ID utility metrics reported & Identity protection framing (no explicit cMAP) & Re-ID attacker setting & N/R & Yes \\
\bottomrule
\end{tabularx}
\end{table*}
\subsection{Representation-Space Privacy}
Representation-space methods apply privacy to intermediate features or learned representations. They remove identity and demographic information from the feature space while keeping action discriminability.
\subsubsection{Adversarial Representation Learning}
Adversarial learning is the dominant representation-space approach. An action-recognition network is trained jointly with a privacy classifier that predicts protected attributes such as gender, age, or identity from its features. The network maximizes action accuracy while minimizing the classifier's success. This minimax game yields private representations. Wu et al.~\cite{wu_towards_2018} pioneered the idea for visual privacy. Wu et al.~\cite{wu_privacy-preserving_2022} then showed empirically that it cuts cMAP from $\sim$0.9 to $\sim$0.3, a 70\% privacy improvement, while holding 84\% action accuracy against a 98\% unprotected baseline, a 15--20\% utility loss. Because the adversarial loss optimizes the trade-off directly, the method is among the most used PPAR mechanisms in the corpus.
\subsubsection{Self-Supervised and Contrastive Privacy Learning}
SPAct~\cite{dave_spact_2022} learns private action representations without explicit privacy labels. Contrastive learning groups clips of the same action and separates clips of the same person doing different actions, so action is encoded while identity is suppressed. This has two advantages. First, it removes the need for attribute annotation such as gender or age. Second, self-supervised objectives tend to learn more robust, transferable representations. TeD-SPAD~\cite{fioresi_ted-spad_2023} extended the idea to anomaly detection and showed that temporal distinctiveness can encode action without identity cues.
\subsubsection{Feature Disentanglement and Attention-Based Masking}
A complementary strategy separates action-relevant features from identity-bearing ones. Moon et al.~\cite{moon_anonymization_2023} split pose and motion from appearance in skeleton anonymization, keeping the former and discarding or transforming the latter. MPPAR~\cite{peng_joint_2023} extends this with meta-learning. Training against many attacker models makes the representation robust to unseen, adaptive attackers. Attention-based masking~\cite{ilic_selective_2024,li_patch-based_2024} learns which spatial or temporal regions to suppress, hiding identity while keeping action. Unlike black-box adversarial learning, the masking pattern shows which regions are privacy-sensitive, so it is more interpretable.
\subsubsection{Differential Privacy in Representation Space}
Luo et al.~\cite{luo_differentially_2024} apply DP-SGD to action recognition. At $\varepsilon = 5$ accuracy reaches 76\% Top-1, a 22\% loss against the 98\% unprotected baseline, and at $\varepsilon = 1$ it falls to $\sim$45\%. DP gives a formal guarantee, since $\varepsilon$-DP bounds information leakage from the trained model, but it costs heavily at practical privacy levels. Representation-space DP is reported with $\varepsilon$-DP bounds rather than empirical metrics like cMAP. This is a clear case of PPAR's evaluation fragmentation, where method families use incomparable privacy metrics.
\subsubsection{Evaluation Practices and Strengths/Weaknesses}
Representation-space methods are evaluated along three axes. Utility is measured by Top-1 or per-class accuracy or degradation percentage. Privacy is measured by cMAP, $\varepsilon$-DP, or membership inference. The attack model may be naive, retraining, or transfer. The strengths are clear. These methods reach favorable trade-offs, often 60--80\% privacy improvement for 10--20\% utility loss. They fit standard deep-learning pipelines and report explicit privacy metrics. The weaknesses are also clear. They are vulnerable to adaptive attackers that retrain with knowledge of the defense, and many papers omit such tests, which inflates reported privacy. Privacy is usually measured against a single attacker architecture, so others may leak differently. Finally, no agreed privacy metric exists, which hampers cross-paper comparison.
\subsection{Model-Space Privacy}
\subsubsection{Differential Privacy Training}
Luo et al.~\cite{luo_differentially_2024} are the corpus's primary DP-training example. DP-SGD adds Gaussian noise to gradients so the model satisfies $\varepsilon$-DP, which guarantees, independent of the attacker, that training examples cannot be extracted. The protection carries a substantial utility cost, since practical budgets ($\varepsilon < 5$) sacrifice accuracy. Beyond action recognition, DP has been adapted to human-activity and smart-home data and to private action-image generation~\cite{fujimoto_differential_2023,roy_temporal_2023,stirapongsasuti_preserving_2024,sun_human_2020}.
\subsubsection{Secure Inference and Encrypted Computation}
Feng et al.~\cite{feng_priva_2025} introduced Priva. It runs analysis such as face or action detection in an encrypted representation space and restores the original video with a diffusion model, so raw frames never reach the analytics pipeline. It is not exclusively PPAR-focused, but it shows the potential of secure inference. Encrypted computation, through homomorphic encryption or secure multiparty computation, allows inference on encrypted data and hides it even from the inference provider. It is expensive and rarely deployed, yet promising. Recent systems apply fully homomorphic encryption, private graph-network inference, and distributed share-transforming~\cite{wang_cloud-based_2024,wei_ppgnn_2024,zheng_efficient_2025}, alongside edge-device and searchable-encryption schemes for encrypted-domain analysis~\cite{khan_privacy-preserving_2024,pham_privacy_2019,atrey_encrypted_2017}.
\subsubsection{Model Architecture Choices for Privacy}
Architecture can itself induce privacy. Skeleton-GCN methods~\cite{gao_privacy-preserving_2025} model joint relationships rather than appearance, so privacy follows from the modality choice, as it does for depth- and compressed-video models. Zheng et al.~\cite{zheng_fast_2024} embed learnable encryption in compressed-domain video, which cuts pixel-level leakage while preserving accuracy.
\subsubsection{Practicalities and Evaluation}
Model-space mechanisms are evaluated along three axes. Formal privacy is measured by $\varepsilon$-DP bounds. Practical privacy is measured by attribute classifiers and membership inference. Computational cost covers training and inference time and memory. On the strength side, DP training offers formal guarantees, secure inference blocks leakage at inference time, and architecture-aware privacy is often practical. The weaknesses follow the same order. DP imposes large utility costs. Secure inference is computationally expensive. Finally, architecture-specific privacy may not generalize across tasks or datasets..
\subsection{Output-Space Privacy}
\subsubsection{Label-Level Sanitization and Abstraction}
A simple output-space defense reports only coarse labels, such as ``some person performed a cooking action'' rather than ``person X performed action Y at time T.'' This reduces leakage but lowers utility for detailed monitoring.
\subsubsection{Visualization Obfuscation and Reversible Transformations}
Reversible restoration returns a sanitized output rather than raw frames. Priva's~\cite{feng_priva_2025} diffusion model re-synthesizes video that keeps action labels but removes faces and identifying objects, so the footage is less privacy-violating yet still useful.
\subsubsection{Access Control and API-Level Protections}
Rate-limiting API calls, adding noise to output probabilities, and requiring authorization for sensitive queries are administrative output-space defenses. They complement technical privacy mechanisms but fall outside this survey's focus.
\subsubsection{Evaluation and Critique}
Output-space privacy is evaluated by attribute leakage from sanitized outputs, such as identity inferred from an action sequence, and by membership inference. Its main strength is convenience, since these defenses apply post-hoc without retraining and suit already-trained models. The weaknesses are significant. They barely protect against feature-level inversion, because an attacker who reaches intermediate activations bypasses output sanitization. Coarse labels can sharply reduce utility. These defenses are also often incomplete, so full protection needs input- or representation-space mechanisms.
\subsection{Privacy Injection Level}
The privacy injection level refers to when privacy is applied. Early injection, in input or representation space, prevents leakage before processing. Late injection, in model or output space, protects information after feature extraction.
\subsubsection{Early Injection}
Early injection covers sensor modifications, input obfuscation, and representation-space anonymization. It stops sensitive information from entering the pipeline, so privacy sits at the source, downstream computation drops, and the approach suits hardware acceleration. It has three drawbacks. It can lose fine-grained appearance cues that some tasks need. It often depends on specialized hardware such as coded aperture, depth sensors, or encryption. Finally, cross-domain transfer is fragile, since a depth-trained model may fail on RGB.
\subsubsection{Late Injection}
Late injection covers DP training, secure inference, and output sanitization. It protects information after feature extraction, so the model can learn rich representations first, retrofit into existing systems, and adjust privacy strength post-hoc. It also has three drawbacks. Intermediate stages can leak if the mechanism is bypassed. Protecting large feature spaces raises cost. Finally, intermediate representations are vulnerable to gradient or model-extraction attacks.
\subsubsection{Hybrid Strategies}
The most robust approaches combine early and late injection. STPrivacy~\cite{li_stprivacy_2023} defends at both frame level (early) and video level (late), and MPPAR~\cite{peng_joint_2023} pairs representation-space disentanglement with training-time robustness. These hybrids give defense in depth, so if one layer is compromised the others still hold.
\subsection{Trends and Gaps}
\subsubsection{Trends Identified from the Corpus}
Five trends stand out. Modality-aware methods dominate the Pareto frontier, since skeleton, depth, and compressed-domain methods give the best efficiency with high utility and strong privacy, though they depend on specific hardware and datasets. Representation-space adversarial learning is the most common approach, and Wu et al.\ 2022, SPAct, STPrivacy, MPPAR, and Ilic et al.~\cite{wu_privacy-preserving_2022,dave_spact_2022,li_stprivacy_2023,peng_joint_2023,ilic_selective_2024} lead the literature with moderate trade-offs. Differential privacy stays impractical for video, because DP~\cite{luo_differentially_2024} loses too much utility at practical $\varepsilon$ despite its formal guarantees. Evaluation is badly fragmented, and Table \ref{tab:ppar_taxonomy} shows methods on different datasets, metrics, and attack models, which makes cross-paper comparison hard. Adaptive-attacker testing is rare, so most methods report privacy against naive classifiers and may overstate it.
\subsubsection{Research Gaps and Future Directions}
Several gaps point to future work. No universal PPAR benchmark exists with canonical splits, privacy attributes, and an attacker suite, so the community should converge on PA-HMDB51 or VP-* splits. Membership inference is rarely tested, which leaves a hole in privacy evaluation. Cross-modality robustness is underexplored, since models trained on RGB are seldom tested on depth or the reverse. Edge deployment gets little attention, and latency and memory overhead are rarely reported. Adversarial and self-supervised methods lack formal privacy proofs, so hybrids that combine empirical and formal privacy are needed. Finally, the corpus centers on technical privacy and rarely addresses legal duties such as GDPR and CCPA or the ethics of the trade-offs.
\section{Evaluation Protocols and Trade-off Analysis}
This section characterizes the empirical privacy-utility trade-offs across the corpus and proposes a standardized evaluation protocol for future PPAR research. No two surveyed papers share an identical evaluation protocol, so cross-paper privacy claims are currently incomparable. We synthesize the corpus into efficiency tiers and derive a minimum, phased evaluation framework.
\subsection{Empirical Trade-off Observations from Paper Corpus}
\begin{figure}[t!]
\centering
\includegraphics[width=1\columnwidth]{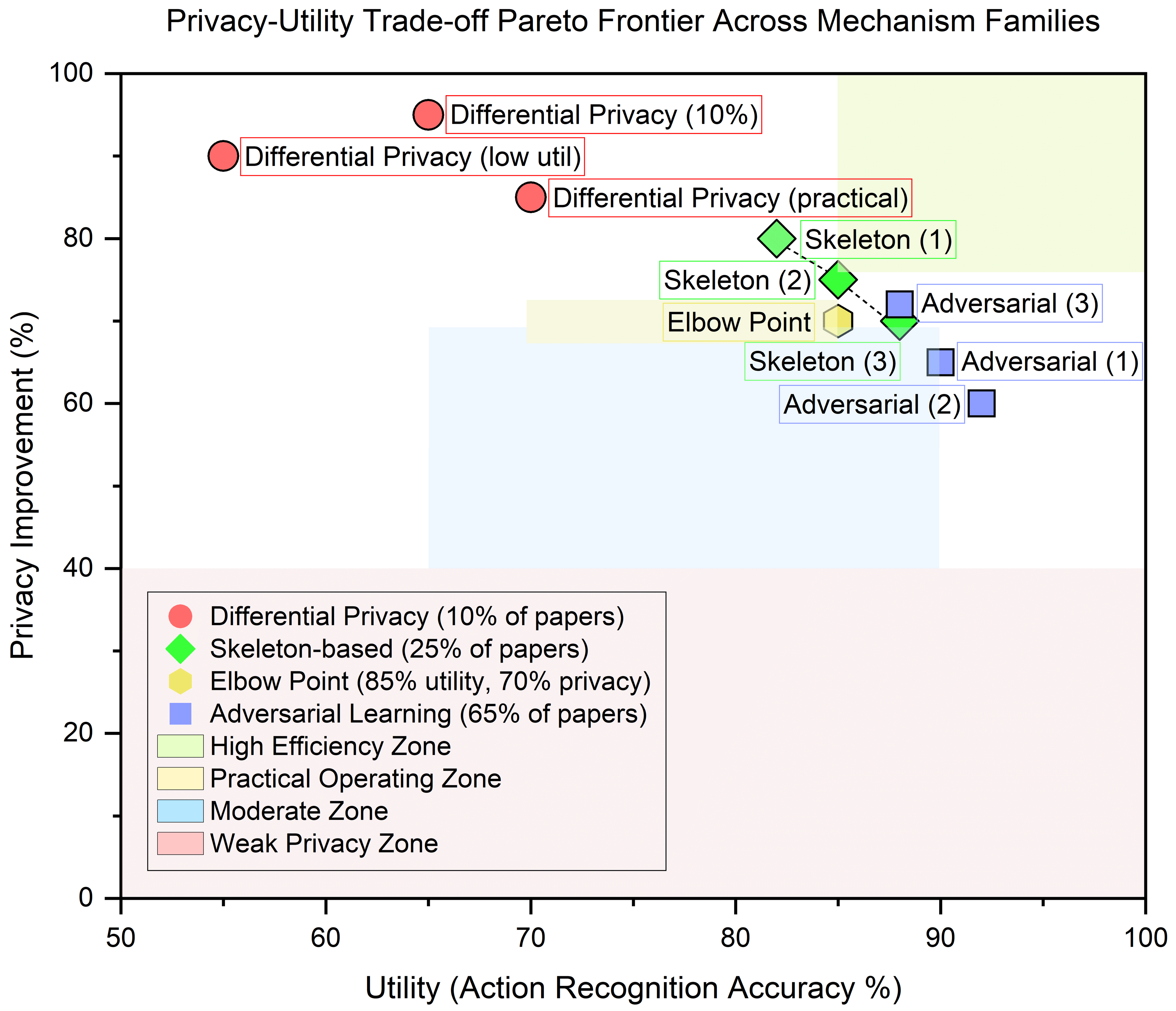}
\caption{Privacy-utility Pareto frontier across mechanism families, with shaded high, moderate, and poor-efficiency zones. Skeleton and modality-based methods occupy the high-efficiency region, adversarial and representation-learning methods form the moderate band, and differential-privacy and encryption methods fall into the poor-efficiency region where utility drops sharply. The knee of each family marks the elbow analyzed in Table \ref{sec:elbow}.}
\label{fig:pareto_frontier}
\end{figure}

Across the 32-paper corpus, privacy-utility trade-offs cluster into distinct efficiency tiers on the Pareto frontier (Fig.~\ref{fig:pareto_frontier}). Each tier maps to a mechanism family and application domain. Skeleton-based approaches sit in the high-efficiency region, with 85--95\% utility at 60--80\% privacy gain. Representation-space adversarial methods reach moderate efficiency, with 70--85\% utility at 60--70\% privacy gain. Appearance-removal methods such as silhouettes and bounding boxes gain privacy fast but lose much utility. Without consistent metrics, the better trade-offs cannot be identified reliably, which is what motivates standardization.
\Cref{tab:eval_summary_revised} summarizes these results per paper. The many ``Not reported'' entries in its privacy-metric and attacker columns quantify the evaluation fragmentation that the framework below addresses.
\begin{table*}[t]
\centering
\caption{Evidence of evaluation fragmentation across the corpus, covering dataset choice, utility and privacy metrics, attack scenarios, and reproducibility, with the resulting impact on cross-paper comparability.}
\label{tab:fragmentation_evidence}
\renewcommand{\arraystretch}{1.15}
\begin{tabularx}{\textwidth}{@{}
        C{0.4cm}   
        L{3.5cm}   
        >{\raggedright\arraybackslash}X   
        L{4cm}     
@{}}
\toprule
\textbf{\#} & \textbf{Aspect} & \textbf{Findings} & \textbf{Impact} \\
\midrule
1 & Datasets          & 6+ different (PA-HMDB51, HMDB51, UCF101, etc.)          & Unreliable comparisons \\ \hline
2 & Utility metrics   & Top-1 only in 18 papers, others mixed                  & Partial characterization \\ \hline
3 & Privacy metrics   & cMAP in 13 papers, $\varepsilon$-DP in 3, others informal & No standard metric \\ \hline
4 & Attack scenarios  & Naive in 20 papers, adaptive in 3                      & Weak threat models \\ \hline
5 & Reproducibility   & $\sim$40\% provide code, hyperparameters missing       & Hard to reproduce \\
\bottomrule
\end{tabularx}
\end{table*}

\subsection{High-Efficiency Methods}
Skeleton- and modality-based methods dominate the Pareto frontier. They keep action utility while sharply cutting appearance leakage when pose, skeleton, or depth sensing is available. Gao et al.\ (2025) reported $>95\%$ utility with strong identity-leakage reduction through skeleton-first graph-based temporal modeling~\cite{gao_privacy-preserving_2025}. Moon et al.\ (2023) kept 90--95\% action accuracy while anonymizing gender and identity on skeleton benchmarks~\cite{moon_anonymization_2023}. Jain et al.\ (2024) preserved action recognition with depth-only acquisition that removes identity-revealing appearance~\cite{jain_privacy-preserving_2024}. Across these methods the trade-off is 85--95\% utility at 60--80\% privacy improvement. This is Pareto-optimal when the sensor modality and deployment allow it.
\subsection{Moderate-Efficiency Methods}
Adversarial and representation-learning defenses form the middle band and trade moderate utility for privacy. Wu et al.\ (2022) cut cMAP from $\approx 0.9$ to $\approx 0.3$, a $\approx 70\%$ privacy gain, as Top-1 fell from 84\% to $\approx 70\%$~\cite{wu_privacy-preserving_2022}. SPAct (Dave et al., 2022) reported $\approx 10$--15\% utility loss from its self-supervised multi-attribute defense~\cite{dave_spact_2022}. STPrivacy (Li et al., 2023) lost $\approx 10$--20\% accuracy through frame- and video-level anonymization and introduced the VP-HMDB51 and VP-UCF101 benchmarks~\cite{li_stprivacy_2023}. MPPAR (Peng et al., 2023) improved generalization to retrained attackers at moderate cost through adaptive-attacker-aware meta-learning~\cite{peng_joint_2023}. The band's trade-off is 70--85\% utility at 60--70\% privacy improvement, which many deployments can accept.
\subsection{Poor-Efficiency Methods}
Formal guarantees and heavy encryption cost a lot of utility in current PPAR systems. Differential privacy (Luo et al., 2024) applied DP-SGD to video. It gave $\approx 76\%$ Top-1 at $\varepsilon = 5$ against a $\approx 98\%$ baseline, and $\approx 45\%$ at $\varepsilon = 1$~\cite{luo_differentially_2024}. Heavy encryption (Ishikawa et al., 2024; Zheng et al., 2024) uses cube-based and compressed-domain pixel encryption. It trades visual fidelity, and so recognition, for strong appearance protection~\cite{ishikawa_learnable_2024,zheng_fast_2024}. Gait and silhouette deformation (Hirose et al., 2022) preserves recognition for some tasks but gives only informal guarantees, so biometrics may leak under stronger attackers~\cite{hirose_anonymization_2022}. The trade-off here is 45--76\% utility for formally strong privacy, which is impractical unless utility tolerance is low.
\subsection{Trade-off Visualization and Elbow Analysis}\label{sec:elbow}
The three efficiency tiers separate cleanly on the Pareto frontier (Fig.~\ref{fig:pareto_frontier}), and each mechanism family shows a characteristic elbow. Skeleton-based methods show almost none, holding utility near baseline ($>95\%$) until privacy constraints turn extreme. Adversarial methods hit an elbow near 60--70\% privacy improvement, past which utility drops fast. Differential privacy shows the sharpest elbow, near $\varepsilon \approx 1$, where utility collapses. Most defenses can therefore reach about 60--70\% privacy improvement before utility falls off. Beyond that point, the choice of method should follow deployment constraints and acceptable utility loss.
\subsection{Proposed PPAR Unified Evaluation Protocol}
\begin{figure*}[h!]
\centering
\includegraphics[width=0.85\textwidth]{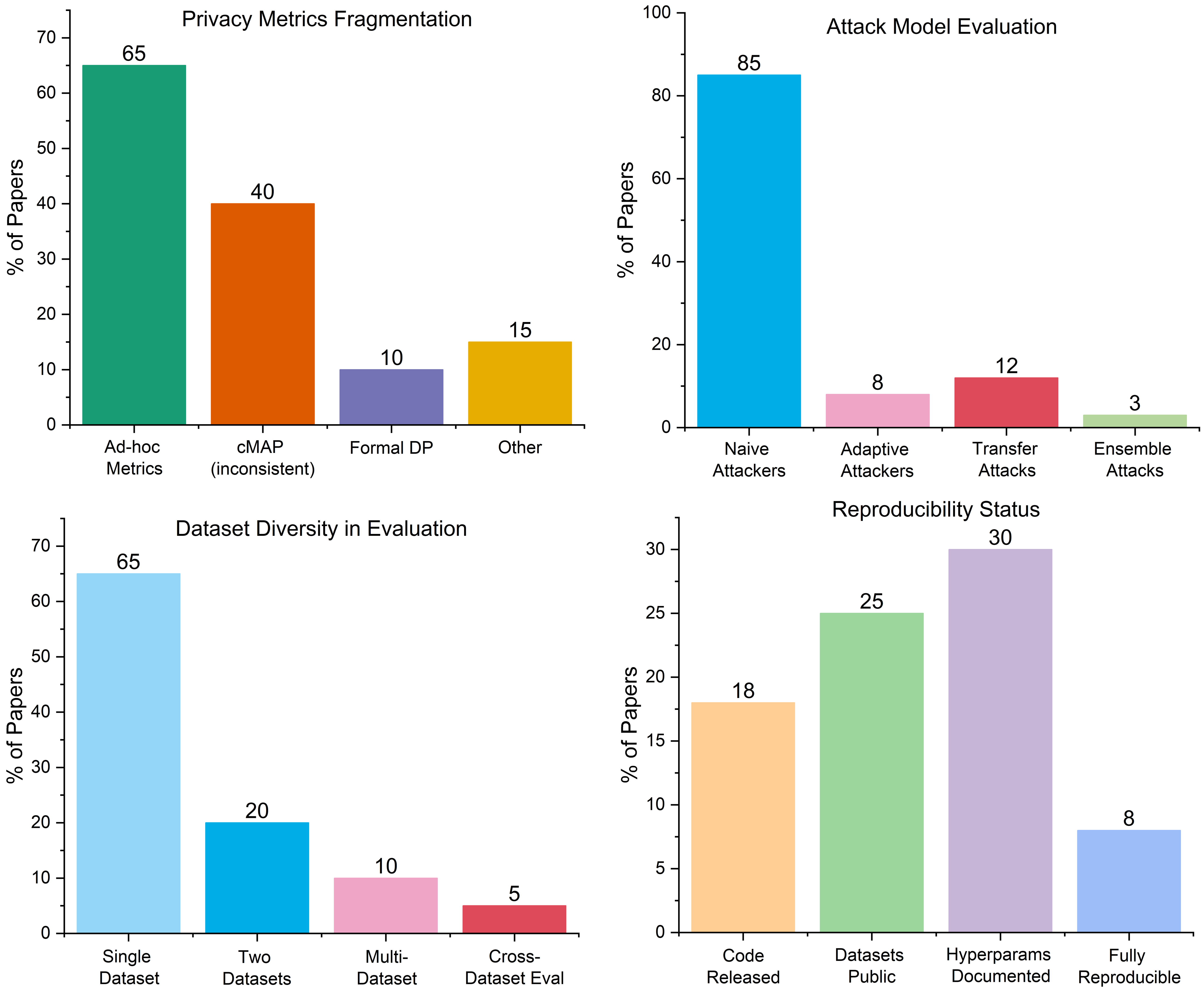}
\caption{Evaluation fragmentation across the corpus along four axes, namely privacy-metric choice, attack-model strength, dataset diversity, and reproducibility. Ad-hoc privacy metrics and naive attackers dominate, and few studies release code or are fully reproducible.}
\label{fig:evaluation_fragmentation}
\end{figure*}

Corpus analysis reveals severe fragmentation. No two of the reviewed papers use the same evaluation protocol, and only $\sim$40\% release code. Privacy claims are therefore evaluated under incomparable conditions, and cross-paper rankings are unreliable. Fig.~\ref{fig:evaluation_fragmentation} and Table \ref{tab:fragmentation_evidence} quantify the fragmentation across datasets, metrics, attack protocols, and reproducibility.

The protocol specifies five components, namely utility metrics (A), privacy metrics (B), attacker specification (C), dataset and reproducibility (D), and a phased adoption roadmap (E).
\subsubsection{A. Utility Metrics}
The primary utility metric is Top-1 accuracy on the official split. The secondary metrics are per-class accuracy, given as mean $\pm$ std, and accuracy degradation, defined as
$$\text{Accuracy Degradation (\%)} = \frac{A_{\text{baseline}} - A_{\text{privacy}}}{A_{\text{baseline}}} \times 100\%.$$
Optional metrics are mAP for localization and F1 for imbalanced tasks. Papers should report on PA-HMDB51 as the privacy-aware primary dataset, with UCF101 or HMDB51 as secondary~\cite{wu_privacy-preserving_2022,li_stprivacy_2023,dave_spact_2022}. Results should use official splits, give confidence intervals over multiple runs, and state the backbone and pretraining.
\subsubsection{B. Privacy Metrics}
Two privacy metrics are recommended. The first is cMAP-based attribute prediction. It needs a fixed reference attacker, such as a 2-layer MLP on features or a ResNet-18 on frames. The training regimen (optimizer, learning rate, epochs) and a fixed attacker split such as 50/50 should be stated. Papers should report the average cMAP across protected attributes, including identity, gender, age, and clothing. Three thresholds apply, namely strict (cMAP $< 0.2$), moderate (cMAP $< 0.4$), and loose (cMAP $< 0.6$). The second metric is differential privacy. Papers should report $\varepsilon$ at $\delta = 10^{-5}$ or a curve across $\varepsilon$, state the mechanism (Gaussian or Laplace) and sampling strategy, and give the utility-privacy curve.
\subsubsection{C. Attack Model Specification}
The attacker model should be specified along four dimensions. Attacker access is white-box or black-box. Attacker knowledge is naive, adaptive, or transfer. Attacker training data is supervised or unsupervised. The implementation should state the classifier architecture, the hyperparameters (learning rate, batch size, epochs), the train/test split, and the random seeds.
\subsubsection{D. Dataset \& Reproducibility}
Three items support reproducibility. Papers should state the dataset version and split, such as PA-HMDB51 official split 1, together with the protected attributes and dataset sizes. They should release code, pretrained weights, preprocessing scripts, and attacker implementations, plus a docker or conda environment. Finally, they should compare against at least three existing PPAR methods on the same splits and backbones.
\subsubsection{E. Phased Implementation Strategy}
The framework can be adopted in three phases. Phase 1, minimal compliance, reports Top-1 and degradation percentage, one primary privacy metric (cMAP or $\varepsilon$-DP), the threat model, and released code and splits. Phase 2, full compliance, adds per-class metrics, further privacy metrics such as transfer and membership, several attacker scenarios, and Pareto ranking. Phase 3, the standard, adds full multi-metric utility, formal privacy metrics, an exhaustive attacker suite covering white- and black-box, adaptive, and transfer attackers, and cross-dataset generalization.
\begin{figure*}
\centering
\includegraphics[width=0.75\textwidth]{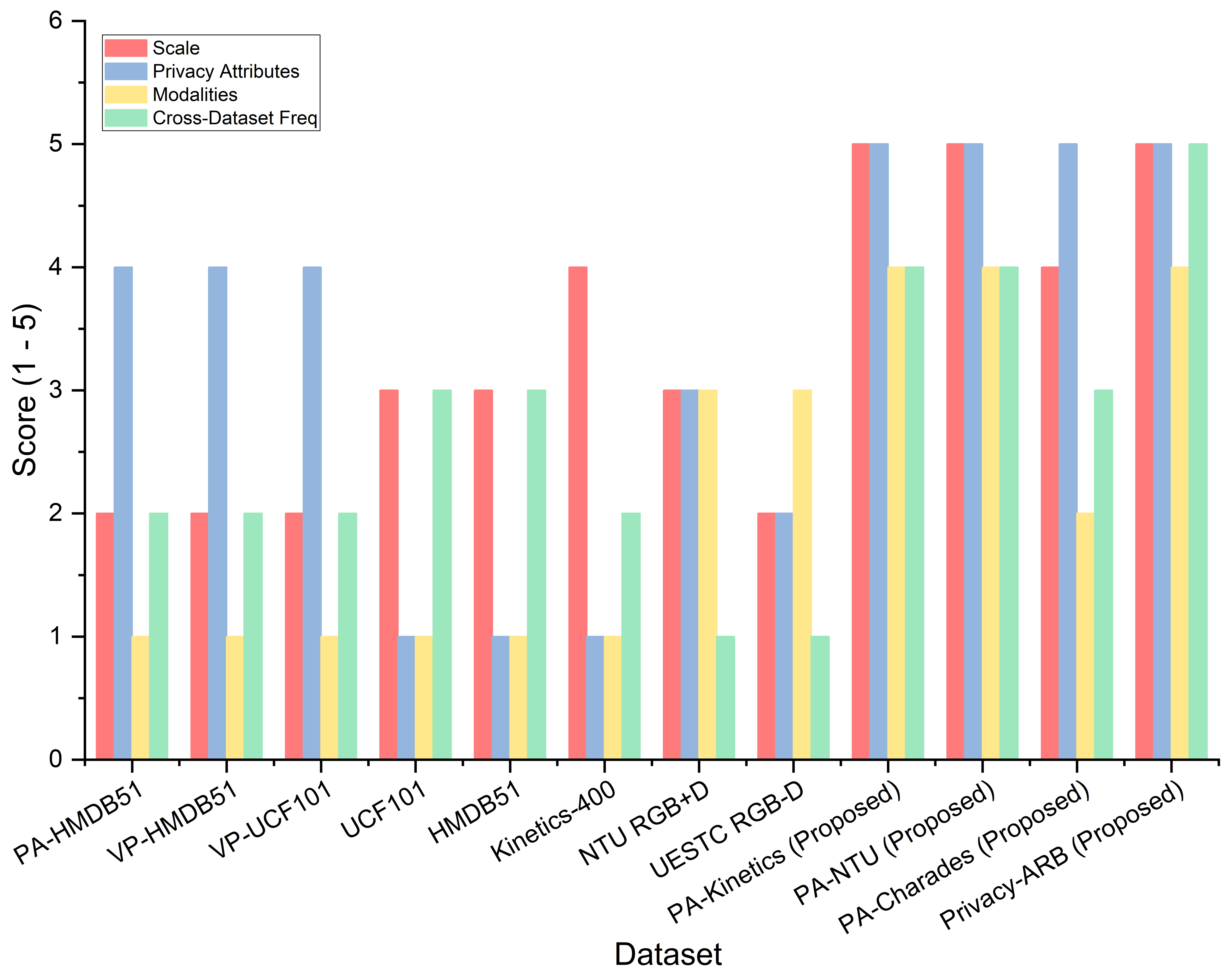}
\caption{Dataset landscape for PPAR, showing existing and proposed action-recognition benchmarks rated 1--5 on scale, privacy-attribute coverage, modality coverage, and cross-dataset usage. The ratings expose the field's concentration on a few utility datasets and the scarcity of privacy-aware, multi-modal benchmarks, which the proposed benchmarks aim to fill.}
\label{fig:dataset_landscape}
\end{figure*}
\section{Datasets and Benchmarks}
Dataset choice sits at the center of PPAR. Measured utility and perceived privacy both depend on the data used, its annotation, and the attacker models applied. Standard action-recognition benchmarks are needed to measure task utility but cannot prove privacy claims, since most were not built to quantify attribute leakage or identity-disclosure risk. This section separates general-purpose utility datasets from privacy-aware benchmarks, reviews the datasets that shaped PPAR, and points to the central gap. No universally accepted, attacker-aware privacy benchmark exists.
\subsection{Existing Datasets}
\subsubsection{Standard Action-Recognition Datasets}
UCF101, HMDB51, and Kinetics, with their variants, remain the utility backbone of PPAR work. They supply canonical splits, diverse action classes, and established protocols for comparing action accuracy. They ship without privacy labels, because they were built for discriminative power and temporal diversity rather than for measuring identity leakage. When authors rely on them alone, privacy is evaluated post-hoc with external attribute classifiers. These heterogeneous attacker designs complicate cross-paper comparison.
\subsubsection{Modality- and Deployment-Specific Datasets}
A second family targets modalities or settings that are private by design or deployment-relevant. These include skeleton datasets such as NTU and Kinetics-skeletons, depth and RGB-D, gait, assisted-living, and surveillance video. Depth sensors and pose estimators already reduce appearance leakage, so skeleton- and depth-based methods can make stronger utility-preserving privacy claims, though under sensor and algorithmic constraints. Each domain also brings its own risks. Assisted-living data emphasizes continuous monitoring, gait data overlaps with biometric identity, and surveillance data adds crowd and occlusion challenges. Representative works include Moon et al.\ and Gao et al.\ on skeleton anonymization~\cite{moon_anonymization_2023,gao_privacy-preserving_2025} and Jain et al.\ on assisted-living evaluation~\cite{jain_privacy-preserving_2024}.
\subsubsection{Corpus-Grounded Examples and How Dataset Choice Shapes Claims}
Dataset selection shapes each paper's claims and limitations. At the acquisition level, Wang et al.\ used coded-aperture input to weaken appearance cues~\cite{wang_privacy-preserving_2019}, and Liu et al.\ embedded defenses in compressed-sensing pipelines~\cite{liu_video_2021}. On standard data, Tomei et al.\ measured face-obfuscation effects on Kinetics-derived clips~\cite{tomei_estimating_2021}, and Kumawat and Nagahara applied motion-difference quantization to keep motion while removing appearance~\cite{kumawat_privacy-preserving_2022}. Toward privacy-aware benchmarking, Wu et al.\ and related adversarial methods used PA-HMDB51~\cite{wu_privacy-preserving_2022}, while SPAct and STPrivacy added privacy labels and test protocols~\cite{dave_spact_2022,li_stprivacy_2023}.
\subsection{Privacy-Aware Datasets}
Standard benchmarks lack the labels and protocols needed for privacy evaluation, so a distinct class of privacy-aware datasets has emerged.
\subsubsection{What We Mean by Privacy-Aware Datasets}
A privacy-aware dataset does at least one of three things. It carries ground-truth privacy labels such as identity or attributes. It provides anonymized variants alongside the originals. It may also ship with a benchmark protocol that defines attacker models and reporting formats. These features are essential, since privacy evaluation needs ground truth for both utility and leakage, and attacks must be reproducible across papers.
\subsubsection{Canonical Privacy-Aware Datasets in the Corpus}
Canonical privacy-aware resources include PA-HMDB51, an adapted HMDB51 split used by many adversarial and anonymization methods~\cite{wu_privacy-preserving_2022}. STPrivacy introduced the VP-HMDB51 and VP-UCF101 variants for spatio-temporal anonymization and cross-dataset tests~\cite{li_stprivacy_2023}. Task- and modality-specific collections for skeleton, gait, and assisted-living carry identity or attribute labels~\cite{moon_anonymization_2023,gao_privacy-preserving_2025,jain_privacy-preserving_2024}. Building these benchmarks has reshaped evaluation practice. SPAct added self-supervised protocols and privacy labels, which allow evaluation without full attribute annotation~\cite{dave_spact_2022}. STPrivacy's VP variants eased cross-method and transfer-aware comparison~\cite{li_stprivacy_2023}. MPPAR and TeD-SPAD showed that generalization- and anomaly-aware benchmark design is feasible and necessary for deployment~\cite{peng_joint_2023,fioresi_ted-spad_2023}.
\subsection{Limitations and Design Principles for Better Benchmarks}
Despite this progress, privacy-aware datasets are still few, mostly single-paper, and narrow in attribute coverage. They rarely standardize inversion, membership, or adaptive-attacker protocols. A practical benchmark should provide several things. It needs canonical train, validation, and test splits to prevent cherry-picking. It should cover multiple privacy attributes such as identity, face visibility, gender, age, and clothing, with documented annotation protocols. It should define an attacker suite that spans naive classifiers, retraining or adaptive attackers, inversion or membership attacks, and transfer attackers. It should include cross-domain or cross-dataset splits to measure generalization. Finally, it should ship reproducible preprocessing scripts and baseline attacker implementations with fixed seeds. Fig.~\ref{fig:dataset_landscape} rates existing and proposed datasets on scale, privacy attributes, modality coverage, and cross-dataset usage, which exposes the concentration and imbalance across the 32-paper corpus.
\begin{figure}[h!]
\centering
\includegraphics[width=0.48\textwidth]{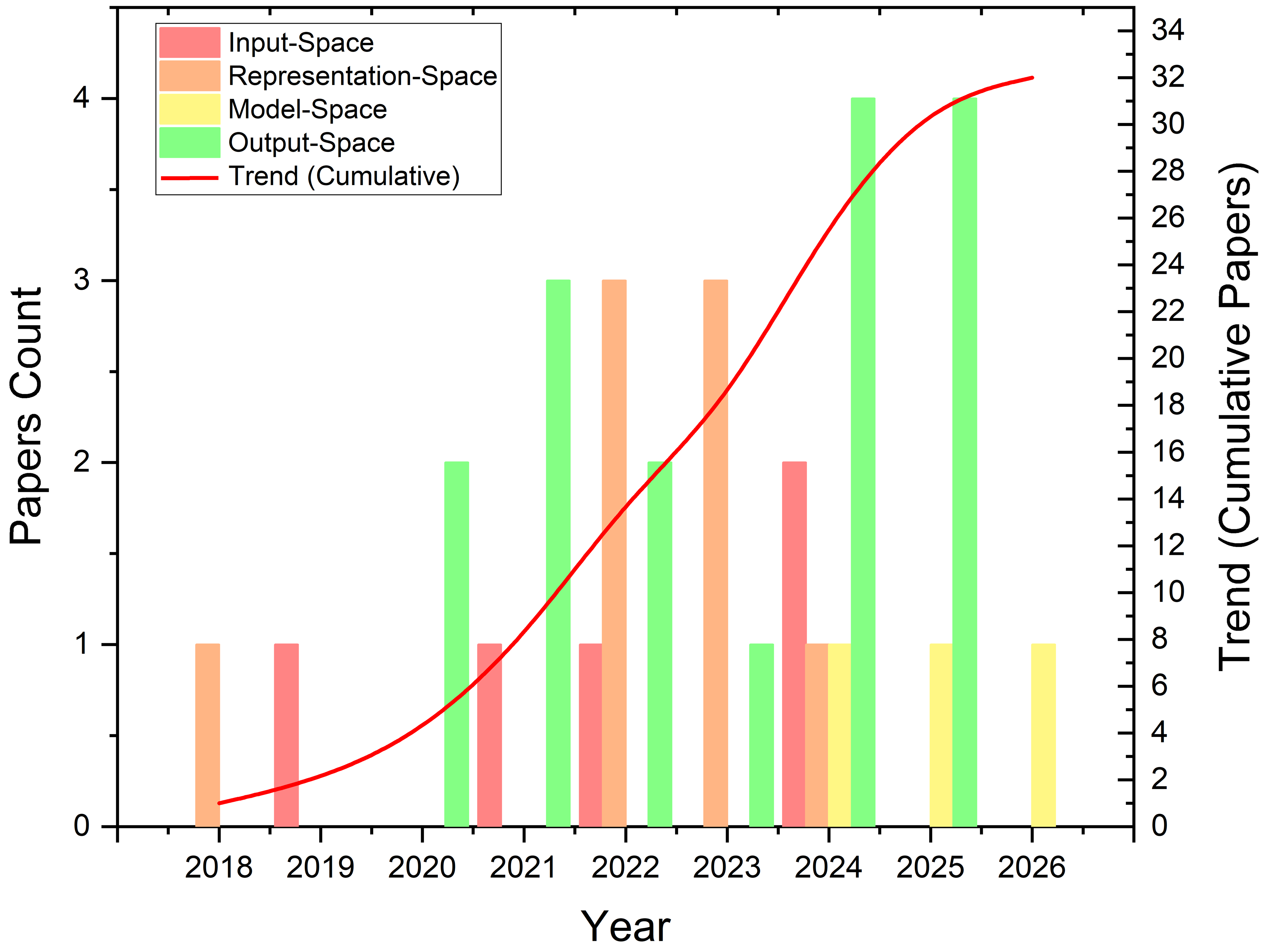}
\caption{Temporal evolution of PPAR research from 2018 to 2026, showing yearly publications by privacy injection space and the cumulative trend, with a post-2023 shift toward formal-privacy and task-specific methods.}
\label{fig:temporal_evolution}
\end{figure}
\subsection{The Critical Gap}
PPAR still lacks a benchmark suite that rivals ImageNet or Kinetics in scope and adoption. The dataset-level fragmentation quantified in Table \ref{tab:fragmentation_evidence} blocks cross-paper synthesis and delays consolidation. Two consequences follow. First, evaluation settings are inconsistent. Some papers add a single attribute classifier to utility-only metrics on UCF or HMDB, while others use PA-HMDB51 or VP-* splits with richer protocols, so equal-footing comparison is impossible. Second, generalization is underexplored. Many defenses are validated on one dataset, so cross-dataset transfer and robustness to unseen attackers stay underreported, which risks brittle deployment. The field needs a community-driven benchmark. It should combine standard utility datasets such as UCF101, HMDB51, and Kinetics with privacy-aware variants such as PA-HMDB, VP-HMDB51, and VP-UCF101. It should add an open-source attacker suite covering naive, adaptive, inversion, and membership attacks, together with a clear reporting template. SPAct and STPrivacy are stepping stones, and MPPAR shows how generalization-aware evaluation can work. What remains is a coordinated effort to adopt a shared benchmark, release reproducible attacker baselines, and require a minimal privacy-evaluation checklist.
\section{Comparative Analysis}
\begin{figure*}[h!]
\centering
\includegraphics[width=0.9\textwidth]{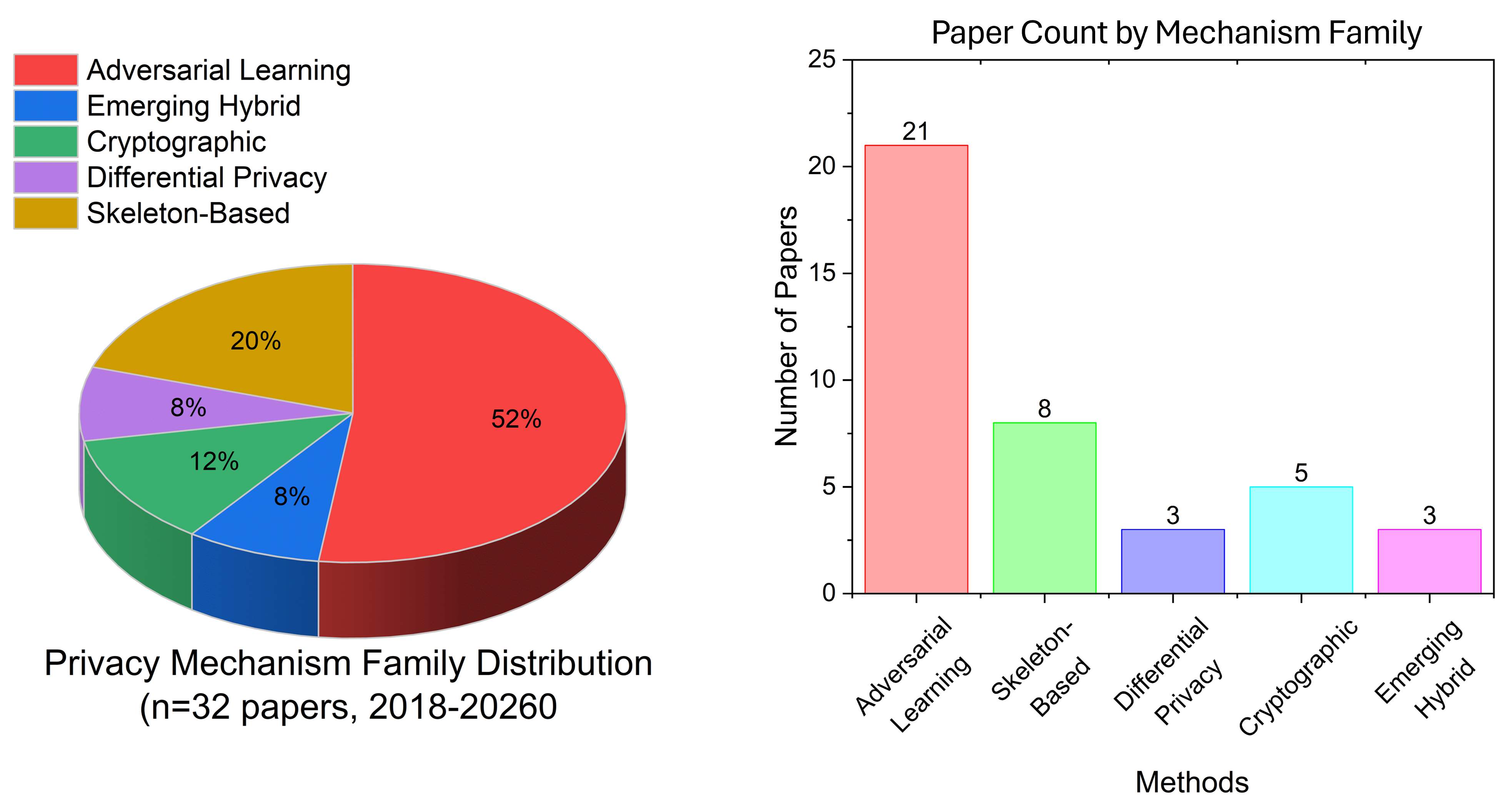}
\caption{Distribution of the surveyed papers by privacy-mechanism family. Adversarial learning dominates, followed by skeleton-based methods, while cryptographic, differential-privacy, and emerging-hybrid approaches remain comparatively rare.}
\label{fig:mechanism_distribution}
\end{figure*}

This section compares the PPAR method families on the axes that matter for research and deployment, namely task utility, privacy in both empirical and formal forms, and robustness to realistic attackers and dataset shifts. Across the 32 papers it asks which families preserve utility best, which give the strongest privacy claims, and which stay robust under realistic attack and transfer settings. The comparison is organized by mechanism family rather than by individual paper, which exposes structural trade-offs and recurring evaluation gaps. Representative works anchor the claims (Wu et al.\ 2018, 2022; Wang et al.\ 2019; Dave et al.\ 2022; Li et al.\ 2023; Luo et al.\ 2024)~\cite{wu_towards_2018,wu_privacy-preserving_2022,wang_privacy-preserving_2019,dave_spact_2022,li_stprivacy_2023,luo_differentially_2024}. Two figures support the analysis. Fig.~\ref{fig:mechanism_distribution} groups the 32 methods by mechanism family, and Fig.~\ref{fig:temporal_evolution} traces their emergence over 2018--2026.
\subsection{Quantitative Method-Category View}
\Cref{tab:comparative_summary} summarizes representative work by method category. It shows consistent utility advantages for modality-based methods and recurring evaluation gaps.
\subsubsection{Dominant Families and Numerical Dominance}
Representation-space adversarial methods and skeleton or modality-specific methods together account for most included papers (Fig.~\ref{fig:mechanism_distribution}). Adversarial learning emerged early and spawned many descendants (Wu et al.\ 2018, 2022; Zou et al.)~\cite{wu_towards_2018,wu_privacy-preserving_2022,zou_privacy-preserving_2022}. Skeleton and modality approaches (Moon et al.\ 2023; Gao et al.\ 2025; Jain et al.\ 2024) use structured signals to avoid appearance leakage~\cite{moon_anonymization_2023,gao_privacy-preserving_2025,jain_privacy-preserving_2024}. Input-, acquisition-, and compressed-domain techniques are rarer but influential for edge efficiency~\cite{wang_privacy-preserving_2019,zheng_fast_2024,liu_video_2021}.
\subsubsection{Metrics in Use and Their Limits}
Top-1 action accuracy is still the default utility metric, and few papers report per-class, transfer, or calibration measures. Privacy metrics are more fragmented. cMAP, attribute-classifier accuracy, and face-recognition rates dominate, while formal $\varepsilon$-DP appears in only a few works (Luo et al.\ 2024; Feng et al.\ Priva 2025)~\cite{luo_differentially_2024,feng_priva_2025}. This creates a reporting asymmetry. Utility numbers compare across papers, but privacy guarantees compare only qualitatively.
\subsubsection{Attack Modeling Breadth and Adaptivity}
Attacker models follow three habits. Naive attribute classifiers are trained and tested on the same sanitized distribution. Simple retraining attackers adapt to transformed inputs. Specialized inversion, membership, or differential attacks appear in a few works, notably Luo et al.\ 2024~\cite{luo_differentially_2024}. Adaptive attackers, who know the defense and adapt their architecture or loss, stay underused, which inflates apparent privacy.
\subsubsection{Utility-Privacy Trade-offs by Family}
Skeleton- and modality-based methods keep the best action utility while driving appearance attributes near chance (Gao et al.\ 2025; Moon et al.\ 2023)~\cite{gao_privacy-preserving_2025,moon_anonymization_2023}. Representation-adversarial methods are a practical middle ground, but their privacy strength depends heavily on attacker assumptions. Input-space transforms such as coded aperture, compressed sensing, and block shuffles preserve motion, yet can leave residual structure that stronger attackers exploit. Encryption and compressed-domain defenses promise strong privacy but are unevenly evaluated and often omit inversion tests, which leaves cross-paper claims fragile.
\subsubsection{Reporting Practices and Evaluation Rigor}
Four evaluation weaknesses recur. Most papers lean on Top-1 accuracy and a single privacy proxy such as cMAP or one attribute classifier. Transfer and cross-dataset evaluation is sparse, and where it is reported it reveals more leakage than within-dataset tests. Adaptive attackers, inversion, and membership-inference tests are rarely used. Few papers pair formal privacy such as DP with action benchmarks at scale. Those that do, such as Luo et al.\ 2024, show a large utility cost~\cite{luo_differentially_2024}.
\subsubsection{Separation Between Modalities and Pixel-Level Methods}
The table reveals a structural separation. Modality-changing methods such as skeleton, depth, and optical-flow-only gain clear privacy-utility advantages, because identity cues are inherently suppressed. Pixel-level adversarial or obfuscation methods face a harder task. They must remove identity while keeping the fine appearance signals that separate similar actions. That tougher trade-off explains their lower utility on the same tasks.
\subsubsection{Practical Conclusions from the Table}
Where sensors permit, a skeleton or modality pipeline is the best first choice for utility-preserving privacy. For RGB streams, adversarial-representation and compressed-domain defenses are promising but need validation against adaptive attackers and cross-dataset transfer. Formal privacy is a separate axis that offers provable guarantees at reduced utility.
\subsection{Trend Analysis}
The field's temporal evolution shows how these trade-offs and evaluation practices took shape.
\subsubsection{Temporal Evolution of the Field}
The field evolved in four phases (Fig.~\ref{fig:temporal_evolution}). Pre-2019 work, not included here, explored heuristic obfuscation and masking~\cite{dai_towards_2015,qi_privacy_2016,salama_multi-level_2017}. From 2018 to 2020, adversarial learning and learned anonymization rose (Wu et al.\ 2018)~\cite{wu_towards_2018}. From 2020 to 2022, papers added stronger empirical evaluation and modality-aware approaches (Moon et al.\ 2023; Dave et al.\ SPAct 2022)~\cite{moon_anonymization_2023,dave_spact_2022}. From 2023, three tendencies converged. These were edge-viable compressed-domain designs (Zheng et al.\ 2024; Liu et al.\ 2021), privacy-aware benchmarks such as STPrivacy with VP-HMDB51 and VP-UCF101, and formal privacy with reversible analytics (Luo et al.\ 2024; Feng et al.\ Priva 2025)~\cite{zheng_fast_2024,liu_video_2021,li_stprivacy_2023,luo_differentially_2024,feng_priva_2025}. The trajectory moves from heuristic pixel-space fixes to principled representation and systems designs.
\subsubsection{Shift in Privacy Metrics}
cMAP and attribute-classifier accuracy dominate for their simplicity, but their meaning depends on the attacker family and dataset splits. No more composable, deployment-oriented standard has emerged yet. Differential privacy and membership inference are growing but stay specialized, held back by utility cost and complexity.
\subsubsection{Attacker Modeling Trends}
Early work tested privacy against the same classifiers used in training, which gave optimistic results. Recent papers add retraining or transfer attacks, and a growing subset tries inversion or membership attacks. Adaptive attacker modeling stays rare. A defense that stops naive classifiers may fail against adaptive adversaries.
\subsubsection{Dataset and Benchmark Shifts}
HMDB51 and UCF101 remain the common utility benchmarks, while privacy-aware variants such as PA-HMDB, VP-HMDB51, and VP-UCF101 are starting to standardize privacy evaluation~\cite{li_stprivacy_2023}. Adoption is partial. Many papers still use original datasets with ad-hoc attribute annotations, which complicates cross-paper comparison. Cross-dataset generalization tests are increasingly seen as necessary but are not yet standard.
\subsubsection{Representation Design Evolution}
Designs have moved from raw pixel obfuscation such as pixelation and blurring toward structured representations such as skeletons, depth, motion differences, and disentangled latents. Action recognition is mostly motion-based, so representations that keep motion while dropping appearance give better privacy-utility trade-offs. These approaches can be sensor- or environment-dependent, and they may need separate pipelines for unconstrained scenes.
\subsubsection{Critical Observations on Field Trajectory}
The field shows methodological breadth but evaluation fragmentation, and architectural progress outpaces convergent evaluation standards. More papers now include a transfer or retraining attacker, but a standard evaluation suite covering per-class utility, transfer, adaptive, and inversion or membership attacks remains rare.
\subsection{Key Observations}
Five observations distill the mechanism-family and temporal evidence into the field's core findings.
\subsubsection{Observation 1}
Utility-preserving privacy is easiest when the representation suppresses identity cues. Skeleton and depth pipelines (Moon et al.\ 2023; Gao et al.\ 2025; Jain et al.\ 2024) consistently keep action accuracy while reducing attribute leakage~\cite{moon_anonymization_2023,gao_privacy-preserving_2025,jain_privacy-preserving_2024}. They collapse the appearance manifold but preserve the kinematics that action classification needs, at the cost of sensor dependency.
\subsubsection{Observation 2}
Adversarial and anonymization papers often report privacy as a drop in attribute-classifier mAP (Wu et al.\ 2022; Dave et al.\ SPAct 2022; Li et al.\ STPrivacy 2023)~\cite{wu_privacy-preserving_2022,dave_spact_2022,li_stprivacy_2023}. However, cMAP couples tightly to the attacker family and split, so defenses that overfit to a narrow attacker look stronger than they are under general threat models (\cref{sec:challenges}).
\subsubsection{Observation 3}
Luo et al.\ 2024 shows differential privacy works for video but at a noticeable utility cost~\cite{luo_differentially_2024}. Feng et al.\ Priva 2025 offers a system-level alternative through reversible analytics and controlled access, though it assumes stronger trust and more operational complexity~\cite{feng_priva_2025}.
\subsubsection{Observation 4}
Most papers evaluate privacy against only one or two attacker variants, so membership inference, model inversion, and adaptive attackers stay underrepresented. Methods that are strong against naive classifiers may fail badly under these threats (\cref{sec:challenges}).
\subsubsection{Observation 5}
Some papers report high accuracy but cursory privacy evaluation. Others showcase anonymization yet never test utility or realistic deployment. High-impact work balances both axes and reports across strong, adaptive attackers and transfer settings. Taken together, the literature is method-rich but evaluation-poor, and representation-learning methods have matured faster than evaluation standards. The most promising near-term direction is not simply stronger privacy algorithms, but standardized, adversary-aware evaluations that give calibrated privacy-utility trade-offs under realistic assumptions.
\renewcommand{\arraystretch}{1.4}
\sloppy
\begin{table*}[t]
\centering
\scriptsize
\setlength{\tabcolsep}{3pt}
\caption{Comparative summary of the PPAR literature by method category, listing representative papers, datasets, utility and privacy metrics, attacker models, and the main empirical result for each of the six categories. Skeleton- and modality-based methods achieve the strongest utility-privacy trade-offs, whereas formal-privacy and encryption categories offer stronger or provable guarantees at higher utility cost.}
\label{tab:comparative_summary}
\begin{tabularx}{\textwidth}{@{}
    C{0.3cm}   
    L{1.9cm}   
    L{2.6cm}   
    C{0.7cm}   
    L{1.9cm}   
    L{1.3cm}   
    L{1.4cm}   
    L{2.0cm}   
    >{\arraybackslash}X
@{}}
\toprule
\textbf{\#} & \textbf{Method Category} & \textbf{Representative Papers} & \textbf{Year} & \textbf{Dataset(s)} & \textbf{Utility Metric(s)} & \textbf{Privacy Metric(s)} & \textbf{Attack Model(s)} & \textbf{Main Empirical Result} \\
\midrule
1 & Representation-space adversarial & Wu et al., Zou et al., Dave et al.\ (SPAct)~\cite{wu_towards_2018,wu_privacy-preserving_2022,zou_privacy-preserving_2022,dave_spact_2022} & 2018--2022 & PA-HMDB51, UCF101, HMDB51 & Top-1 accuracy, F1 & cMAP, attribute classifier & Adversarial attribute classifiers, naive retraining & Adversarial training yields competitive utility with reduced attribute leakage but assumes non-adaptive attackers \\ \hline
2 & Input-space / acquisition-space & Wang et al., Liu et al., Ishikawa et al.~\cite{wang_privacy-preserving_2019,liu_video_2021,ishikawa_learnable_2024} & 2019--2024 & UCF101, HMDB51, custom & Top-1 accuracy, mAP & cMAP, attribute classifiers & Attribute inference, simple classifiers & Acquisition transforms retain motion while removing identity cues, though they are often tested against weak attackers \\ \hline
3 & Compression / encryption & Zheng et al., Liu et al., Maximov (CIAGAN)~\cite{zheng_fast_2024,liu_video_2021,maximov_ciagan_2020} & 2020--2024 & Compressed datasets, I-frame tests, face & Top-1 on compressed domain & cMAP, face recognition rate & Reconstruction, frame-level classifiers & Compressed-domain defenses are promising for edge deployment, though evaluations often omit strong inversion attempts \\ \hline
4 & Output-space / deployment & Zhou et al., Tomei et al., Ilic et al., Hellmann et al.~\cite{zhou_privacy-sensitive_2020,tomei_estimating_2021,ilic_selective_2024,hellmann_ganonymization_2024} & 2020--2024 & Kinetics subsets, PA-HMDB & Top-1, per-class accuracy & Recognition rate, attribute classifiers & Real-time attackers, object trackers & Output-space masking is operationally simple but risks collateral damage, though selective approaches reduce it \\ \hline
5 & Skeleton / modality-specific & Moon et al., Gao et al., Jain et al., Hirose et al.~\cite{moon_anonymization_2023,gao_privacy-preserving_2025,jain_privacy-preserving_2024,hirose_anonymization_2022} & 2022--2025 & Skeleton datasets, NTU, HMDB51 subsets & Top-1, per-class accuracy & Attribute classifier on skeletons & Attribute inference adapted to skeletons & Modality filtering preserves motion utility well and greatly reduces appearance leakage, but is limited when skeleton extraction fails \\ \hline
6 & Model-space / formal privacy & Luo et al., Feng et al.\ (Priva)~\cite{luo_differentially_2024,feng_priva_2025} & 2024--2025 & CIFAR variants, task-specific video & Accuracy under DP-fine-tuning & $\varepsilon$-DP, membership leakage & White-box DP attacker models, inversion & Differential privacy gives formal guarantees but at notable utility cost, and system models trade accuracy for provable guarantees \\
\bottomrule
\end{tabularx}
\end{table*}
\begin{figure*}
\centering
\includegraphics[width=0.9\textwidth]{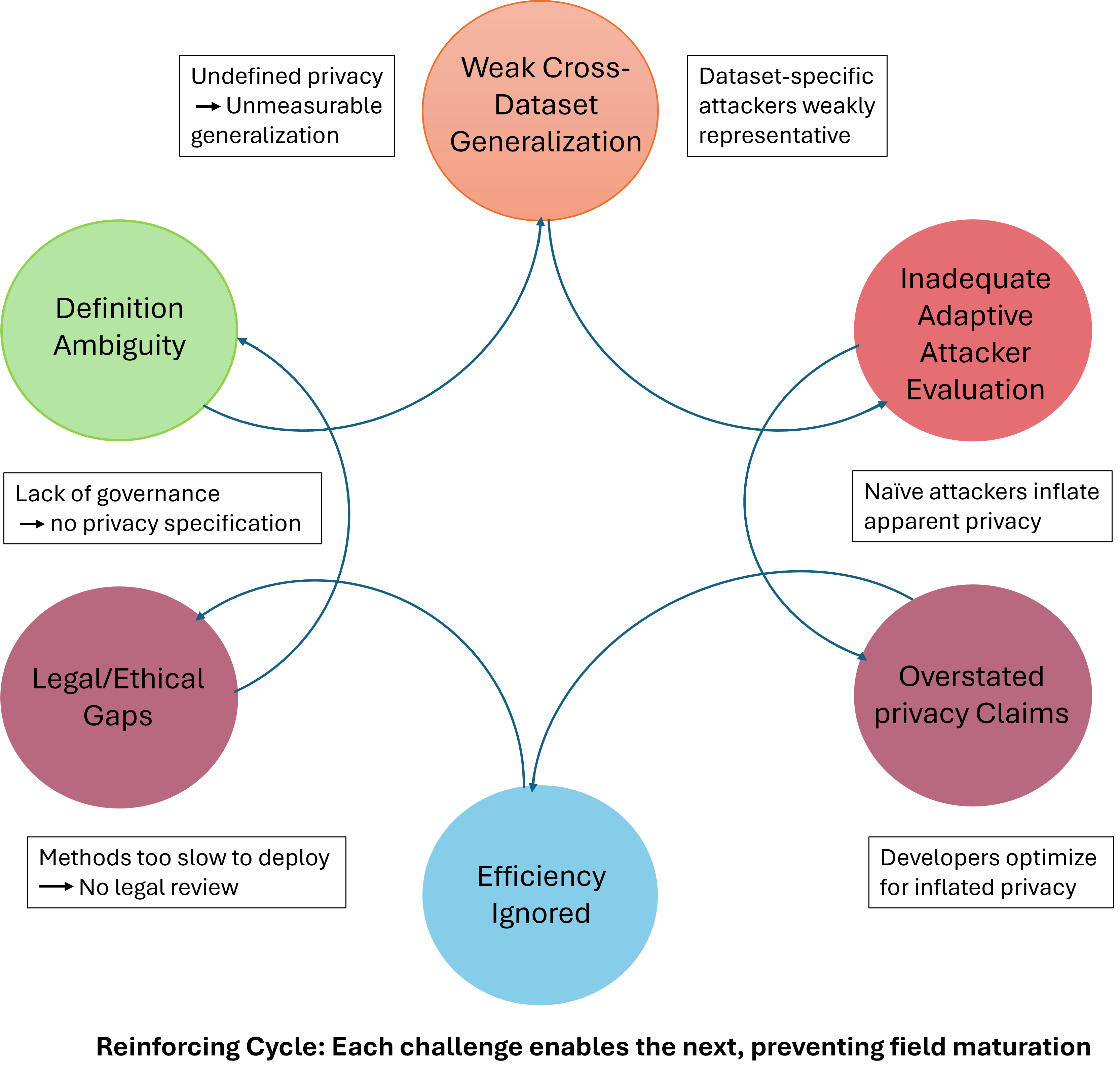}
\caption{Interconnections among the open challenges in PPAR. Definition ambiguity and weak evaluation mutually reinforce one another, so field maturation requires coordinated standardization rather than isolated fixes.}
\label{fig:challenge_interconnection}
\end{figure*}
\section{Open Challenges}\label{sec:challenges}
Five interconnected barriers hold back the maturation of PPAR, namely definitional ambiguity, weak cross-dataset generalization, inadequate adaptive-attacker evaluation, neglected real-time constraints, and unaddressed legal and governance obligations. We discuss each in turn, draw evidence from the corpus, pair each barrier with the standardization it calls for, and then show how the five reinforce one another.
\subsection{No Unified Privacy Definition}
The most basic obstacle is that the field has no shared meaning for the word ``privacy.'' At least six distinct notions coexist in the corpus, and papers often use them interchangeably, which makes headline claims incomparable. The first is empirical attribute suppression, the reduced success of attribute classifiers such as gender or age, measured by cMAP or an accuracy drop~\cite{wu_towards_2018,wu_privacy-preserving_2022}. The second is anonymization and identity removal, the reduced success of identity or re-identification attacks, as in CIAGAN, Hirose et al., Moon et al., and Hellmann et al.~\cite{maximov_ciagan_2020,hirose_anonymization_2022,moon_anonymization_2023,hellmann_ganonymization_2024}. The third is formal differential privacy, an $(\varepsilon, \delta)$-DP guarantee that holds for any attacker, model, and time horizon, applied to video action recognition through DP-SGD by Luo et al.~\cite{luo_differentially_2024}. The fourth is task-conditioned empirical privacy, defined relative to specific attacks, datasets, and attacker models, as in SPAct, STPrivacy, MPPAR, and TeD-SPAD~\cite{dave_spact_2022,li_stprivacy_2023,peng_joint_2023,fioresi_ted-spad_2023}. The fifth is encryption and confidentiality, resistance to signal recovery without the key, used by Wang et al., Ishikawa et al., Zheng et al., and Feng et al.~\cite{wang_privacy-preserving_2019,ishikawa_learnable_2024,zheng_fast_2024,feng_priva_2025}. The sixth is reconstruction and feature-inversion resistance, the inability to invert learned representations back to identity-bearing data, which is usually assumed rather than tested. These notions are not equivalent, so conflating them yields claims that cannot be compared. A defense that suppresses an attribute classifier is not thereby differentially private, and a system that resists reconstruction may still leak demographics. Every paper should therefore state its privacy target (identity, demographics, or appearance), the attacker model and its resources, the metric used such as cMAP or $\varepsilon$-DP, and whether the guarantee is empirical or formal. Claims should also sit on an explicit strength hierarchy, from formal DP through encryption and anonymization down to attribute suppression, so readers compare guarantees rather than vocabulary.
\subsection{Weak Cross-Dataset Generalization}
Even where privacy is defined consistently, the evidence for it is thin. Fewer than 15\% of the surveyed papers test cross-dataset generalization, and most evaluate on a single benchmark against one attack and one attribute. Reported robustness is therefore benchmark-dependent, and claims rarely transfer to data the method was not tuned on. Credible evaluation should require cross-dataset testing on at least two benchmarks, ideally privacy-aware expansions such as PA-Kinetics and PA-NTU. It should also include multi-attribute evaluation and transfer-attack testing that probes robustness against attributes and attacks unseen during training.
\subsection{Inadequate Adaptive-Attacker Evaluation}
A related weakness concerns the adversary. Fewer than 10\% of papers evaluate adaptive attackers, and most report results only against naive attackers with no knowledge of the defense. Wu et al.\ 2022, SPAct, STPrivacy, and MPPAR are among the rare exceptions~\cite{wu_privacy-preserving_2022,dave_spact_2022,li_stprivacy_2023,peng_joint_2023}. A defense that is strong against a naive classifier can fail badly once the attacker adapts to the mechanism. Attacker evaluation should therefore span both naive and mechanism-aware adaptive adversaries, with transparent reporting of the attacker's access, training data, and knowledge of the defense. It should also include transfer and ensemble attacks that stress the defense beyond a single fixed adversary.

\subsection{Real-Time Constraints Inadequately Addressed}
The practical target for much of PPAR is edge deployment, yet real-time feasibility is rarely shown. Operating budgets of 30+ FPS, sub-100 ms latency, and constrained memory and power are seldom reported. Wang et al.\ 2019, Zheng et al.\ 2024, and EAST (Zhicai et al.\ 2025, 37 FPS on-device) are among the few exceptions~\cite{wang_privacy-preserving_2019,zheng_fast_2024,zhicai_east_2025}. This silence matters, because privacy mechanisms add real overhead. Adversarial training, multi-attribute classifiers, multi-stage pipelines, and skeleton extraction each cost 5 to 50 ms per frame. Papers should report efficiency metrics such as FPS, latency, memory, and parameter count on both standard and resource-constrained hardware. They should state the deployment target (cloud, edge, or mobile) and characterize the efficiency-utility-privacy trade-off through a tri-objective Pareto analysis rather than optimizing privacy alone.
\subsection{Legal, Ethical, and Governance Gaps}
PPAR is deployed in high-stakes settings such as surveillance and assisted living, yet legal and ethical obligations are largely ignored. Jain et al.\ 2024 and Hellmann et al.\ 2024 are rare exceptions~\cite{jain_privacy-preserving_2024,hellmann_ganonymization_2024}. A 70\% improvement in cMAP does not satisfy GDPR if consent was never obtained, and technical privacy gains cannot substitute for governance. Papers that make deployment claims should document IRB approval, consent, and data retention and deletion policies. They should establish GDPR or CCPA compliance, record privacy-by-design decisions, and provide user-facing disclosures that quantify the residual privacy risk.
\subsection{Interconnections and Field-Level Implications}
These five challenges are not independent, and each reinforces the next. Definitional ambiguity lets weak generalization pass unnoticed, which in turn hides inadequate attacker evaluation, so overstated privacy claims accumulate while efficiency is ignored and legal deployment stays infeasible. Fig.~\ref{fig:challenge_interconnection} maps this network of dependencies. The field cannot mature through isolated algorithmic fixes. Because the barriers are coupled, progress depends on coordinated standardization across all five, namely a privacy-definition consensus, a shared benchmark suite with cross-dataset evaluation, standardized naive, adaptive, and transfer attacker models, mandatory efficiency reporting, and governance-aware protocols. PPAR will advance only when scientific rigor is treated as seriously as algorithmic novelty.
\begin{figure*}
\centering
\includegraphics[width=0.99\textwidth]{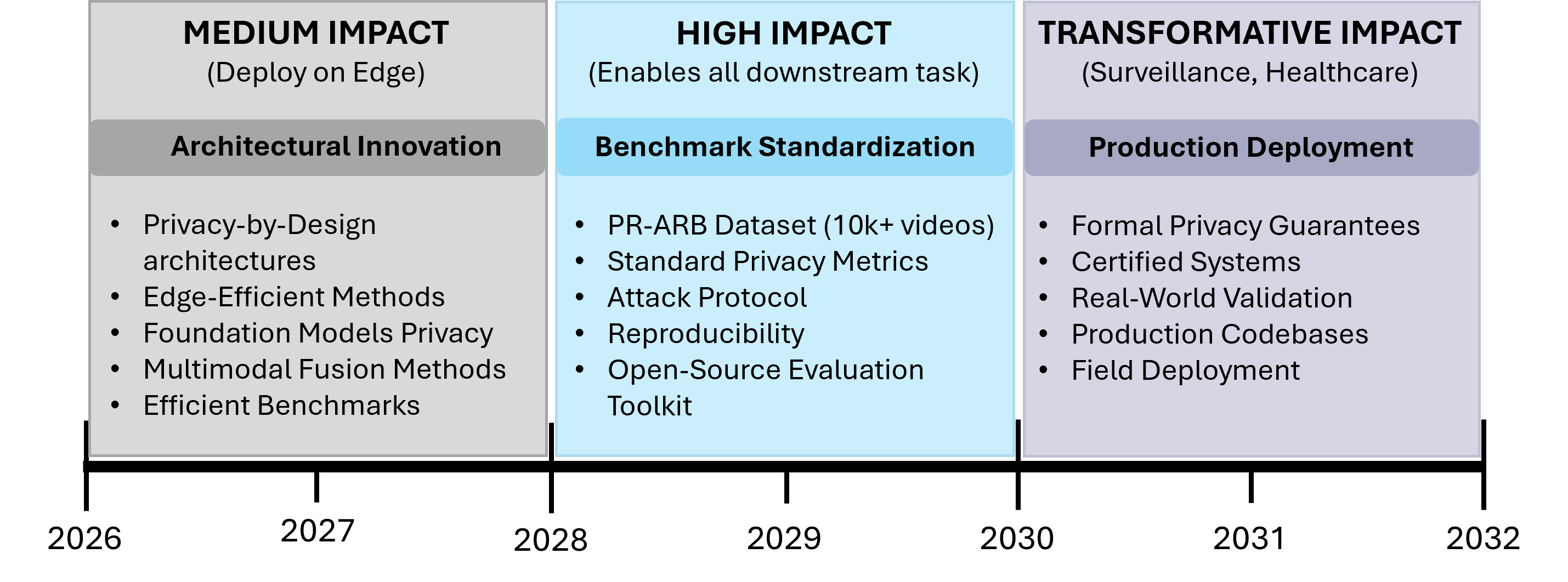}
\caption{Three-tier research roadmap for PPAR. \textit{Medium impact} covers emerging architectural advances including privacy-by-design, edge-efficient methods, and multimodal fusion. \textit{High impact} covers benchmark standardization through the PR-ARB dataset, unified metrics, and an open evaluation toolkit. \textit{Transformative impact} covers certified, formally guaranteed systems deployed in high-stakes domains such as surveillance and healthcare.}
\label{fig:research_roadmap}
\end{figure*}
\section{Future Directions}
The PPAR field stands at an inflection point, and the challenges of Section~\ref{sec:challenges} are tractable through coordinated effort. This section charts five complementary directions, namely foundation models, multimodal privacy, privacy-by-design architectures, edge AI, and benchmark standardization. Together they advance architectural innovation, practical deployment, evaluation maturity, and governance integration, and they form a research roadmap (Fig.~\ref{fig:research_roadmap}) toward deployed, trustworthy PPAR.
\subsection{Foundation-Model-Based Privacy}
Large-scale pre-trained vision models such as Vision Transformers, CLIP, and multimodal foundation models introduce information flows unlike those in CNNs, since they replace local convolutions with self-attention over patch embeddings. Their privacy properties are underexplored and may show different leakage patterns.
\subsubsection{Scaling Privacy via Pre-trained Vision Transformers}
A key question is whether adversarial privacy in ViT latent space differs from that in CNNs. If identity localizes to specific tokens or attention patterns, transformer attention could enable new injection mechanisms. One is selective masking of identity-bearing tokens at inference. Another is hierarchical attention, where lower layers preserve action features and upper layers suppress appearance cues. STPrivacy showed ViT action recognition can be privacy-aware but did not investigate these options systematically~\cite{li_stprivacy_2023}. A concrete direction is privacy-aware fine-tuning for pre-trained ViTs, since downstream adaptation may erode pre-training privacy. Knowledge distillation offers one route. It trains a teacher ViT with strong privacy guarantees, then distills to a lightweight student that keeps privacy at lower deployment cost.
\subsubsection{Privacy-Aware Foundation Model Adaptation}
Fine-tuning foundation models for downstream tasks risks catastrophic forgetting, since adapting a privacy-aware model to new action data may re-enable leakage. LoRA and adapter modules can constrain adaptation and may preserve privacy properties. Prompt-based fine-tuning of vision-language models such as CLIP instead allows semantic privacy specifications, where users state preferences such as ``hide identity but preserve action'' in natural language.
\subsubsection{Cross-Modal Privacy Leakage}
Foundation models encode information across image, text, audio, and video channels. A model may suppress appearance in visual features yet re-encode identity through correlated audio or text encoders. This cross-modal leakage is largely unexplored~\cite{zhao_audio-visual_2024}. Future work should develop cross-modal attack algorithms and multi-modal privacy metrics that account for leakage across modalities.
\subsection{Multimodal Privacy}
Multimodal fusion imposes different privacy requirements across modalities. It also opens a new threat vector, since cross-modal attacks can exploit inter-modal leakage.
\subsubsection{Heterogeneous Privacy Requirements}
Modalities carry different identity risks, including residual re-identification from skeleton data (Section~\ref{sec:taxonomy})~\cite{hirose_anonymization_2022,moon_anonymization_2023,gao_privacy-preserving_2025}. However, the literature treats them independently and applies a mechanism designed for RGB to skeleton unchanged, despite these differing needs. A key direction is modality-specific privacy budgets. RGB receives the strongest mechanism such as adversarial learning or encryption. Skeleton uses lighter anonymization through joint suppression. Audio uses speech anonymization. A fusion module then gates modalities adaptively, suppressing low-quality RGB when privacy risk is high and boosting skeleton when it is available. This adaptive privacy gating has not been studied systematically.
\subsubsection{Cross-Modal Privacy Attacks}
If skeleton is protected but RGB leaks, an adversary can train skeleton-to-appearance reconstruction with GANs or diffusion to invert it. Worse, leaked audio gender and age can strengthen identity inference on the skeleton stream. These cross-modal vectors remain largely unexplored in PPAR. Adversarial and backdoor attacks already expose the fragility~\cite{kumar_finding_2020,jin_frequency_2025,yan_efficient_2021,wang_joint_2025,ko_privmon_2023,zhang_backdoor_2024}, and query-efficient variants now target skeleton recognition directly~\cite{cao_bones_2026}. A concrete direction is to formalize privacy through modality complementarity, fusing modalities such as skeleton and depth so no single one leaks identity fully, as emerging federated and vision-language methods explore~\cite{yang_cross-modal_2024,palit_federated_2026}.
\subsection{Privacy-by-Design Architectures}
Current PPAR methods bolt privacy reactively onto standard architectures. Privacy-by-Design (PbD) instead optimizes the architecture for privacy-utility trade-offs from the start.
\subsubsection{Architectural Inductive Biases for Privacy}
Rather than the post-hoc adversarial layer of prior work~\cite{wu_privacy-preserving_2022,dave_spact_2022,li_stprivacy_2023}, PbD encodes privacy through architectural inductive biases. One is hierarchical attention, where lower transformer layers encode task-relevant actions and upper layers suppress appearance cues by design. A second is feature disentanglement into separate streams with different privacy constraints. The identity stream receives the strongest protection through heavy adversarial training or differential privacy. The action stream receives minimal protection. The appearance stream is discarded entirely.
\subsubsection{Modular Privacy Knobs}
Applications sit at different privacy-utility points, yet traversing the frontier now requires retraining. PbD should expose tunable privacy knobs, such as learnable masks or selective layer freezing, that interpolate privacy without retraining~\cite{ishikawa_learnable_2024}. Once trained at maximal privacy, the model serves any operating point.
\subsubsection{Hardware-Aware Privacy Architectures}
PbD must co-optimize privacy and hardware efficiency on the edge. ViTs may offer inherent privacy advantages but are costly there, while lightweight MobileNets are efficient yet expose fewer privacy injection points. Research should map privacy-efficiency Pareto frontiers across architecture families and edge devices.
\subsection{Edge AI and On-Device Privacy}
Real-world PPAR runs on resource-constrained edge devices, with limited compute and memory, no cloud offloading, and continuous operation.
\subsubsection{Privacy-Efficiency Trade-Offs on Edge}
Luo et al.\ 2024's differentially private video recognition adds substantial and possibly edge-infeasible overhead~\cite{luo_differentially_2024}. Skeleton methods (Gao et al.\ 2025; Moon et al.\ 2023) are lightweight but drop appearance information~\cite{gao_privacy-preserving_2025,moon_anonymization_2023}. Systematic profiling is missing. It is unclear which mechanisms suit which devices, or whether skeleton plus local DP beats adversarial learning on the edge. A key direction is privacy-efficiency benchmarks that profile major PPAR methods on edge hardware such as ARM, mobile GPUs, and edge TPUs, reporting FPS, latency, memory, and energy alongside privacy-utility metrics. Early edge systems already pursue real-time on-device privacy-preserving analytics, including lightweight and spiking-neural-network designs~\cite{huang_semantic_2024,barthelemy_safety_2024,rodriguez-conde_-device_2021,singh_real-time_2023,le_dsc-snn_2025,dilmaghani_lightweight_2026}.
\subsubsection{On-Device Learning and Privacy}
On-device training enables privacy-positive settings that the cloud cannot match, since models learn locally and video never leaves the camera. Federated learning extends this across devices without centralizing data, though federated DP is costly and client dropout weakens guarantees. A growing body applies federated learning to skeleton-based action recognition and studies privacy attacks, personalization, and poisoning defenses~\cite{aouedi_federated_2024,guo_fsar_2023,jourdan_privacy_2021,kim_feddet_2025,bala_privacy_2025,barros_personalized_2025,swami_privacy-preserving_2025,wu_video_2025,doshi_federated_2022,dinh_floweraction_2025,nguyen_fedfslar_2024,tu_benchmarking_2024,kim_fedpure_2026,gad_joint_2023,rao_privacy-preserving_2025}.
\subsubsection{Hardware Security Primitives}
Edge devices increasingly embed hardware security modules such as TEEs, secure enclaves, and TPMs. A PbD opportunity places privacy-critical computations in these secure regions while task inference runs on the main processor. This gives hardware-backed privacy guarantees and reduces software vulnerabilities.
\subsection{Benchmark Standardization and Reproducibility}
The fragmentation of Section~\ref{sec:challenges} motivates standardization, the infrastructure that enables every other direction.
\subsubsection{Unified Privacy Metrics Standard}
About 65\% of papers use ad-hoc metrics, only 10\% use formal DP, and cMAP (40\%) is defined inconsistently, so privacy claims are not comparable. Three options could build consensus. One is to adopt cMAP universally with a formal definition. Another is to adopt formal DP for all claims and calibrate $\varepsilon$ across methods. A third is a hierarchical framework with a primary metric, cMAP or DP, plus secondary metrics.
\subsubsection{Standardized Attack Protocol}
A consensus protocol would standardize four attack tiers. The naive tier is a same-distribution classifier. The unseen-attacker tier uses a different domain or different attributes. The adaptive tier assumes the attacker knows the mechanism. The strong tier covers reconstruction, membership inference, and GAN inversion. Each tier needs a reference implementation, and papers should report robustness against at least three types.
\subsubsection{Open-Source Ecosystem}
Three commitments would speed progress, namely mandatory code release, a standardized evaluation codebase (privacy-eval-kit), and a pre-trained model zoo. Together they cut duplication and support reproducibility.
\subsection{Bringing the Directions Together}
The five directions are interdependent. Foundation-model and multimodal advances need clean evaluation, and privacy-by-design architectures suit edge deployment. Benchmark standardization is the enabling infrastructure that measures progress across all the others. The concluding section consolidates them into the survey's contributions.
\section{Conclusion}
This survey reviewed the recent literature on privacy-preserving action recognition (2018--2026). Five mechanism families have emerged, namely adversarial learning, skeleton-based, differential privacy, cryptographic, and hybrid methods, yet none dominates, and the privacy-utility trade-off stays steep, since skeleton-based methods preserve the most utility, adversarial methods sit in the middle, and differential privacy reaches formal guarantees only at heavy utility cost, with most work clustering near a practical operating point of 60--70\% privacy at 70--85\% utility. However, the larger obstacle is evaluation rather than algorithms, because most papers use ad-hoc metrics and naive attackers, fewer than 10\% test adaptive attacks, and fewer than 15\% test cross-dataset generalization, so reported robustness rarely transfers, while cMAP is defined inconsistently, formal DP is rare, benchmarks are small, only about 40\% of papers release code, and real-time edge deployment stays largely undemonstrated. To address this fragmentation, we contributed a two-dimensional privacy-space taxonomy, a formal problem definition, a comparative analysis, and a unified evaluation protocol. These findings expose five coupled challenges, namely the lack of a shared privacy definition, weak cross-dataset generalization, thin adaptive-attacker evaluation, neglected real-time constraints, and unmet legal and governance duties, and each one lets the others pass unchecked, so isolated fixes will not move the field. The more promising path runs through foundation-model and multimodal privacy, privacy-by-design architectures, on-device and edge deployment, and above all benchmark standardization, which is the infrastructure that makes every other advance measurable. Taken together, the evidence suggests that PPAR is technically feasible but not yet trustworthy or deployable at scale, and that closing the gap depends less on new algorithms than on shared foundations such as a consensus privacy definition, large-scale privacy-aware benchmarks with standardized attack protocols, mandatory efficiency reporting, and open, reproducible codebases, so that the field can consolidate its fragmented methods into a few standardized, certified approaches within about five years, and finally show that privacy and utility are co-optimizable through principled design rather than traded off after the fact.

\bibliographystyle{IEEEtran}
\bibliography{Ref}

@article{wu_privacy-preserving_2022,
	title = {Privacy-Preserving Deep Action Recognition: An Adversarial Learning Framework and A New Dataset},
	volume = {44},
	issn = {1939-3539},
	doi = {10.1109/TPAMI.2020.3026709},
	shorttitle = {Privacy-Preserving Deep Action Recognition},
	pages = {2126--2139},
	number = {4},
	journaltitle = {{IEEE} Transactions on Pattern Analysis and Machine Intelligence},
	author = {Wu, Zhenyu and Wang, Haotao and Wang, Zhaowen and Jin, Hailin and Wang, Zhangyang},
	date = {2022-04},
}

@inproceedings{zou_privacy-preserving_2022,
	title = {Privacy-Preserving Action Recognition},
	issn = {2379-190X},
	doi = {10.1109/ICASSP43922.2022.9747456},
	eventtitle = {{ICASSP} 2022 - 2022 {IEEE} International Conference on Acoustics, Speech and Signal Processing ({ICASSP})},
	pages = {2175--2179},
	booktitle = {{ICASSP} 2022 - 2022 {IEEE} International Conference on Acoustics, Speech and Signal Processing ({ICASSP})},
	author = {Zou, Chengming and Yuan, Ducheng and Lan, Long and Chi, Haoang},
	date = {2022-05},
	note = {{ISSN}: 2379-190X},
}

@inproceedings{li_patch-based_2024,
	title = {Patch-Based Privacy Attention for Weakly-Supervised Privacy-Preserving Action Recognition},
	issn = {2770-8330},
	doi = {10.1109/FG59268.2024.10582006},
	eventtitle = {2024 {IEEE} 18th International Conference on Automatic Face and Gesture Recognition ({FG})},
	pages = {1--9},
	booktitle = {2024 {IEEE} 18th International Conference on Automatic Face and Gesture Recognition ({FG})},
	author = {Li, Xiao and Qiu, Yu-Kun and Peng, Yi-Xing and Zheng, Wei-Shi},
	date = {2024-05},
	note = {{ISSN}: 2770-8330},
}

@inproceedings{zheng_fast_2024,
	title = {A Fast and Tunable Privacy-Preserving Action Recognition Framework over Compressed Video},
	issn = {1945-788X},
	doi = {10.1109/ICME57554.2024.10687883},
	eventtitle = {2024 {IEEE} International Conference on Multimedia and Expo ({ICME})},
	pages = {1--6},
	booktitle = {2024 {IEEE} International Conference on Multimedia and Expo ({ICME})},
	author = {Zheng, Qingfeng and Zheng, Peijia and Luo, Weiqi and Lu, Wei},
	date = {2024-07},
	note = {{ISSN}: 1945-788X},
}

@inproceedings{dave_spact_2022,
	title = {{SPAct}: Self-supervised Privacy Preservation for Action Recognition},
	issn = {2575-7075},
	doi = {10.1109/CVPR52688.2022.01953},
	shorttitle = {{SPAct}},
	eventtitle = {2022 {IEEE}/{CVF} Conference on Computer Vision and Pattern Recognition ({CVPR})},
	pages = {20132--20141},
	booktitle = {2022 {IEEE}/{CVF} Conference on Computer Vision and Pattern Recognition ({CVPR})},
	author = {Dave, Ishan Rajendrakumar and Chen, Chen and Shah, Mubarak},
	date = {2022-06},
	note = {{ISSN}: 2575-7075},
}

@inproceedings{ishikawa_learnable_2024,
	title = {Learnable Cube-based Video Encryption for Privacy-Preserving Action Recognition},
	issn = {2642-9381},
	doi = {10.1109/WACV57701.2024.00685},
	eventtitle = {2024 {IEEE}/{CVF} Winter Conference on Applications of Computer Vision ({WACV})},
	pages = {6988--6998},
	booktitle = {2024 {IEEE}/{CVF} Winter Conference on Applications of Computer Vision ({WACV})},
	author = {Ishikawa, Yuchi and Kondo, Masayoshi and Kataoka, Hirokatsu},
	date = {2024-01},
	note = {{ISSN}: 2642-9381},
}

@inproceedings{ilic_selective_2024,
	title = {Selective, Interpretable and Motion Consistent Privacy Attribute Obfuscation for Action Recognition},
	issn = {2575-7075},
	doi = {10.1109/CVPR52733.2024.01772},
	eventtitle = {2024 {IEEE}/{CVF} Conference on Computer Vision and Pattern Recognition ({CVPR})},
	pages = {18730--18739},
	booktitle = {2024 {IEEE}/{CVF} Conference on Computer Vision and Pattern Recognition ({CVPR})},
	author = {Ilic, Filip and Zhao, He and Pock, Thomas and Wildes, Richard P.},
	date = {2024-06},
	note = {{ISSN}: 2575-7075},
}

@article{gao_privacy-preserving_2025,
	title = {Privacy-Preserving 3-D Skeleton-Based Video Action Recognition via Graph Convolution Network},
	volume = {71},
	issn = {1558-4127},
	doi = {10.1109/TCE.2024.3476273},
	pages = {6627--6641},
	number = {2},
	journaltitle = {{IEEE} Transactions on Consumer Electronics},
	author = {Gao, Xuesong and Li, Keqiu and Liu, Xiulong and Nie, Jie and Chen, Weiqiang and Tian, Yonghong},
	date = {2025-05},
}

@inproceedings{li_stprivacy_2023,
	title = {{STPrivacy}: Spatio-Temporal Privacy-Preserving Action Recognition},
	issn = {2380-7504},
	doi = {10.1109/ICCV51070.2023.00471},
	shorttitle = {{STPrivacy}},
	eventtitle = {2023 {IEEE}/{CVF} International Conference on Computer Vision ({ICCV})},
	pages = {5083--5092},
	booktitle = {2023 {IEEE}/{CVF} International Conference on Computer Vision ({ICCV})},
	author = {Li, Ming and Xu, Xiangyu and Fan, Hehe and Zhou, Pan and Liu, Jun and Liu, Jia-Wei and Li, Jiahe and Keppo, Jussi and Shou, Mike Zheng and Yan, Shuicheng},
	date = {2023-10},
	note = {{ISSN}: 2380-7504},
}

@inproceedings{ren_privacy_2022,
	title = {A Privacy Preserving Video Surveillance System for Trauma Rooms},
	doi = {10.1109/SmartWorld-UIC-ATC-ScalCom-DigitalTwin-PriComp-Metaverse56740.2022.00244},
	eventtitle = {2022 {IEEE} Smartworld, Ubiquitous Intelligence \& Computing, Scalable Computing \& Communications, Digital Twin, Privacy Computing, Metaverse, Autonomous \& Trusted Vehicles ({SmartWorld}/{UIC}/{ScalCom}/{DigitalTwin}/{PriComp}/Meta)},
	pages = {1716--1721},
	booktitle = {2022 {IEEE} Smartworld, Ubiquitous Intelligence \& Computing, Scalable Computing \& Communications, Digital Twin, Privacy Computing, Metaverse, Autonomous \& Trusted Vehicles ({SmartWorld}/{UIC}/{ScalCom}/{DigitalTwin}/{PriComp}/Meta)},
	author = {Ren, Zhengyong and Ghazinour, Kambiz and Guan, Qiang and Bayramzadeh, Sara and Yang, Yuxin},
	date = {2022-12},
}

@inproceedings{fioresi_ted-spad_2023,
	title = {{TeD}-{SPAD}: Temporal Distinctiveness for Self-supervised Privacy-preservation for video Anomaly Detection},
	issn = {2380-7504},
	doi = {10.1109/ICCV51070.2023.01251},
	shorttitle = {{TeD}-{SPAD}},
	eventtitle = {2023 {IEEE}/{CVF} International Conference on Computer Vision ({ICCV})},
	pages = {13552--13563},
	booktitle = {2023 {IEEE}/{CVF} International Conference on Computer Vision ({ICCV})},
	author = {Fioresi, Joseph and Dave, Ishan Rajendrakumar and Shah, Mubarak},
	date = {2023-10},
	note = {{ISSN}: 2380-7504},
}

@inproceedings{luo_differentially_2024,
	title = {Differentially Private Video Activity Recognition},
	issn = {2642-9381},
	doi = {10.1109/WACV57701.2024.00652},
	eventtitle = {2024 {IEEE}/{CVF} Winter Conference on Applications of Computer Vision ({WACV})},
	pages = {6643--6653},
	booktitle = {2024 {IEEE}/{CVF} Winter Conference on Applications of Computer Vision ({WACV})},
	author = {Luo, Zelun and Zou, Yuliang and Yang, Yijin and Durante, Zane and Huang, De-An and Yu, Zhiding and Xiao, Chaowei and Fei-Fei, Li and Anandkumar, Animashree},
	date = {2024-01},
	note = {{ISSN}: 2642-9381},
}

@inproceedings{houshidari_three_2025,
	title = {Three Decades of Smart Noninvasive Older Adults Monitoring: Trends in Objectives, Privacy Preserving, Modality and {AI} Assistance},
	issn = {2572-6927},
	doi = {10.1109/SGC69320.2025.11372285},
	shorttitle = {Three Decades of Smart Noninvasive Older Adults Monitoring},
	eventtitle = {2025 15th Smart Grid Conference ({SGC})},
	pages = {1--7},
	booktitle = {2025 15th Smart Grid Conference ({SGC})},
	author = {Houshidari, Kimia and Houshidari, Farhad and Fereidunian, Alireza},
	date = {2025-12},
	note = {{ISSN}: 2572-6927},
}

@article{hirose_anonymization_2022,
	title = {Anonymization of Human Gait in Video Based on Silhouette Deformation and Texture Transfer},
	volume = {17},
	issn = {1556-6021},
	doi = {10.1109/TIFS.2022.3206422},
	pages = {3375--3390},
	journaltitle = {{IEEE} Transactions on Information Forensics and Security},
	author = {Hirose, Yuki and Nakamura, Kazuaki and Nitta, Naoko and Babaguchi, Noboru},
	date = {2022},
}

@inproceedings{tomei_estimating_2021,
	title = {Estimating (and fixing) the Effect of Face Obfuscation in Video Recognition},
	issn = {2160-7516},
	doi = {10.1109/CVPRW53098.2021.00364},
	eventtitle = {2021 {IEEE}/{CVF} Conference on Computer Vision and Pattern Recognition Workshops ({CVPRW})},
	pages = {3257--3263},
	booktitle = {2021 {IEEE}/{CVF} Conference on Computer Vision and Pattern Recognition Workshops ({CVPRW})},
	author = {Tomei, Matteo and Baraldi, Lorenzo and Bronzin, Simone and Cucchiara, Rita},
	date = {2021-06},
	note = {{ISSN}: 2160-7516},
}

@article{jain_privacy-preserving_2024,
	title = {Privacy-Preserving Human Activity Recognition System for Assisted Living Environments},
	volume = {5},
	issn = {2691-4581},
	doi = {10.1109/TAI.2023.3323272},
	pages = {2342--2357},
	number = {5},
	journaltitle = {{IEEE} Transactions on Artificial Intelligence},
	author = {Jain, Ankit and Akerkar, Rajendra and Srivastava, Abhishek},
	date = {2024-05},
}

@inproceedings{zhicai_east_2025,
	title = {{EAST}: Edge-Aware Self-Supervised Transformer for Privacy-Compliant Video Analysis},
	doi = {10.1109/ICCR67387.2025.11291780},
	shorttitle = {{EAST}},
	eventtitle = {2025 3rd International Conference on Cyber Resilience ({ICCR})},
	pages = {1--7},
	booktitle = {2025 3rd International Conference on Cyber Resilience ({ICCR})},
	author = {Zhicai, Li and Abdullah, Siti Norul Huda Sheikh and Jiale, Qi and Ghani, Ahmad Tarmizi Abdul},
	date = {2025-07},
}

@inproceedings{maximov_ciagan_2020,
	title = {{CIAGAN}: Conditional Identity Anonymization Generative Adversarial Networks},
	issn = {2575-7075},
	doi = {10.1109/CVPR42600.2020.00549},
	shorttitle = {{CIAGAN}},
	pages = {5446--5455},
	booktitle = {2020 {IEEE}/{CVF} Conference on Computer Vision and Pattern Recognition ({CVPR})},
	author = {Maximov, Maxim and Elezi, Ismail and Leal-Taix{\'e}, Laura},
	date = {2020-06},
	note = {{ISSN}: 2575-7075},
}

@article{zhao_visual_2025,
	title = {Visual Content Privacy Protection: A Survey},
	volume = {57},
	issn = {0360-0300},
	doi = {10.1145/3708501},
	shorttitle = {Visual Content Privacy Protection},
	pages = {122:1--122:36},
	number = {5},
	journaltitle = {{ACM} Comput. Surv.},
	author = {Zhao, Ruoyu and Zhang, Yushu and Wang, Tao and Wen, Wenying and Xiang, Yong and Cao, Xiaochun},
	date = {2025-01-24},
}

@article{tong_image_2021,
	title = {An Image Privacy Protection Algorithm Based on Adversarial Perturbation Generative Networks},
	volume = {17},
	issn = {1551-6857},
	doi = {10.1145/3381088},
	pages = {43:1--43:14},
	number = {2},
	journaltitle = {{ACM} Trans. Multimedia Comput. Commun. Appl.},
	author = {Tong, Chao and Zhang, Mengze and Lang, Chao and Zheng, Zhigao},
	date = {2021-04-21},
}

@inproceedings{zhou_privacy-sensitive_2020,
	location = {New York, {NY}, {USA}},
	title = {Privacy-sensitive Objects Pixelation for Live Video Streaming},
	isbn = {978-1-4503-7988-5},
	doi = {10.1145/3394171.3413972},
	series = {{MM} '20},
	pages = {3025--3033},
	booktitle = {Proceedings of the 28th {ACM} International Conference on Multimedia},
	publisher = {Association for Computing Machinery},
	author = {Zhou, Jizhe and Pun, Chi-Man and Tong, Yu},
	date = {2020-10-12},
}

@article{cai_generative_2021,
	title = {Generative Adversarial Networks: A Survey Toward Private and Secure Applications},
	volume = {54},
	issn = {0360-0300},
	doi = {10.1145/3459992},
	shorttitle = {Generative Adversarial Networks},
	pages = {132:1--132:38},
	number = {6},
	journaltitle = {{ACM} Comput. Surv.},
	author = {Cai, Zhipeng and Xiong, Zuobin and Xu, Honghui and Wang, Peng and Li, Wei and Pan, Yi},
	date = {2021-07-13},
}

@article{hellmann_ganonymization_2024,
	title = {{GANonymization}: A {GAN}-Based Face Anonymization Framework for Preserving Emotional Expressions},
	volume = {21},
	issn = {1551-6857},
	doi = {10.1145/3641107},
	shorttitle = {{GANonymization}},
	pages = {6:1--6:27},
	number = {1},
	journaltitle = {{ACM} Trans. Multimedia Comput. Commun. Appl.},
	author = {Hellmann, Fabio and Mertes, Silvan and Benouis, Mohamed and Hustinx, Alexander and Hsieh, Tzung-Chien and Conati, Cristina and Krawitz, Peter and Andr{\'e}, Elisabeth},
	date = {2024-12-14},
}

@article{dou_person_2026,
	location = {Beijing},
	title = {Person identity shift for privacy-preserving person re-identification},
	volume = {69},
	issn = {1674-733X, 1869-1919},
	doi = {10.1007/s11432-023-4583-x},
	pages = {112103},
	number = {1},
	journaltitle = {Science China-Information Sciences},
	shortjournal = {Sci. China-Inf. Sci.},
	publisher = {Science Press},
	author = {Dou, Shuguang and Jiang, Xinyang and Zhao, Qingsong and Wang, Yansen and Li, Dongsheng and Zhao, Cairong},
	date = {2026-01-04},
	note = {Num Pages: 14
Web of Science {ID}: {WOS}:001660148200005},
}

@article{feng_priva_2025,
	location = {Bellingham},
	title = {Priva: privacy reversible and intelligible video analytics system through diffusion restoration models},
	volume = {34},
	issn = {1017-9909, 1560-229X},
	doi = {10.1117/1.JEI.34.2.023022},
	shorttitle = {Priva},
	number = {2},
	journaltitle = {Journal of Electronic Imaging},
	shortjournal = {J. Electron. Imaging},
	publisher = {Spie-Soc Photo-Optical Instrumentation Engineers},
	author = {Feng, Hanqi and Hu, Chuang and Cheng, Dazhao},
	date = {2025-03-01},
	note = {Num Pages: 15
Web of Science {ID}: {WOS}:001488782700014},
}

@article{cucchiara_video_2024,
	location = {Los Alamitos},
	title = {Video Surveillance and Privacy: A Solvable Paradox?},
	volume = {57},
	issn = {0018-9162, 1558-0814},
	doi = {10.1109/MC.2023.3316696},
	shorttitle = {Video Surveillance and Privacy},
	pages = {91--100},
	number = {3},
	journaltitle = {Computer},
	shortjournal = {Computer},
	publisher = {{IEEE} Computer Soc},
	author = {Cucchiara, Rita and Baraldi, Lorenzo and Cornia, Marcella and Sarto, Sara},
	date = {2024-03},
	note = {Num Pages: 10
Web of Science {ID}: {WOS}:001180702200010},
}

@inproceedings{moon_anonymization_2023,
	location = {Palo Alto},
	title = {Anonymization for Skeleton Action Recognition},
	issn = {2159-5399, 2374-3468},
	pages = {15028--15036},
	booktitle = {Thirty-Seventh Aaai Conference on Artificial Intelligence, Vol 37 No 12},
	publisher = {Assoc Advancement Artificial Intelligence},
	author = {Moon, Saemi and Kim, Myeonghyeon and Qin, Zhenyue and Liu, Yang and Kim, Dongwoo},
	editor = {Williams, B. and Chen, Y. and Neville, J.},
	date = {2023},
	note = {Num Pages: 9
Series Title: {AAAI} Conference on Artificial Intelligence
Web of Science {ID}: {WOS}:001243755000107},
}

@inproceedings{peng_joint_2023,
	location = {La Jolla},
	title = {Joint Attribute and Model Generalization Learning for Privacy-Preserving Action Recognition},
	issn = {1049-5258},
	eventtitle = {37th Conference on Neural Information Processing Systems ({NeurIPS})},
	booktitle = {Advances in Neural Information Processing Systems 36 (neurips 2023)},
	publisher = {Neural Information Processing Systems (nips)},
	author = {Peng, Duo and Xu, Li and Ke, Qiuhong and Hu, Ping and Liu, Jun},
	editor = {Oh, A. and Neumann, T. and Globerson, A. and Saenko, K. and Hardt, M. and Levine, S.},
	date = {2023},
	note = {Num Pages: 13
Series Title: Advances in Neural Information Processing Systems
Web of Science {ID}: {WOS}:001229751901020},
}

@inproceedings{kumawat_privacy-preserving_2022,
	location = {Cham},
	title = {Privacy-Preserving Action Recognition via Motion Difference Quantization},
	volume = {13673},
	isbn = {978-3-031-19777-2 978-3-031-19778-9},
	issn = {0302-9743, 1611-3349},
	doi = {10.1007/978-3-031-19778-9_30},
	pages = {518--534},
	booktitle = {Computer Vision, Eccv 2022, Pt Xiii},
	publisher = {Springer International Publishing Ag},
	author = {Kumawat, Sudhakar and Nagahara, Hajime},
	editor = {Avidan, S. and Brostow, G. and Cisse, M. and Farinella, G. M. and Hassner, T.},
	date = {2022},
	note = {Num Pages: 17
Series Title: Lecture Notes in Computer Science
Web of Science {ID}: {WOS}:000897100100030},
}

@article{liu_video_2021,
	location = {Amsterdam},
	title = {Video action recognition with visual privacy protection based on compressed sensing},
	volume = {113},
	issn = {1383-7621, 1873-6165},
	doi = {10.1016/j.sysarc.2020.101882},
	pages = {101882},
	journaltitle = {Journal of Systems Architecture},
	shortjournal = {J. Syst. Architect.},
	publisher = {Elsevier},
	author = {Liu, Jixin and Zhang, Ruxue and Han, Guang and Sun, Ning and Kwong, Sam},
	date = {2021-02},
	note = {Num Pages: 14
Web of Science {ID}: {WOS}:000723108000005},
}

@inproceedings{wang_privacy-preserving_2019,
	location = {New York},
	title = {Privacy-Preserving Action Recognition using Coded Aperture Videos},
	isbn = {978-1-7281-2506-0},
	issn = {2160-7508},
	doi = {10.1109/CVPRW.2019.00007},
	eventtitle = {32nd {IEEE}/{CVF} Conference on Computer Vision and Pattern Recognition ({CVPR})},
	pages = {1--10},
	booktitle = {2019 Ieee/Cvf Conference on Computer Vision and Pattern Recognition Workshops (cvprw 2019)},
	publisher = {{IEEE}},
	author = {Wang, Zihao W. and Vineet, Vibhav and Pittaluga, Francesco and Sinha, Sudipta N. and Cossairt, Oliver and Kang, Sing Bing},
	date = {2019},
	note = {Num Pages: 10
Series Title: {IEEE} Computer Society Conference on Computer Vision and Pattern Recognition Workshops
Web of Science {ID}: {WOS}:000569983600001},
}

@inproceedings{wu_towards_2018,
	location = {Cham},
	title = {Towards Privacy-Preserving Visual Recognition via Adversarial Training: A Pilot Study},
	volume = {11220},
	isbn = {978-3-030-01270-0 978-3-030-01269-4},
	issn = {0302-9743, 1611-3349},
	doi = {10.1007/978-3-030-01270-0_37},
	shorttitle = {Towards Privacy-Preserving Visual Recognition via Adversarial Training},
	eventtitle = {15th European Conference on Computer Vision ({ECCV})},
	pages = {627--645},
	booktitle = {Computer Vision - Eccv 2018, Pt Xvi},
	publisher = {Springer International Publishing Ag},
	author = {Wu, Zhenyu and Wang, Zhangyang and Wang, Zhaowen and Jin, Hailin},
	editor = {Ferrari, V. and Hebert, M. and Sminchisescu, C. and Weiss, Y.},
	date = {2018},
	note = {Num Pages: 19
Series Title: Lecture Notes in Computer Science
Web of Science {ID}: {WOS}:000603403700037},
}

@article{al-obaidi_modeling_2020,
	title = {Modeling Temporal Visual Salience for Human Action Recognition Enabled Visual Anonymity Preservation},
	volume = {8},
	issn = {2169-3536},
	doi = {10.1109/ACCESS.2020.3039740},
	pages = {213806--213824},
	journaltitle = {{IEEE} Access},
	author = {Al-Obaidi, Salah and Al-Khafaji, Hiba and Abhayaratne, Charith},
	date = {2020},
	}

@inproceedings{fujimoto_differential_2023,
	title = {Differential Privacy with Weighted {$\epsilon$} for Privacy-Preservation in Human Activity Recognition},
	issn = {2766-8576},
	doi = {10.1109/PerComWorkshops56833.2023.10150239},
	eventtitle = {2023 {IEEE} International Conference on Pervasive Computing and Communications Workshops and other Affiliated Events ({PerCom} Workshops)},
	pages = {634--639},
	booktitle = {2023 {IEEE} International Conference on Pervasive Computing and Communications Workshops and other Affiliated Events ({PerCom} Workshops)},
	author = {Fujimoto, Ryusei and Nakamura, Yugo and Arakawa, Yutaka},
	date = {2023-03},
	note = {{ISSN}: 2766-8576},
}

@inproceedings{liu_indoor_2020,
	title = {Indoor Privacy-preserving Action Recognition via Partially Coupled Convolutional Neural Network},
	doi = {10.1109/ICAICE51518.2020.00062},
	eventtitle = {2020 International Conference on Artificial Intelligence and Computer Engineering ({ICAICE})},
	pages = {292--295},
	booktitle = {2020 International Conference on Artificial Intelligence and Computer Engineering ({ICAICE})},
	author = {Liu, Jixin and Zhang, Leilei},
	date = {2020-10},
}

@inproceedings{roy_temporal_2023,
	title = {Temporal Differential Privacy for Human Activity Recognition},
	doi = {10.1109/DSAA60987.2023.10302475},
	eventtitle = {2023 {IEEE} 10th International Conference on Data Science and Advanced Analytics ({DSAA})},
	pages = {1--10},
	booktitle = {2023 {IEEE} 10th International Conference on Data Science and Advanced Analytics ({DSAA})},
	author = {Roy, Debaditya and Girdzijauskas, {\v S}ar{\=u}nas},
	date = {2023-10},
}

@article{stirapongsasuti_preserving_2024,
	title = {Preserving Data Utility in Differentially Private Smart Home Data},
	volume = {12},
	issn = {2169-3536},
	doi = {10.1109/ACCESS.2024.3390039},
	pages = {56571--56581},
	journaltitle = {{IEEE} Access},
	author = {Stirapongsasuti, Sopicha and Tiausas, Francis Jerome and Nakamura, Yugo and Yasumoto, Keiichi},
	date = {2024},
}

@inproceedings{htoo_privacy_2023,
	title = {Privacy Preserving Human Fall Recognition Using Human Skeleton Data},
	doi = {10.1109/ICCA51723.2023.10181509},
	eventtitle = {2023 {IEEE} Conference on Computer Applications ({ICCA})},
	pages = {276--281},
	booktitle = {2023 {IEEE} Conference on Computer Applications ({ICCA})},
	author = {Htoo, Chit Kyin and Sein, Myint Myint},
	date = {2023-02},
}

@inproceedings{jourdan_privacy_2021,
	title = {Privacy Assessment of Federated Learning Using Private Personalized Layers},
	issn = {1551-2541},
	doi = {10.1109/MLSP52302.2021.9596237},
	eventtitle = {2021 {IEEE} 31st International Workshop on Machine Learning for Signal Processing ({MLSP})},
	pages = {1--6},
	booktitle = {2021 {IEEE} 31st International Workshop on Machine Learning for Signal Processing ({MLSP})},
	author = {Jourdan, Th{\'e}o and Boutet, Antoine and Frindel, Carole},
	date = {2021-10},
	note = {{ISSN}: 1551-2541},
}

@inproceedings{carr_user_2024,
	title = {User Privacy in Skeleton-based Motion Data},
	issn = {2573-2978},
	doi = {10.1109/BigData62323.2024.10825650},
	eventtitle = {2024 {IEEE} International Conference on Big Data ({BigData})},
	pages = {8219--8221},
	booktitle = {2024 {IEEE} International Conference on Big Data ({BigData})},
	author = {Carr, Thomas and Xu, Depeng},
	date = {2024-12},
	note = {{ISSN}: 2573-2978},
}

@article{liang_multi-modal_2020,
	title = {Multi-Modal Human Action Recognition With Sub-Action Exploiting and Class-Privacy Preserved Collaborative Representation Learning},
	volume = {8},
	issn = {2169-3536},
	doi = {10.1109/ACCESS.2020.2976496},
	pages = {39920--39933},
	journaltitle = {{IEEE} Access},
	author = {Liang, Chengwu and Liu, Deyin and Qi, Lin and Guan, Ling},
	date = {2020},
}

@inproceedings{li_multi-objective_2025,
	title = {Multi-Objective Feature Selection for User Privacy Protection in Human Daily Activity Recognition},
	issn = {2325-2944},
	doi = {10.1109/DCOSS-IoT65416.2025.00017},
	eventtitle = {2025 21st International Conference on Distributed Computing in Smart Systems and the Internet of Things ({DCOSS}-{IoT})},
	pages = {61--68},
	booktitle = {2025 21st International Conference on Distributed Computing in Smart Systems and the Internet of Things ({DCOSS}-{IoT})},
	author = {Li, Qiyang and Vincent, Johanne and Espes, David and Gogniat, Guy},
	date = {2025-06},
	note = {{ISSN}: 2325-2944},
}

@article{swami_privacy-preserving_2025,
	title = {Privacy-Preserving On-Screen Activity Recognition via One-Shot Federated Learning},
	volume = {13},
	issn = {2169-3536},
	doi = {10.1109/ACCESS.2025.3638797},
	pages = {201625--201644},
	journaltitle = {{IEEE} Access},
	author = {Swami, Parayush and Priya, Annu and Palanisamy, Balamurugan and Roy, Debangshu and Hassija, Vikas and Chalapathi, G. S. S.},
	date = {2025},
}

@inproceedings{le_dsc-snn_2025,
	title = {{DSC}-{SNN}: A Depthwise Separable Convolutional Spiking Neural Network for Efficient, Privacy-Preserving Action Recognition from Event-Based Data},
	issn = {2770-6850},
	doi = {10.1109/MAPR67746.2025.11133952},
	shorttitle = {{DSC}-{SNN}},
	eventtitle = {2025 International Conference on Multimedia Analysis and Pattern Recognition ({MAPR})},
	pages = {1--6},
	booktitle = {2025 International Conference on Multimedia Analysis and Pattern Recognition ({MAPR})},
	author = {Le, Nguyen Tan Khang and Nguyen, Ha Dung and Nguyen, Thanh Binh},
	date = {2025-08},
	note = {{ISSN}: 2770-6850},
}

@inproceedings{carr_anonvis_2025,
	title = {{AnonVis}: A Visualization Tool for Human Motion Anonymization},
	issn = {2771-1110},
	doi = {10.1109/ISMAR-Adjunct68609.2025.00278},
	shorttitle = {{AnonVis}},
	eventtitle = {2025 {IEEE} International Symposium on Mixed and Augmented Reality Adjunct ({ISMAR}-Adjunct)},
	pages = {988--989},
	booktitle = {2025 {IEEE} International Symposium on Mixed and Augmented Reality Adjunct ({ISMAR}-Adjunct)},
	author = {Carr, Thomas and Flanagan, Ruby and Bastakoti, Albert and Xu, Depeng and Lu, Aidong},
	date = {2025-10},
	note = {{ISSN}: 2771-1110},
}

@inproceedings{low_adverfacial_2022,
	title = {{AdverFacial}: Privacy-Preserving Universal Adversarial Perturbation Against Facial Micro-Expression Leakages},
	issn = {2379-190X},
	doi = {10.1109/ICASSP43922.2022.9746848},
	shorttitle = {{AdverFacial}},
	eventtitle = {{ICASSP} 2022 - 2022 {IEEE} International Conference on Acoustics, Speech and Signal Processing ({ICASSP})},
	pages = {2754--2758},
	booktitle = {{ICASSP} 2022 - 2022 {IEEE} International Conference on Acoustics, Speech and Signal Processing ({ICASSP})},
	author = {Low, Yin-Yin and Tanvy, Angeline and Phan, Rapha{\"e}l C.-W. and Chang, Xiaojun},
	date = {2022-05},
	note = {{ISSN}: 2379-190X},
}

@inproceedings{gutev_depth-based_2025,
	title = {Depth-based Human Action Recognition for Private Settings using Motion Saliency Detection},
	issn = {2471-8963},
	doi = {10.1109/EUVIP66349.2025.11238648},
	eventtitle = {2025 13th European Workshop on Visual Information Processing ({EUVIP})},
	pages = {1--7},
	booktitle = {2025 13th European Workshop on Visual Information Processing ({EUVIP})},
	author = {Gutev, Alexander and Debono, Carl James},
	date = {2025-10},
	note = {{ISSN}: 2471-8963},
}

@inproceedings{tanwar_preserving_2023,
	title = {Preserving Privacy in Image Database through Bit-planes Obfuscation},
	issn = {2473-3490},
	doi = {10.1109/ICDEW58674.2023.00027},
	eventtitle = {2023 {IEEE} 39th International Conference on Data Engineering Workshops ({ICDEW})},
	pages = {132--137},
	booktitle = {2023 {IEEE} 39th International Conference on Data Engineering Workshops ({ICDEW})},
	author = {Tanwar, Vishesh K. and Gupta, Ashish and Madria, Sanjay and Das, Sajal K.},
	date = {2023-04},
	note = {{ISSN}: 2473-3490},
}

@inproceedings{bala_privacy_2025,
	title = {Privacy Preserving Federated Random Forest Approach for Human Activity Recognition},
	issn = {2996-4393},
	doi = {10.1109/ICHORA65333.2025.11016851},
	eventtitle = {2025 7th International Congress on Human-Computer Interaction, Optimization and Robotic Applications ({ICHORA})},
	pages = {1--6},
	booktitle = {2025 7th International Congress on Human-Computer Interaction, Optimization and Robotic Applications ({ICHORA})},
	author = {Bala, Liza and Sultana, Nayeema and Islam, Md Rashedul and Ahmed, Nadeem},
	date = {2025-05},
	note = {{ISSN}: 2996-4393},
}

@article{dilmaghani_lightweight_2026,
	title = {A Lightweight 3D-{CNN} for Event-Based Human Action Recognition With Privacy-Preserving Potential},
	volume = {14},
	issn = {2169-3536},
	doi = {10.1109/ACCESS.2026.3660117},
	pages = {18193--18205},
	journaltitle = {{IEEE} Access},
	author = {Dilmaghani, Mehdi Sefidgar and Fowley, Frank and Corcoran, Peter},
	date = {2026},
}

@inproceedings{dai_towards_2015,
	title = {Towards privacy-preserving recognition of human activities},
	doi = {10.1109/ICIP.2015.7351605},
	eventtitle = {2015 {IEEE} International Conference on Image Processing ({ICIP})},
	pages = {4238--4242},
	booktitle = {2015 {IEEE} International Conference on Image Processing ({ICIP})},
	author = {Dai, Ji and Saghafi, Behrouz and Wu, Jonathan and Konrad, Janusz and Ishwar, Prakash},
	date = {2015-09},
}

@inproceedings{sun_human_2020,
	title = {Human Action Image Generation with Differential Privacy},
	issn = {1945-788X},
	doi = {10.1109/ICME46284.2020.9102767},
	eventtitle = {2020 {IEEE} International Conference on Multimedia and Expo ({ICME})},
	pages = {1--6},
	booktitle = {2020 {IEEE} International Conference on Multimedia and Expo ({ICME})},
	author = {Sun, Mingxuan and Wang, Qing and Liu, Zicheng},
	date = {2020-07},
	note = {{ISSN}: 1945-788X},
}

@article{aouedi_federated_2024,
	title = {Federated Learning for Human Activity Recognition: Overview, Advances, and Challenges},
	volume = {5},
	issn = {2644-125X},
	doi = {10.1109/OJCOMS.2024.3484228},
	shorttitle = {Federated Learning for Human Activity Recognition},
	pages = {7341--7367},
	journaltitle = {{IEEE} Open Journal of the Communications Society},
	author = {Aouedi, Ons and Sacco, Alessio and Khan, Latif U. and Nguyen, Dinh C. and Guizani, Mohsen},
	date = {2024},
}

@article{arguello_learning_2024,
	title = {Learning to Describe Scenes via Privacy-Aware Designed Optical Lens},
	volume = {10},
	issn = {2333-9403},
	doi = {10.1109/TCI.2024.3426975},
	pages = {1069--1079},
	journaltitle = {{IEEE} Transactions on Computational Imaging},
	author = {Arguello, Paula and Lopez, Jhon and Sanchez, Karen and Hinojosa, Carlos and Rojas-Morales, Fernando and Arguello, Henry},
	date = {2024},
}

@inproceedings{salama_multi-level_2017,
	title = {Multi-Level Privacy-Preserving Access Control as a Service for Personal Healthcare Monitoring},
	doi = {10.1109/ICWS.2017.111},
	eventtitle = {2017 {IEEE} International Conference on Web Services ({ICWS})},
	pages = {878--881},
	booktitle = {2017 {IEEE} International Conference on Web Services ({ICWS})},
	author = {Salama, Usama and Yao, Lina and Wang, Xianzhi and Paik, Hye-Young and Beheshti, Amin},
	date = {2017-06},
}

@inproceedings{guo_fsar_2023,
	title = {{FSAR}: Federated Skeleton-based Action Recognition with Adaptive Topology Structure and Knowledge Distillation},
	issn = {2380-7504},
	doi = {10.1109/ICCV51070.2023.00954},
	shorttitle = {{FSAR}},
	eventtitle = {2023 {IEEE}/{CVF} International Conference on Computer Vision ({ICCV})},
	pages = {10366--10376},
	booktitle = {2023 {IEEE}/{CVF} International Conference on Computer Vision ({ICCV})},
	author = {Guo, Jingwen and Liu, Hong and Sun, Shitong and Guo, Tianyu and Zhang, Min and Si, Chenyang},
	date = {2023-10},
	note = {{ISSN}: 2380-7504},
}

@inproceedings{al-obaidi_privacy_2019,
	title = {Privacy protected recognition of activities of daily living in video},
	doi = {10.1049/cp.2019.0101},
	eventtitle = {3rd {IET} International Conference on Technologies for Active and Assisted Living ({TechAAL} 2019)},
	pages = {1--6},
	booktitle = {3rd {IET} International Conference on Technologies for Active and Assisted Living ({TechAAL} 2019)},
	author = {Al-Obaidi, Salah and Abhayaratne, Charith},
	date = {2019-03},
}

@inproceedings{kim_feddet_2025,
	title = {{FedDet}: Data Poisoning Attack Detection for Federated Skeleton-based Action Recognition},
	doi = {10.1109/ICRA55743.2025.11128277},
	shorttitle = {{FedDet}},
	eventtitle = {2025 {IEEE} International Conference on Robotics and Automation ({ICRA})},
	pages = {15159--15166},
	booktitle = {2025 {IEEE} International Conference on Robotics and Automation ({ICRA})},
	author = {Kim, Min Hyuk and Lee, Eun-Gi and Yoo, Seok Bong},
	date = {2025-05},
}

@inproceedings{shahi_impacts_2022,
	title = {Impacts of Image Obfuscation on Fine-grained Activity Recognition in Egocentric Video},
	doi = {10.1109/PerComWorkshops53856.2022.9767447},
	eventtitle = {2022 {IEEE} International Conference on Pervasive Computing and Communications Workshops and other Affiliated Events ({PerCom} Workshops)},
	pages = {341--346},
	booktitle = {2022 {IEEE} International Conference on Pervasive Computing and Communications Workshops and other Affiliated Events ({PerCom} Workshops)},
	author = {Shahi, Soroush and Alharbi, Rawan and Gao, Yang and Sen, Sougata and Katsaggelos, Aggelos K and Hester, Josiah and Alshurafa, Nabil},
	date = {2022-03},
}

@article{sepehri_privacy-preserving_2023,
	title = {Privacy-Preserving Image Acquisition for Neural Vision Systems},
	volume = {25},
	issn = {1941-0077},
	doi = {10.1109/TMM.2022.3207018},
	pages = {6232--6244},
	journaltitle = {{IEEE} Transactions on Multimedia},
	author = {Sepehri, Yamin and Pad, Pedram and K{\"u}ndig, Cl{\'e}ment and Frossard, Pascal and Dunbar, L. Andrea},
	date = {2023},
}

@inproceedings{baselizadeh_privacy-preserving_2025,
	title = {Privacy-Preserving 3D Lidar-Based Multi-Modal Activity Recognition in Human-Robot Interaction},
	issn = {2837-5149},
	doi = {10.1109/ICCMA67641.2025.11369549},
	eventtitle = {2025 13th International Conference on Control, Mechatronics and Automation ({ICCMA})},
	pages = {516--523},
	booktitle = {2025 13th International Conference on Control, Mechatronics and Automation ({ICCMA})},
	author = {Baselizadeh, Adel and Uddin, Md Zia and Khaksar, Weria and Lindblom, Diana Saplacan and Torresen, Jim},
	date = {2025-11},
	note = {{ISSN}: 2837-5149},
}

@inproceedings{triess_exploring_2024,
	title = {Exploring {AI}-Based Anonymization of Industrial Image and Video Data in the Context of Feature Preservation},
	issn = {2076-1465},
	doi = {10.23919/EUSIPCO63174.2024.10715265},
	eventtitle = {2024 32nd European Signal Processing Conference ({EUSIPCO})},
	pages = {471--475},
	booktitle = {2024 32nd European Signal Processing Conference ({EUSIPCO})},
	author = {Triess, Sabrina C. and Leitritz, Timo and Jauch, Christian},
	date = {2024-08},
	note = {{ISSN}: 2076-1465},
}

@inproceedings{hukkelas_does_2023,
	title = {Does Image Anonymization Impact Computer Vision Training?},
	issn = {2160-7516},
	doi = {10.1109/CVPRW59228.2023.00019},
	eventtitle = {2023 {IEEE}/{CVF} Conference on Computer Vision and Pattern Recognition Workshops ({CVPRW})},
	pages = {140--150},
	booktitle = {2023 {IEEE}/{CVF} Conference on Computer Vision and Pattern Recognition Workshops ({CVPRW})},
	author = {Hukkel{\aa}s, H{\aa}kon and Lindseth, Frank},
	date = {2023-06},
	note = {{ISSN}: 2160-7516},
}

@inproceedings{thapar_anonymizing_2021,
	title = {Anonymizing Egocentric Videos},
	issn = {2380-7504},
	doi = {10.1109/ICCV48922.2021.00232},
	eventtitle = {2021 {IEEE}/{CVF} International Conference on Computer Vision ({ICCV})},
	pages = {2300--2309},
	booktitle = {2021 {IEEE}/{CVF} International Conference on Computer Vision ({ICCV})},
	author = {Thapar, Daksh and Nigam, Aditya and Arora, Chetan},
	date = {2021-10},
	note = {{ISSN}: 2380-7504},
}

@inproceedings{barros_personalized_2025,
	title = {Personalized federated learning for sedentary behavior classification with heterogeneous feature distributions under adversarial threats},
	issn = {2161-4407},
	doi = {10.1109/IJCNN64981.2025.11227203},
	eventtitle = {2025 International Joint Conference on Neural Networks ({IJCNN})},
	pages = {1--8},
	booktitle = {2025 International Joint Conference on Neural Networks ({IJCNN})},
	author = {Barros, Pedro H. and Polido, T{\'u}lio and Guevara, Judy C. and Villas, Leandro and Guidoni, Daniel and Da Fonseca, Nelson L. S. and Ramos, Heitor S.},
	date = {2025-06},
	note = {{ISSN}: 2161-4407},
}

@inproceedings{perelli_analysis_2024,
	location = {New York, {NY}, {USA}},
	title = {Analysis of Human Action Recognition Features in Person Identification Systems for Anti-Bullying Applications},
	isbn = {979-8-4007-1159-6},
	doi = {10.1145/3701268.3701288},
	series = {{HCAIep} '24},
	pages = {58},
	booktitle = {Proceedings of the 2024 Conference on Human Centred Artificial Intelligence - Education and Practice},
	publisher = {Association for Computing Machinery},
	author = {Perelli, Gianpaolo and Micheletto, Marco and Orru, Giulia and Avvisati, Giulia and Capozza, Massimo and Luca Marcialis, Gian},
	date = {2024-12-02},
}

@inproceedings{kumar_finding_2020,
	location = {New York, {NY}, {USA}},
	title = {Finding Achilles' Heel: Adversarial Attack on Multi-modal Action Recognition},
	isbn = {978-1-4503-7988-5},
	doi = {10.1145/3394171.3413531},
	series = {{MM} '20},
	shorttitle = {Finding Achilles' Heel},
	pages = {3829--3837},
	booktitle = {Proceedings of the 28th {ACM} International Conference on Multimedia},
	publisher = {Association for Computing Machinery},
	author = {Kumar, Deepak and Kumar, Chetan and Seah, Chun Wei and Xia, Siyu and Shao, Ming},
	date = {2020-10-12},
}

@article{wang_joint_2025,
	title = {Joint Spatiotemporal Adversarial Attacks on Video Transformer Models Through {XAI}-guided Perturbation},
	issn = {1551-6857},
	doi = {10.1145/3766071},
	journaltitle = {{ACM} Trans. Multimedia Comput. Commun. Appl.},
	author = {Wang, Zerui and Liu, Yan},
	date = {2025-09-05},
	note = {Just Accepted},
}

@inproceedings{park_privacy-driven_2025,
	location = {New York, {NY}, {USA}},
	title = {Privacy-Driven Faces: A Survey on Generative Facial De-identification},
	isbn = {979-8-4007-1419-1},
	doi = {10.1145/3709022.3736543},
	series = {{WDC} '25},
	shorttitle = {Privacy-Driven Faces},
	pages = {27--32},
	booktitle = {Proceedings of the 4th Workshop on Security Implications of Deepfakes and Cheapfakes},
	publisher = {Association for Computing Machinery},
	author = {Park, Sunyoung and Kim, Hyunji and Choi, Seul-Ki and Kim, Taeeun and Park, Eunil},
	date = {2025-08-25},
}

@inproceedings{yan_efficient_2021,
	location = {New York, {NY}, {USA}},
	title = {Efficient Sparse Attacks on Videos using Reinforcement Learning},
	isbn = {978-1-4503-8651-7},
	doi = {10.1145/3474085.3475395},
	series = {{MM} '21},
	pages = {2326--2334},
	booktitle = {Proceedings of the 29th {ACM} International Conference on Multimedia},
	publisher = {Association for Computing Machinery},
	author = {Yan, Huanqian and Wei, Xingxing},
	date = {2021-10-17},
}

@article{zhang_backdoor_2024,
	title = {Backdoor Attacks and Defenses Targeting Multi-Domain {AI} Models: A Comprehensive Review},
	volume = {57},
	issn = {0360-0300},
	doi = {10.1145/3704725},
	shorttitle = {Backdoor Attacks and Defenses Targeting Multi-Domain {AI} Models},
	pages = {87:1--87:35},
	number = {4},
	journaltitle = {{ACM} Comput. Surv.},
	author = {Zhang, Shaobo and Pan, Yimeng and Liu, Qin and Yan, Zheng and Choo, Kim-Kwang Raymond and Wang, Guojun},
	date = {2024-12-10},
}

@article{zheng_efficient_2025,
	title = {Efficient Privacy-Preserving Video Analytics via Share Transforming in Distributed Clouds},
	volume = {21},
	issn = {1551-6857},
	doi = {10.1145/3744248},
	pages = {228:1--228:29},
	number = {8},
	journaltitle = {{ACM} Trans. Multimedia Comput. Commun. Appl.},
	author = {Zheng, Tengfei and Wang, Bo and Li, Gen and Tang, Yuxing and Dou, Qiang},
	date = {2025-08-12},
}

@inproceedings{ko_privmon_2023,
	location = {New York, {NY}, {USA}},
	title = {{PrivMon}: A Stream-Based System for Real-Time Privacy Attack Detection for Machine Learning Models},
	isbn = {979-8-4007-0765-0},
	doi = {10.1145/3607199.3607232},
	series = {{RAID} '23},
	shorttitle = {{PrivMon}},
	pages = {264--281},
	booktitle = {Proceedings of the 26th International Symposium on Research in Attacks, Intrusions and Defenses},
	publisher = {Association for Computing Machinery},
	author = {Ko, Myeongseob and Yang, Xinyu and Ji, Zhengjie and Just, Hoang Anh and Gao, Peng and Kumar, Anoop and Jia, Ruoxi},
	date = {2023-10-16},
}

@article{zhao_federation_2025,
	title = {The Federation Strikes Back: A Survey of Federated Learning Privacy Attacks, Defenses, Applications, and Policy Landscape},
	volume = {57},
	issn = {0360-0300},
	doi = {10.1145/3724113},
	shorttitle = {The Federation Strikes Back},
	pages = {230:1--230:37},
	number = {9},
	journaltitle = {{ACM} Comput. Surv.},
	author = {Zhao, Joshua and Bagchi, Saurabh and Avestimehr, Salman and Chan, Kevin and Chaterji, Somali and Dimitriadis, Dimitris and Li, Jiacheng and Li, Ninghui and Nourian, Arash and Roth, Holger},
	date = {2025-04-03},
}

@inproceedings{singh_human_2021,
	location = {New York, {NY}, {USA}},
	title = {Human Attributes Prediction under Privacy-preserving Conditions},
	isbn = {978-1-4503-8651-7},
	doi = {10.1145/3474085.3475687},
	series = {{MM} '21},
	pages = {4698--4706},
	booktitle = {Proceedings of the 29th {ACM} International Conference on Multimedia},
	publisher = {Association for Computing Machinery},
	author = {Singh, Anshu and Fan, Shaojing and Kankanhalli, Mohan},
	date = {2021-10-17},
}

@inproceedings{wei_ppgnn_2024,
	location = {New York, {NY}, {USA}},
	title = {{PPGNN}: Fast and Accurate Privacy-Preserving Graph Neural Network Inference via Parallel and Pipelined Arithmetic-and-Logic {FHE} Accelerator},
	isbn = {979-8-4007-0601-1},
	doi = {10.1145/3649329.3656517},
	series = {{DAC} '24},
	shorttitle = {{PPGNN}},
	pages = {1--6},
	booktitle = {Proceedings of the 61st {ACM}/{IEEE} Design Automation Conference},
	publisher = {Association for Computing Machinery},
	author = {Wei, Yuntao and Wang, Xueyan and Bian, Song and Huang, Yicheng and Zhao, Weisheng and Jin, Yier},
	date = {2024-11-07},
}

@inproceedings{singh_real-time_2023,
	location = {New York, {NY}, {USA}},
	title = {Real-Time Privacy Preserving Human Activity Recognition on Mobile using 1DCNN-{BiLSTM} Deep Learning},
	isbn = {978-1-4503-9838-1},
	doi = {10.1145/3591156.3591159},
	series = {{IVSP} '23},
	pages = {18--26},
	booktitle = {Proceedings of the 2023 5th International Conference on Image, Video and Signal Processing},
	publisher = {Association for Computing Machinery},
	author = {Singh, Ishneet Sukhvinder and Kaza, Pradyoth and Hosler Iv, Peter Gregory and Chin, Zheng Yang and Ang, Kai Keng},
	date = {2023-06-16},
}

@inproceedings{qi_privacy_2016,
	location = {New York, {NY}, {USA}},
	title = {Privacy preserving via interval covering based subclass division and manifold learning based bi-directional obfuscation for effort estimation},
	isbn = {978-1-4503-3845-5},
	doi = {10.1145/2970276.2970302},
	series = {{ASE} '16},
	pages = {75--86},
	booktitle = {Proceedings of the 31st {IEEE}/{ACM} International Conference on Automated Software Engineering},
	publisher = {Association for Computing Machinery},
	author = {Qi, Fumin and Jing, Xiao-Yuan and Zhu, Xiaoke and Wu, Fei and Cheng, Li},
	date = {2016-08-25},
}

@inproceedings{jin_frequency_2025,
	location = {New York, {NY}, {USA}},
	title = {Frequency Domain Distributed Perturbations: Towards Query-Efficient Black-Box Adversarial Video Attack},
	isbn = {979-8-4007-2035-2},
	doi = {10.1145/3746027.3755456},
	series = {{MM} '25},
	shorttitle = {Frequency Domain Distributed Perturbations},
	pages = {8360--8368},
	booktitle = {Proceedings of the 33rd {ACM} International Conference on Multimedia},
	publisher = {Association for Computing Machinery},
	author = {Jin, Teng and He, Ziwen and Fu, Zhangjie and Wang, Songping and Lyu, Yueming and Shi, Yufei},
	date = {2025-10-27},
}

@article{zhao_audio-visual_2024,
	title = {Audio-Visual Contrastive Pre-train for Face Forgery Detection},
	volume = {21},
	issn = {1551-6857},
	doi = {10.1145/3651311},
	pages = {45:1--45:16},
	number = {2},
	journaltitle = {{ACM} Trans. Multimedia Comput. Commun. Appl.},
	author = {Zhao, Hanqing and Zhou, Wenbo and Chen, Dongdong and Zhang, Weiming and Guo, Ying and Cheng, Zhen and Yan, Pengfei and Yu, Nenghai},
	date = {2024-12-24},
}

@article{huh_novel_2025,
	title = {A Novel Intelligent Video Surveillance System Using Low-Traffic Scene-Preserving Video Anonymization},
	volume = {16},
	issn = {2157-6904},
	doi = {10.1145/3709001},
	pages = {32:1--32:24},
	number = {2},
	journaltitle = {{ACM} Trans. Intell. Syst. Technol.},
	author = {Huh, Jungwoo and Kang, Jiwoo and Woo, Jongwook and Lee, Sanghoon},
	date = {2025-02-15},
}

@inproceedings{adra_e2priv_2025,
	location = {New York},
	title = {E2PRIV: Privacy-Preserving Event-to-Video Reconstruction with Face Anonymization},
	isbn = {979-8-3315-3716-6 979-8-3315-3715-9},
	doi = {10.1109/IWBF63717.2025.11113401},
	shorttitle = {E2PRIV},
	eventtitle = {13th International Workshop on Biometrics and Forensics-{IWBF}},
	pages = {110--115},
	booktitle = {2025 13th International Workshop on Biometrics and Forensics, Iwbf},
	publisher = {{IEEE}},
	author = {Adra, Mira and Dugelay, Jean-Luc},
	editor = {Sequeira, A. F. and Raja, K.},
	date = {2025},
	note = {Num Pages: 6
Web of Science {ID}: {WOS}:001572133900018},
}

@article{wu_video_2025,
	location = {Henderson},
	title = {Video Action Recognition Method Based on Personalized Federated Learning and Spatiotemporal Features},
	volume = {83},
	issn = {1546-2218, 1546-2226},
	doi = {10.32604/cmc.2025.061396},
	pages = {4961--4978},
	number = {3},
	journaltitle = {Cmc-Computers Materials \& Continua},
	shortjournal = {{CMC}-Comput. Mat. Contin.},
	publisher = {Tech Science Press},
	author = {Wu, Rongsen and Xu, Jie and Zhang, Yuhang and Zhao, Changming and Xie, Yiweng and Wu, Zelei and Li, Yunji and Guo, Jinhong and Tang, Shiyang},
	date = {2025},
	note = {Num Pages: 18
Web of Science {ID}: {WOS}:001498222000001},
}

@article{park_human_2025,
	location = {Piscataway},
	title = {Human Daily Indoor Action ({HDIA}) Dataset: Privacy-Preserving Human Action Recognition Using Infrared Camera and Wearable Armband Sensors},
	volume = {13},
	issn = {2169-3536},
	doi = {10.1109/ACCESS.2025.3556001},
	shorttitle = {Human Daily Indoor Action ({HDIA}) Dataset},
	pages = {60822--60832},
	journaltitle = {Ieee Access},
	shortjournal = {{IEEE} Access},
	publisher = {Ieee-Inst Electrical Electronics Engineers Inc},
	author = {Park, Jongbum and Ok Yang, Kyoung and Park, Sunme and Choi, Jun Won},
	date = {2025},
	note = {Num Pages: 11
Web of Science {ID}: {WOS}:001464931200009},
}

@article{zhang_chaotic_2025,
	location = {Amsterdam},
	title = {Chaotic loss-based spiking neural network for privacy-preserving bullying detection in public places},
	volume = {169},
	issn = {1568-4946, 1872-9681},
	doi = {10.1016/j.asoc.2024.112643},
	pages = {112643},
	journaltitle = {Applied Soft Computing},
	shortjournal = {Appl. Soft. Comput.},
	publisher = {Elsevier},
	author = {Zhang, Jing and Yang, Tianlang and Jiang, Cheng and Liu, Jingwei and Zhang, Haoran},
	date = {2025-01},
	note = {Num Pages: 12
Web of Science {ID}: {WOS}:001390933600001},
}

@inproceedings{zadeh_self-supervised_2021,
	location = {New York, {NY}, {USA}},
	title = {Self-Supervised Human Activity Recognition by Augmenting Generative Adversarial Networks},
	isbn = {978-1-4503-8792-7},
	doi = {10.1145/3453892.3453893},
	series = {{PETRA} '21},
	pages = {171--176},
	booktitle = {Proceedings of the 14th {PErvasive} Technologies Related to Assistive Environments Conference},
	publisher = {Association for Computing Machinery},
	author = {Zadeh, Mohammad Zaki and Ramesh Babu, Ashwin and Jaiswal, Ashish and Kyrarini, Maria and Makedon, Fillia},
	date = {2021-06-29},
}

@article{barthelemy_safety_2024,
	location = {Basel},
	title = {Safety After Dark: A Privacy Compliant and Real-Time Edge Computing Intelligent Video Analytics for Safer Public Transportation},
	volume = {24},
	issn = {1424-8220},
	doi = {10.3390/s24248102},
	shorttitle = {Safety After Dark},
	pages = {8102},
	number = {24},
	journaltitle = {Sensors},
	shortjournal = {Sensors},
	publisher = {{MDPI}},
	author = {Barthelemy, Johan and Iqbal, Umair and Qian, Yan and Amirghasemi, Mehrdad and Perez, Pascal},
	date = {2024-12},
	note = {Num Pages: 17
Web of Science {ID}: {WOS}:001386760300001},
}

@article{wang_cloud-based_2024,
	location = {Dordrecht},
	title = {Cloud-based secure human action recognition with fully homomorphic encryption},
	volume = {81},
	issn = {0920-8542, 1573-0484},
	doi = {10.1007/s11227-024-06512-z},
	pages = {12},
	number = {1},
	journaltitle = {Journal of Supercomputing},
	shortjournal = {J. Supercomput.},
	publisher = {Springer},
	author = {Wang, Ruyan and Zeng, Qinglin and Yang, Zhigang and Zhang, Puning},
	date = {2024-10-16},
	note = {Num Pages: 27
Web of Science {ID}: {WOS}:001333859100003},
}

@inproceedings{shim_mosaic_2023,
	location = {New York},
	title = {Mosaic: Extremely Low-resolution {RFID} Vision for Visually-anonymized Action Recognition},
	isbn = {979-8-4007-0118-4},
	doi = {10.1145/3583120.3586968},
	shorttitle = {Mosaic},
	eventtitle = {22nd {ACM}/{IEEE} International Conference on Information Processing in Sensor Networks ({IPSN})},
	pages = {247--260},
	booktitle = {Proceedings of the 2023 the 22nd International Conference on Information Processing in Sensor Networks, Ipsn 2023},
	publisher = {Assoc Computing Machinery},
	author = {Shim, Seungwoo and Shin, Hyeonho and Cho, Myeongkyun and Lee, Youngki and Shin, Jinwoo and Kim, Song Min},
	date = {2023},
	note = {Num Pages: 14
Web of Science {ID}: {WOS}:001112123000019},
}

@article{climent-perez_privacy-preserving_2022,
	location = {Basel},
	title = {Privacy-Preserving Human Action Recognition with a Many-Objective Evolutionary Algorithm},
	volume = {22},
	issn = {1424-8220},
	doi = {10.3390/s22030764},
	pages = {764},
	number = {3},
	journaltitle = {Sensors},
	shortjournal = {Sensors},
	publisher = {{MDPI}},
	author = {Climent-Perez, Pau and Florez-Revuelta, Francisco},
	date = {2022-02},
	note = {Num Pages: 14
Web of Science {ID}: {WOS}:000760117500001},
}

@inproceedings{hinojosa_privhar_2022,
	location = {Cham},
	title = {{PrivHAR}: Recognizing Human Actions from Privacy-Preserving Lens},
	volume = {13664},
	isbn = {978-3-031-19771-0 978-3-031-19772-7},
	issn = {0302-9743, 1611-3349},
	doi = {10.1007/978-3-031-19772-7_19},
	shorttitle = {{PrivHAR}},
	eventtitle = {17th European Conference on Computer Vision ({ECCV})},
	pages = {314--332},
	booktitle = {Computer Vision - Eccv 2022, Pt Iv},
	publisher = {Springer International Publishing Ag},
	author = {Hinojosa, Carlos and Marquez, Miguel and Arguello, Henry and Adeli, Ehsan and Fei-Fei, Li and Niebles, Juan Carlos},
	editor = {Avidan, S. and Brostow, G. and Cisse, M. and Farinella, G. M. and Hassner, T.},
	date = {2022},
	note = {Num Pages: 19
Series Title: Lecture Notes in Computer Science
Web of Science {ID}: {WOS}:000898297000019},
}

@article{rodriguez-conde_-device_2021,
	location = {Basel},
	title = {On-Device Object Detection for More Efficient and Privacy-Compliant Visual Perception in Context-Aware Systems},
	volume = {11},
	issn = {2076-3417},
	doi = {10.3390/app11199173},
	pages = {9173},
	number = {19},
	journaltitle = {Applied Sciences-Basel},
	shortjournal = {Appl. Sci.-Basel},
	publisher = {{MDPI}},
	author = {Rodriguez-Conde, Ivan and Campos, Celso and Fdez-Riverola, Florentino},
	date = {2021-10},
	note = {Num Pages: 34
Web of Science {ID}: {WOS}:000707895400001},
}

@inproceedings{liu_local_2018,
	location = {Paris},
	title = {Local Chaotic Encryption Based on Privacy Protection on Video Surveillance},
	volume = {238},
	isbn = {978-94-6252-535-1},
	issn = {2352-5398},
	eventtitle = {8th International Conference on Social Science and Education Research ({SSER})},
	pages = {390--393},
	booktitle = {Proceedings of the 2018 8th International Conference on Social Science and Education Research (sser 2018)},
	publisher = {Atlantis Press},
	author = {Liu, Suolan and Kong, Lizhi},
	editor = {Wang, Z. and Kun, Z. and Miracle, J.},
	date = {2018},
	note = {Num Pages: 4
Series Title: Advances in Social Science Education and Humanities Research
Web of Science {ID}: {WOS}:000468213400082},
}

@inproceedings{huang_semantic_2024,
	location = {New York, {NY}, {USA}},
	title = {Semantic Privacy-Preserving for Video Surveillance Services on the Edge},
	isbn = {979-8-4007-0123-8},
	doi = {10.1145/3583740.3626820},
	series = {{SEC} '23},
	pages = {300--305},
	booktitle = {Proceedings of the Eighth {ACM}/{IEEE} Symposium on Edge Computing},
	publisher = {Association for Computing Machinery},
	author = {Huang, Alexander Y. C. and Chen, Yitao and Huang, Dijiang and Zhao, Ming},
	date = {2024-08-07},
}

@article{li_survey_2024,
	title = {A Survey of Robustness and Safety of 2D and 3D Deep Learning Models against Adversarial Attacks},
	volume = {56},
	issn = {0360-0300},
	doi = {10.1145/3636551},
	pages = {138:1--138:37},
	number = {6},
	journaltitle = {{ACM} Comput. Surv.},
	author = {Li, Yanjie and Xie, Bin and Guo, Songtao and Yang, Yuanyuan and Xiao, Bin},
	date = {2024-01-22},
}

@article{wang_generative_2021,
	title = {Generative Adversarial Networks in Computer Vision: A Survey and Taxonomy},
	volume = {54},
	issn = {0360-0300},
	doi = {10.1145/3439723},
	shorttitle = {Generative Adversarial Networks in Computer Vision},
	pages = {37:1--37:38},
	number = {2},
	journaltitle = {{ACM} Comput. Surv.},
	author = {Wang, Zhengwei and She, Qi and Ward, Tom{\'a}s E.},
	date = {2021-02-09},
}

@inproceedings{li2024privacy,
  title={Privacy-Preserving Action Recognition: A Survey},
  author={Li, Xiao and Qiu, Yu-Kun and Peng, Yi-Xing and Zeng, Ling-An and Zheng, Wei-Shi},
  booktitle={Chinese Conference on Pattern Recognition and Computer Vision (PRCV)},
  pages={454--468},
  year={2024},
  doi = {10.1007/978-981-97-8511-7_32},
  organization={Springer}
}

@inproceedings{carr_review_2024,
	title = {A Review of Privacy and Utility in Skeleton-based Data in Virtual Reality Metaverses},
	doi = {10.1109/MetaCom62920.2024.00041},
	eventtitle = {2024 {IEEE} International Conference on Metaverse Computing, Networking, and Applications ({MetaCom})},
	pages = {198--205},
	booktitle = {2024 {IEEE} International Conference on Metaverse Computing, Networking, and Applications ({MetaCom})},
	author = {Carr, Thomas and Xu, Depeng and Lu, Aidong},
	date = {2024-08},
}

@inproceedings{dai_towards_2015-1,
	title = {Towards privacy-preserving activity recognition using extremely low temporal and spatial resolution cameras},
	issn = {2160-7516},
	doi = {10.1109/CVPRW.2015.7301356},
	eventtitle = {2015 {IEEE} Conference on Computer Vision and Pattern Recognition Workshops ({CVPRW})},
	pages = {68--76},
	booktitle = {2015 {IEEE} Conference on Computer Vision and Pattern Recognition Workshops ({CVPRW})},
	author = {Dai, Ji and Wu, Jonathan and Saghafi, Behrouz and Konrad, Janusz and Ishwar, Prakash},
	date = {2015-06},
	note = {{ISSN}: 2160-7516},
}

@inproceedings{chaudhary_deep_2022,
	title = {Deep Network for Extremely Low-Resolution Human Action Recognition},
	doi = {10.1109/AVSS56176.2022.9959612},
	eventtitle = {2022 18th {IEEE} International Conference on Advanced Video and Signal Based Surveillance ({AVSS})},
	pages = {1--8},
	booktitle = {2022 18th {IEEE} International Conference on Advanced Video and Signal Based Surveillance ({AVSS})},
	author = {Chaudhary, Sachin and Patil, Prashant W. and Dudhane, Akshay and Murala, Subrahmanyam},
	date = {2022-11},
}

@article{bai_extreme_2023,
	location = {Dordrecht},
	title = {Extreme Low-Resolution Action Recognition with Confident Spatial-Temporal Attention Transfer},
	volume = {131},
	issn = {0920-5691, 1573-1405},
	doi = {10.1007/s11263-023-01771-4},
	pages = {1550--1565},
	number = {6},
	journaltitle = {International Journal of Computer Vision},
	shortjournal = {Int. J. Comput. Vis.},
	publisher = {Springer},
	author = {Bai, Yucai and Zou, Qin and Chen, Xieyuanli and Li, Lingxi and Ding, Zhengming and Chen, Long},
	date = {2023-06},
	note = {Num Pages: 16
Web of Science {ID}: {WOS}:000945810600001},
}

@inproceedings{mucha_beyond_2022,
	location = {New York, {NY}, {USA}},
	title = {Beyond Privacy of Depth Sensors in Active and Assisted Living Devices},
	isbn = {978-1-4503-9631-8},
	doi = {10.1145/3529190.3534764},
	series = {{PETRA} '22},
	pages = {425--429},
	booktitle = {Proceedings of the 15th International Conference on {PErvasive} Technologies Related to Assistive Environments},
	publisher = {Association for Computing Machinery},
	author = {Mucha, Wiktor and Kampel, Martin},
	date = {2022-07-11},
}

@inproceedings{ballester_action_2024,
	location = {Cham},
	title = {Action Recognition from 4D Point Clouds for Privacy-Sensitive Scenarios in Assistive Contexts},
	volume = {14751},
	isbn = {978-3-031-62848-1 978-3-031-62849-8},
	issn = {0302-9743, 1611-3349},
	doi = {10.1007/978-3-031-62849-8_44},
	eventtitle = {19th International Conference on Computers Helping People with Special Needs ({ICCHP})},
	pages = {359--364},
	booktitle = {Computers Helping People with Special Needs, Pt Ii, Icchp 2024},
	publisher = {Springer International Publishing Ag},
	author = {Ballester, Irene and Kampel, Martin},
	editor = {Miesenberger, K. and Penaz, P. and Kobayashi, M.},
	date = {2024},
	note = {Num Pages: 6
Series Title: Lecture Notes in Computer Science
Web of Science {ID}: {WOS}:001313663100043},
}

@inproceedings{zong_privacy-preserving_2021,
	title = {Privacy-Preserving Automatic Slipping Detection Method for Elderly in Bathroom Using Depth Sensors},
	issn = {1948-9447},
	doi = {10.1109/CCDC52312.2021.9602301},
	eventtitle = {2021 33rd Chinese Control and Decision Conference ({CCDC})},
	pages = {1990--1994},
	booktitle = {2021 33rd Chinese Control and Decision Conference ({CCDC})},
	author = {Zong, Hengshan and Lei, Huan and Jiao, Zeyu and Zhong, Zhengyu},
	date = {2021-05},
	note = {{ISSN}: 1948-9447},
}

@inproceedings{zakka_action_2024,
	location = {Cham},
	title = {Action Recognition for Privacy-Preserving Ambient Assisted Living},
	volume = {14976},
	isbn = {978-3-031-67284-2 978-3-031-67285-9},
	issn = {0302-9743, 1611-3349},
	doi = {10.1007/978-3-031-67285-9_15},
	eventtitle = {1st International Conference on Artificial Intelligence in Healthcare ({AIiH})},
	pages = {203--217},
	booktitle = {Artificial Intelligence in Healthcare, Pt Ii, Aiih 2024},
	publisher = {Springer International Publishing Ag},
	author = {Zakka, Vincent Gbouna and Dai, Zhuangzhuang and Manso, Luis J.},
	editor = {Xie, X. and Styles, I. and Powathil, G. and Ceccarelli, M.},
	date = {2024},
	note = {Num Pages: 15
Series Title: Lecture Notes in Computer Science
Web of Science {ID}: {WOS}:001308382500015},
}

@article{rajput_privacy-preserving_2020,
	location = {Oxford},
	title = {Privacy-preserving human action recognition as a remote cloud service using {RGB}-D sensors and deep {CNN}},
	volume = {152},
	issn = {0957-4174, 1873-6793},
	doi = {10.1016/j.eswa.2020.113349},
	pages = {113349},
	journaltitle = {Expert Systems with Applications},
	shortjournal = {Expert Syst. Appl.},
	publisher = {Pergamon-Elsevier Science Ltd},
	author = {Rajput, Amitesh Singh and Raman, Balasubramanian and Imran, Javed},
	date = {2020-08-15},
	note = {Num Pages: 15
Web of Science {ID}: {WOS}:000532801200022},
}

@inproceedings{hadano_multiaae_2025,
	title = {{MultiAAE}: Lightweight Anonymizing Autoencoder for Privacy-Aware Multi-Sensor Human Activity Recognition},
	doi = {10.23919/ICMU65253.2025.11219119},
	shorttitle = {{MultiAAE}},
	eventtitle = {2025 Fifteenth International Conference on Mobile Computing and Ubiquitous Networking ({ICMU})},
	pages = {1--6},
	booktitle = {2025 Fifteenth International Conference on Mobile Computing and Ubiquitous Networking ({ICMU})},
	author = {Hadano, Musashi and Nakamura, Yugo and Arakawa, Yutaka},
	date = {2025-09},
}

@inproceedings{carr_explanation-based_2025,
	location = {Singapore},
	title = {Explanation-Based Anonymization Methods for Motion Privacy},
	volume = {15873},
	isbn = {978-981-96-8182-2 978-981-96-8183-9},
	issn = {2945-9133, 1611-3349},
	doi = {10.1007/978-981-96-8183-9_5},
	eventtitle = {29th Pacific Asia Conference on Knowledge Discovery and Data Mining-{PAKDD}-Annual},
	pages = {52--64},
	booktitle = {Advances in Knowledge Discovery and Data Mining, Pakdd 2025, Pt Iv},
	publisher = {Springer-Verlag Singapore Pte Ltd},
	author = {Carr, Thomas and Zhao, Yaxin and Xu, Depeng and Lu, Aidong},
	editor = {Wu, X. and Spiliopoulou, M. and Wang, C. and Kumar, V. and Cao, L. and Wu, Y. and Yao, Y. and Wu, Z.},
	date = {2025},
	note = {Num Pages: 13
Series Title: Lecture Notes in Artificial Intelligence
Web of Science {ID}: {WOS}:001584727800005},
}

@inproceedings{aslam_balancing_2025,
	title = {Balancing Privacy and Action Performance: A Penalty-Driven Approach to Image Anonymization},
	issn = {2160-7516},
	doi = {10.1109/CVPRW67362.2025.00077},
	shorttitle = {Balancing Privacy and Action Performance},
	eventtitle = {2025 {IEEE}/{CVF} Conference on Computer Vision and Pattern Recognition Workshops ({CVPRW})},
	pages = {720--729},
	booktitle = {2025 {IEEE}/{CVF} Conference on Computer Vision and Pattern Recognition Workshops ({CVPRW})},
	author = {Aslam, Nazia and Nasrollahi, Kamal},
	date = {2025-06},
	note = {{ISSN}: 2160-7516},
}

@article{ismail_stealthguard_2025,
	location = {Geneva},
	title = {{StealthGuard}: a new framework of privacy-preserving human action recognition},
	volume = {27},
	issn = {1744-1765, 1744-1773},
	doi = {10.1504/IJICS.2025.146882},
	shorttitle = {{StealthGuard}},
	number = {2},
	journaltitle = {International Journal of Information and Computer Security},
	shortjournal = {Int. J. Inf. Comput. Secur.},
	publisher = {Inderscience Enterprises Ltd},
	author = {Ismail, Gazi Mohammad and Zhang, Xueping and Yang, Junxiang and Li, Bin},
	date = {2025},
	note = {Num Pages: 22
Web of Science {ID}: {WOS}:001515342200003},
}

@inproceedings{li_supplementary_2025,
	title = {Supplementary Material for ``{NoiseActor}: A Noise-Action Collaborative Framework for Privacy-Preserving Action Recognition without Privacy Labels''},
	issn = {1945-788X},
	doi = {10.1109/ICME59968.2025.11209811},
	shorttitle = {Supplementary Material for ``{NoiseActor}},
	eventtitle = {2025 {IEEE} International Conference on Multimedia and Expo ({ICME})},
	pages = {1--3},
	booktitle = {2025 {IEEE} International Conference on Multimedia and Expo ({ICME})},
	author = {Li, Xiao and Wu, Xiao-Ming and Zhang, Delong and Lin, Kun-Yu and Peng, Yi-Xing and Zeng, Ling-An and Zheng, Wei-Shi},
	date = {2025-06},
	note = {{ISSN}: 1945-788X},
}

@inproceedings{yan_image_2020,
	title = {Image Segmentation Based Privacy-Preserving Human Action Recognition for Anomaly Detection},
	issn = {2379-190X},
	doi = {10.1109/ICASSP40776.2020.9054456},
	eventtitle = {{ICASSP} 2020 - 2020 {IEEE} International Conference on Acoustics, Speech and Signal Processing ({ICASSP})},
	pages = {8931--8935},
	booktitle = {{ICASSP} 2020 - 2020 {IEEE} International Conference on Acoustics, Speech and Signal Processing ({ICASSP})},
	author = {Yan, Jiawei and Angelini, Federico and Naqvi, Syed Mohsen},
	date = {2020-05},
	note = {{ISSN}: 2379-190X},
}

@article{kim_secure_2022,
	location = {Berlin},
	title = {Secure human action recognition by encrypted neural network inference},
	volume = {13},
	issn = {2041-1723},
	doi = {10.1038/s41467-022-32168-5},
	pages = {4799},
	number = {1},
	journaltitle = {Nature Communications},
	shortjournal = {Nat. Commun.},
	publisher = {Nature Portfolio},
	author = {Kim, Miran and Jiang, Xiaoqian and Lauter, Kristin and Ismayilzada, Elkhan and Shams, Shayan},
	date = {2022-08-15},
	note = {Num Pages: 13
Web of Science {ID}: {WOS}:000840984400004},
}

@inproceedings{khan_privacy-preserving_2024,
	title = {Privacy-Preserving Artificial Intelligence on Edge Devices: A Homomorphic Encryption Approach},
	issn = {2836-3868},
	doi = {10.1109/ICWS62655.2024.00061},
	shorttitle = {Privacy-Preserving Artificial Intelligence on Edge Devices},
	eventtitle = {2024 {IEEE} International Conference on Web Services ({ICWS})},
	pages = {395--405},
	booktitle = {2024 {IEEE} International Conference on Web Services ({ICWS})},
	author = {Khan, Muhammad Jahanzeb and Fang, Bo and Cimino, Gaetano and Cirillo, Stefano and Yang, Lei and Zhao, Dongfang},
	date = {2024-07},
	note = {{ISSN}: 2836-3868},
}

@inproceedings{pham_privacy_2019,
	location = {New York, {NY}, {USA}},
	title = {Privacy Preserving Visual Log Service with Temporal Interval Query using Interval Tree-based Searchable Symmetric Encryption},
	isbn = {978-1-4503-7245-9},
	doi = {10.1145/3368926.3369701},
	series = {{SoICT} '19},
	pages = {425--432},
	booktitle = {Proceedings of the 10th International Symposium on Information and Communication Technology},
	publisher = {Association for Computing Machinery},
	author = {Pham, Viet-An and Hoang, Dinh-Hieu and Chung-Nguyen, Huy-Hoang and Tran, Mai-Khiem and Tran, Minh-Triet},
	date = {2019-12-04},
}

@incollection{atrey_encrypted_2017,
	title = {Encrypted domain multimedia content analysis},
	volume = {17},
	isbn = {978-1-970001-07-5},
	pages = {75--104},
	booktitle = {Frontiers of Multimedia Research},
	publisher = {Association for Computing Machinery and Morgan \& Claypool},
	author = {Atrey, Pradeep K. and Lathey, Ankita and Yakubu, Abukari M.},
	date = {2017-12-19},
}

@inproceedings{doshi_federated_2022,
	location = {New York},
	title = {Federated Learning-based Driver Activity Recognition for Edge Devices},
	isbn = {978-1-6654-8739-9},
	issn = {2160-7508},
	doi = {10.1109/CVPRW56347.2022.00377},
	eventtitle = {{IEEE}/{CVF} Conference on Computer Vision and Pattern Recognition ({CVPR})},
	pages = {3337--3345},
	booktitle = {2022 Ieee/Cvf Conference on Computer Vision and Pattern Recognition Workshops, Cvprw 2022},
	publisher = {{IEEE}},
	author = {Doshi, Keval and Yilmaz, Yasin},
	date = {2022},
	note = {Num Pages: 9
Series Title: {IEEE} Computer Society Conference on Computer Vision and Pattern Recognition Workshops
Web of Science {ID}: {WOS}:000861612703050},
}

@article{dinh_floweraction_2025,
	title = {{FlowerAction}: a federated deep learning framework for video-based human action recognition},
	volume = {16},
	issn = {1868-5137},
	doi = {10.1007/s12652-025-04958-4},
	shorttitle = {{FlowerAction}},
	pages = {459--470},
	number = {2},
	journaltitle = {Journal of Ambient Intelligence and Humanized Computing},
	shortjournal = {J. Ambient Intell. Humanized Comput.},
	publisher = {Springer Science and Business Media Deutschland {GmbH}},
	author = {Dinh, Thi Quynh Khanh and Tran, Thanh-Hai and Tran, Trung-Kien and Le, Thi-Lan},
	date = {2025},
}

@inproceedings{nguyen_fedfslar_2024,
	location = {Los Alamitos},
	title = {{FedFSLAR}: A Federated Learning Framework for Few-shot Action Recognition},
	isbn = {979-8-3503-7028-7 979-8-3503-7071-3},
	issn = {2572-4398},
	doi = {10.1109/WACVW60836.2024.00035},
	shorttitle = {{FedFSLAR}},
	eventtitle = {{IEEE}/{CVF} Winter Conference on Applications of Computer Vision ({WACV})},
	pages = {270--279},
	booktitle = {2024 Ieee Winter Conference on Applications of Computer Vision Workshops, Wacvw 2024},
	publisher = {{IEEE} Computer Soc},
	author = {Nguyen, Anh Tu and Abu, Assanali and Aikyn, Nartay and Makhanov, Nursultan and Lee, Min-Ho and Khiem, Le-Huy and Wong, Kok-Seng},
	date = {2024},
	note = {Num Pages: 10
Series Title: {IEEE} Winter Conference on Applications of Computer Vision Workshops
Web of Science {ID}: {WOS}:001223022200070},
}

@article{tu_benchmarking_2024,
	location = {Piscataway},
	title = {Benchmarking Federated Few-Shot Learning for Video-Based Action Recognition},
	volume = {12},
	issn = {2169-3536},
	doi = {10.1109/ACCESS.2024.3519254},
	pages = {193141--193164},
	journaltitle = {Ieee Access},
	shortjournal = {{IEEE} Access},
	publisher = {Ieee-Inst Electrical Electronics Engineers Inc},
	author = {Tu, Nguyen Anh and Aikyn, Nartay and Makhanov, Nursultan and Abu, Assanali and Wong, Kok-Seng and Lee, Min-Ho},
	date = {2024},
	note = {Num Pages: 24
Web of Science {ID}: {WOS}:001383065500001},
}

@article{kim_fedpure_2026,
	location = {New York},
	title = {{FedPure}: Data poisoning attack detection and purification for federated skeleton-based action recognition},
	volume = {725},
	issn = {0020-0255, 1872-6291},
	doi = {10.1016/j.ins.2025.122733},
	shorttitle = {{FedPure}},
	pages = {122733},
	journaltitle = {Information Sciences},
	shortjournal = {Inf. Sci.},
	publisher = {Elsevier Science Inc},
	author = {Kim, Min Hyuk and Lee, Eun-Gi and Yoo, Seok Bong},
	date = {2026-01},
	note = {Num Pages: 23
Web of Science {ID}: {WOS}:001591484600001},
}

@inproceedings{gad_joint_2023,
	title = {Joint Knowledge Distillation and Local Differential Privacy for Communication-Efficient Federated Learning in Heterogeneous Systems},
	issn = {2576-6813},
	doi = {10.1109/GLOBECOM54140.2023.10437358},
	eventtitle = {{GLOBECOM} 2023 - 2023 {IEEE} Global Communications Conference},
	pages = {2051--2056},
	booktitle = {{GLOBECOM} 2023 - 2023 {IEEE} Global Communications Conference},
	author = {Gad, Gad and Fadlullah, Zubair Md and Fouda, Mostafa M. and Ibrahem, Mohamed I. and Nasser, Nidal},
	date = {2023-12},
	note = {{ISSN}: 2576-6813},
}

@inproceedings{rao_privacy-preserving_2025,
	title = {Privacy-Preserving Facial and Action Recognition Using Federated Deep Learning on Edge Devices},
	doi = {10.1109/GCAT66372.2025.11368496},
	eventtitle = {2025 {IEEE} 6th Global Conference for Advancement in Technology ({GCAT})},
	pages = {1--8},
	booktitle = {2025 {IEEE} 6th Global Conference for Advancement in Technology ({GCAT})},
	author = {Rao, Dustakar Surendra and Kollem, Sreedhar},
	date = {2025-10},
}

@article{yang_cross-modal_2024,
	title = {Cross-Modal Federated Human Activity Recognition},
	volume = {46},
	issn = {1939-3539},
	doi = {10.1109/TPAMI.2024.3367412},
	pages = {5345--5361},
	number = {8},
	journaltitle = {{IEEE} Transactions on Pattern Analysis and Machine Intelligence},
	author = {Yang, Xiaoshan and Xiong, Baochen and Huang, Yi and Xu, Changsheng},
	date = {2024-08},
}

@article{palit_federated_2026,
	title = {Federated Multimodal Fusion for Action Recognition Leveraging Vision-Language Embeddings and {SpatioTemporal} {CNNs}},
	volume = {December 2026},
	issn = {2835-8856},
	journaltitle = {Transactions on Machine Learning Research},
	shortjournal = {Transact. mach. learn. res.},
	publisher = {Transactions on Machine Learning Research},
	author = {Palit, Aditi and Yeturu, Kalidas},
	date = {2026},
}

@article{cao_bones_2026,
	location = {Piscataway},
	title = {Bones of Contention: Exploring Query-Efficient Attacks Against Skeleton Recognition Systems},
	volume = {21},
	issn = {1556-6013, 1556-6021},
	doi = {10.1109/TIFS.2025.3639978},
	shorttitle = {Bones of Contention},
	pages = {183--196},
	journaltitle = {Ieee Transactions on Information Forensics and Security},
	shortjournal = {{IEEE} Trans. Inf. Forensic Secur.},
	publisher = {Ieee-Inst Electrical Electronics Engineers Inc},
	author = {Cao, Yuxin and Ye, Kai and Wang, Derui and Xue, Minhui and Ge, Hao and Qian, Chenxiong and Song Dong, Jin},
	date = {2026},
	note = {Num Pages: 14
Web of Science {ID}: {WOS}:001651962500003},
}
\vfill

\clearpage

\end{document}